\documentclass[journal]{vgtc}                     % final (journal style)
\onlineid{0}

\vgtccategory{Research}

\vgtcpapertype{algorithm/technique}

\title{CD-NGP: A Fast Scalable Continual Representation for Dynamic Scenes}

\author{%
  \authororcid{Zhenhuan Liu.},
  Shuai Liu, Zhiwei Ning, Jie Yang, Yifan Zuo, Yuming Fang, 
  Wei Liu
}

\authorfooter{
  \item
   Zhenhuan Liu, Shuai Liu, Zhiwei Ning, Jie Yang, and Wei Liu are with the  Dept. of Automation, Shanghai Jiao Tong University, Shanghai 200240, China. E-mail: $\{$pignfan\_zhilu\_law268, shuailiu,  zwning, jieyang, weiliucv$\}$@sjtu.edu.cn
  \item Yifan Zuo and Yuming Fang are with the School of Computing and Artificial Intelligence, Jiangxi University of Finance and Economics, Nanchang, Jiangxi 330032, China. E-mail: kenny0410@126.com; fa0001ng@e.ntu.edu.sg.  
  }

\abstract{
Novel view synthesis (NVS) in dynamic scenes faces persistent challenges in memory consumption, model complexity, training efficiency, and rendering quality. Offline methods offer high fidelity but suffer from high memory usage and limited scalability, while online approaches often trade quality for speed and compactness. We propose \textit{Continual Dynamic Neural Graphics Primitives (CD-NGP)}, a continual learning framework that reduces memory overhead and enhances scalability through parameter reuse. To avoid feature interference in dynamic scenes and improve rendering quality, our method combines spatial and temporal hash encodings, which compactly represent scene structures and motion patterns. We also introduce a new dataset comprising multi-view, long-duration ($>1200$ frames) videos with both rigid and non-rigid motion, which is not found in existing benchmarks. CD-NGP is evaluated on public datasets and our long video dataset, demonstrating superior scalability and reconstruction quality. It significantly reduces training memory usage (to $<$14GB) and requires only \textbf{0.4MB/frame} in streaming bandwidth on DyNeRF---substantially lower than most online baselines.%
}


\teaser{
  \centering
  \includegraphics[width=\linewidth, alt={ourline figure.}]{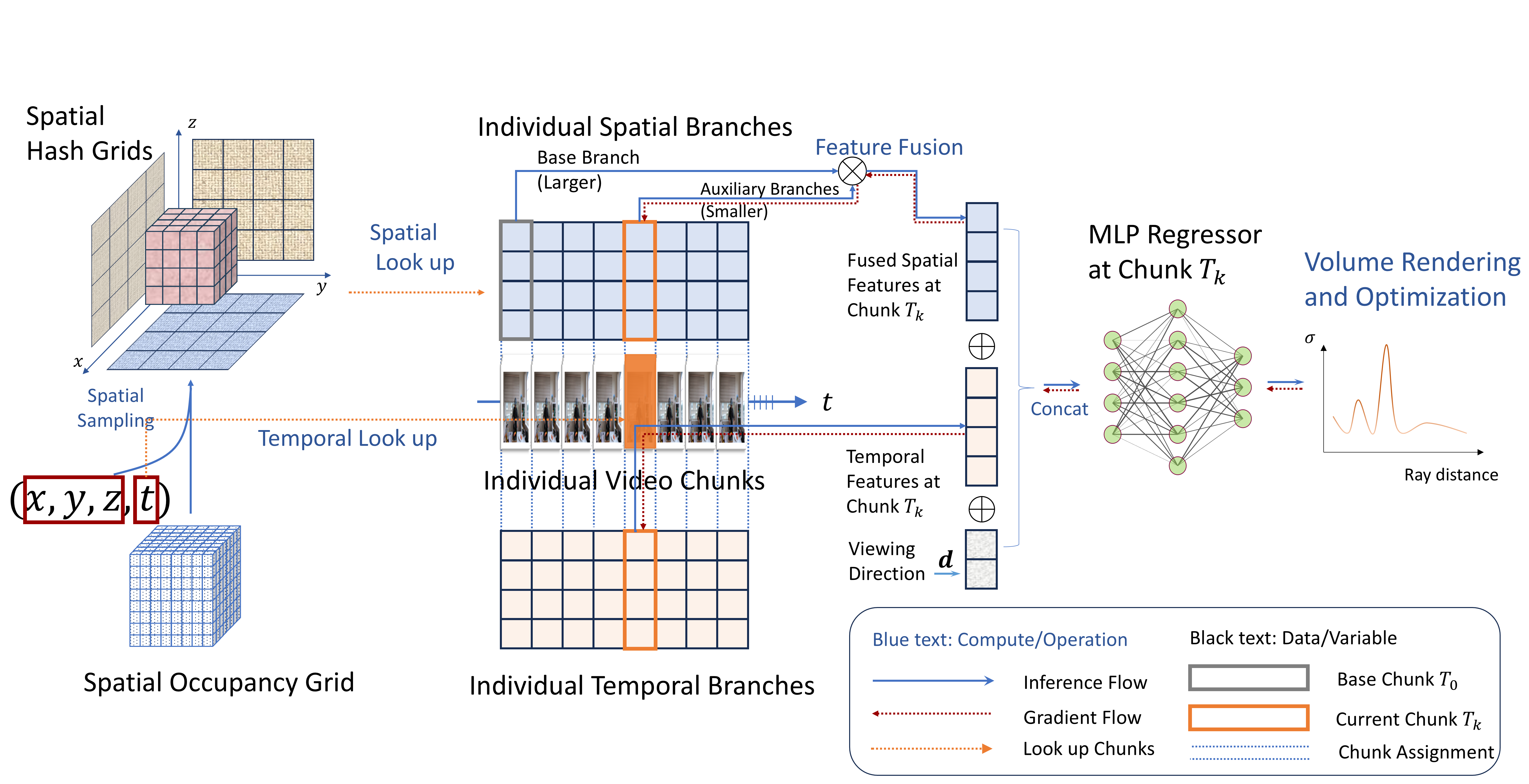}
  \caption{%
  Overview of our continual learning pipeline for dynamic scenes. Our method segments multi-view input videos into multiple chunks and subsequently reconstructs them using individual model branches. Each branch contains individual spatial hash encodings for $(x,y,z)$, temporal hash encodings for $t$, and MLP regressors for downstream rendering and optimization.
Benefiting from the reuse of the larger hash table in the first branch (base branch) and the fusion of features from the subsequent branches (auxiliary branches), our method achieves high scalability and fast convergence. Leveraging the flexible hash encoding structure, it also demonstrates strong adaptability to different encoding types, including 3D voxel grids, 2D plane grids, or hybrid variants. }
  \label{fig:pipeline}
}

\renewcommand{\manuscriptnotetxt}{}

\graphicspath{{figs/}{figures/}{pictures/}{images/}{./}} % where to search for the images

\usepackage[table, svgnames, dvipsnames]{xcolor}
\usepackage{tabu}                      % only used for the table example
\usepackage{booktabs}                  % only used for the table example
\usepackage{lipsum}                    % used to generate placeholder text
\usepackage{mwe}                       % used to generate placeholder figures

\usepackage{amsmath}
\usepackage{amssymb}
\usepackage{multirow}
\usepackage{algorithm}
\usepackage{algorithmic}
\usepackage{wrapfig}
\usepackage{tikz-cd}
\usepackage{mathptmx}                  % use matching math font

\usepackage{colortbl}

\begin{document}

\definecolor{cc1}{RGB}{255,182,193}
\definecolor{cc2}{RGB}{135,206,250}
\definecolor{cc3}{RGB}{255,228,196}

% 定义缩写命令
\makeatletter
\DeclareRobustCommand\onedot{\futurelet\@let@token\@onedot}
\def\@onedot{\ifx\@let@token.\else.\null\fi\xspace}

\def\eg{\emph{e.g}\onedot} 
\def\Eg{\emph{E.g}\onedot}
\def\ie{\emph{i.e}\onedot} 
\def\Ie{\emph{I.e}\onedot}
\def\cf{\emph{cf}\onedot} 
\def\Cf{\emph{Cf}\onedot}
\def\etc{\emph{etc}\onedot} 
\def\vs{\emph{vs}\onedot}
\def\wrt{w.r.t\onedot} 
\def\dof{d.o.f\onedot}
\def\iid{i.i.d\onedot} 
\def\wolog{w.l.o.g\onedot}
\def\etal{\emph{et al}\onedot}
\makeatother

%%%%%%%%%%%%%%%%%%%%%%%%%%%%%%%%%%%%%%%%%%%%%%%%%%%%%%%%%%%%%%%%
%%%%%%%%%%%%%%%%%%%%%% START OF THE PAPER %%%%%%%%%%%%%%%%%%%%%%
%%%%%%%%%%%%%%%%%%%%%%%%%%%%%%%%%%%%%%%%%%%%%%%%%%%%%%%%%%%%%%%%

\firstsection{Introduction}

\maketitle

\label{sec:intro}

Neural radiance field (NeRF) introduced in \cite{nerfeccv} has made significant progress in 3D scene reconstruction and novel view synthesis.
It takes multi-view RGB images with their camera poses as input and leverages multi-layer perceptrons (MLPs) to reconstruct both the geometry and color of the scene.
Many variants have been proposed to improve NeRF's efficiency \cite{tensorf,merf,ingp,bakedsdf,recursivenerf},
 reconstruction quality \cite{Mipnerf,mipnerf360,zipnerf}, 
scalability \cite{F2nerf,scanerf-scalable}, 
 and pose robustness \cite{nopenerf,melonnerf,dsnerf,sinnerf,regnerf,s3im}.
Recently, 3D Gaussian Splatting (3DGS)~\cite{3dgs} proposes an explicit representation with differentiable rendering, which significantly speeds up the reconstruction and enables real-time novel view synthesis.
Despite significant progress having been made in this field, the above-mentioned methods assume that the scene is static and may not work properly in dynamic scenes.

To address the dynamic reconstruction problem, a number of efforts have been made in the offline setting, requiring access to the entire dataset during the training stage. 
For example, the methods in \cite{Dnerf,Dynerf,park2023temporal,hypernerf,NeRFlow,NSFF,neuraltrajectoryF} propose to use MLP-based representation and achieve high-quality results. 
Some methods speed up the optimization by using the tensor factorization (\cite{tensor4d, kplanes,HexPlane_}) and voxel representations \cite{tineuvox,mixvoxels}.  
The more recent continual-learning-based methods \cite{streamRF, invnerf} (also mentioned as online methods) can train models frame by frame, and streamable representations~\cite{nerfplayer} enable the model to be loaded on the fly at inference. 

On the one hand, most offline approaches require loading all video frames of the training set into memory, imposing substantial demands on hardware resources (e.g., approximately 100 GB for the DyNeRF dataset \cite{Dynerf} with 20 views $\times$ 300 frames). Lazy loading reduces peak memory usage \cite{4Dgaussiansplatting}, but still requires prohibitive disk space for full video frame caching. Moreover, the overall resource consumption scales linearly with the number of frames ($O(N_{frames})$), which makes it unsuitable for lengthy video sequences. 
On the other hand, recent online methods \cite{streamRF,invnerf} fail to simultaneously achieve fast convergence and compact model size, which constrains their applicability in long-term dynamic scenarios. 
These limitations motivate us to design a more scalable and memory-efficient representation for modeling dynamic scenes over extended durations.

In this paper, we propose \emph{continual dynamic neural graphics primitives} (CD-NGP), a fast and scalable representation for reconstructing dynamic scenes, which enables dynamic reconstruction in long videos. As illustrated in \cref{fig:pipeline}, CD-NGP derives from three key ideas: dividing the video into segments, encoding spatial and temporal features separately, and reusing features across different segments to ensure high scalability:   
Firstly, we partition dynamic videos into multiple segments according to time stamps $t$ and perform reconstruction using distinct model branches to facilitate continual learning. Each branch of our model includes MLP regressors, spatial hash encoding for spatial coordinates $(x,y,z)$, and temporal hash encoding for the timestamp $t$.  Secondly, the spatial and temporal features are encoded separately to avoid feature interference in different axes. For each model branch, we sample 3D points $(x,y,z)$ according to frame pixels and cached occupancy grids like \cite{ingp}, and encode the points into spatial features using spatial hash grids. We also encode temporal features using hash grids for the separated $t$ axis. These spatial and temporal features and the viewing direction $\mathbf{d}$ are concatenated and input into MLP regressors for rendering and optimization. Thirdly, we leverage the redundancy in the dynamic scenes and fuse the spatial features of different branches to yield high scalability. We set a larger hash table size ($2^{19}$) in the initial branch (the \emph{base branch}), and set a smaller size ($2^{14}$) for the other branches (the \emph{auxiliary branches}). This asymmetric design facilitates reusing the features trained in the base branch, thereby reducing storage consumption and boosting rendering quality.

\begin{figure}[!t]
  \centering
    \includegraphics[width=0.9\linewidth]{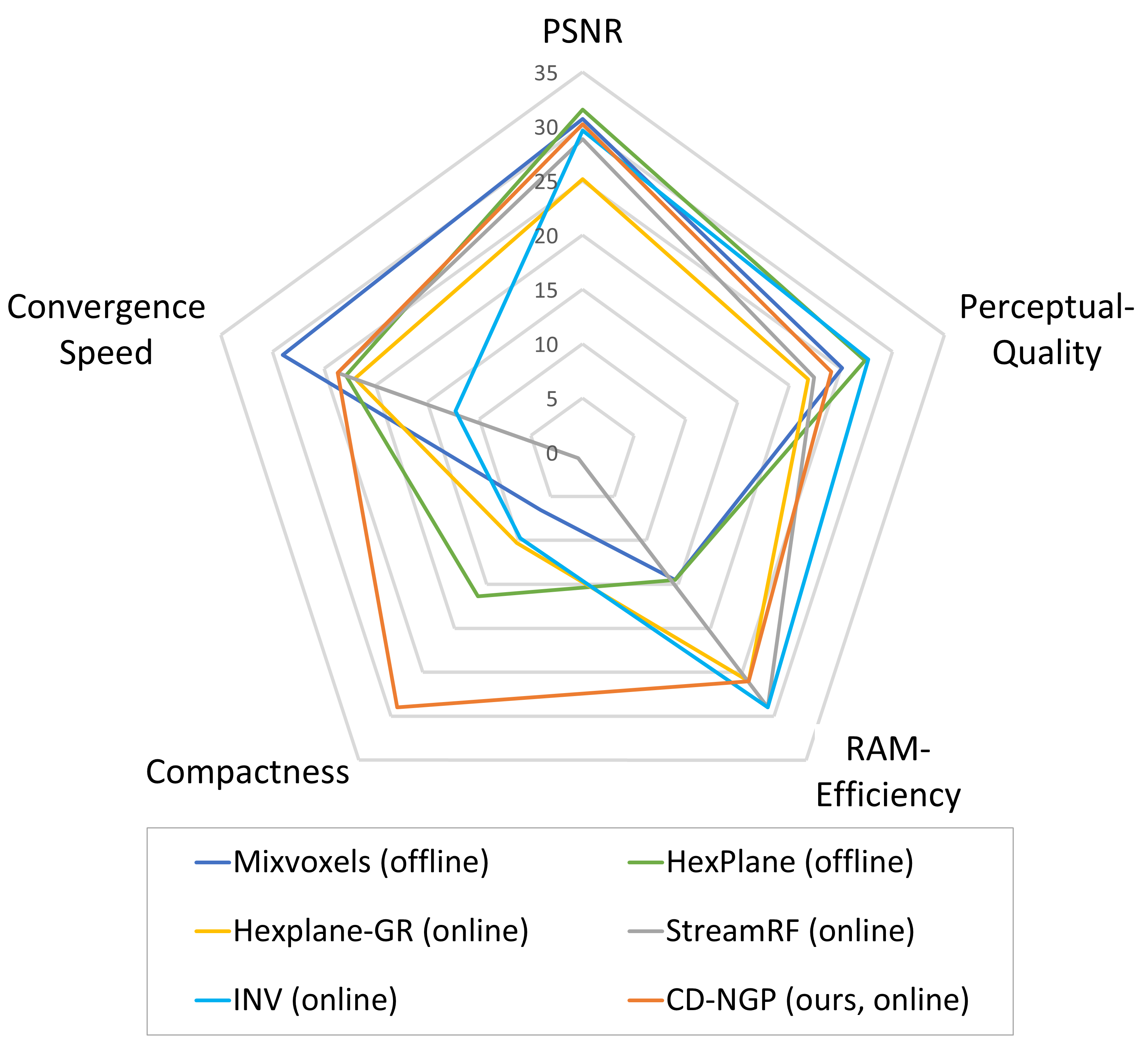}
    \caption{Comparison of various performance indices between our method and recent representative baselines. Perceptual consistency is measured by $1 -$ LPIPS using AlexNet \cite{LPIPS,Alexnet}, and convergence speed is defined as the reciprocal of training time. Compactness and RAM efficiency are derived by applying a $-\log$ transformation to model size and memory usage, respectively. }
  \label{fig:Visualized comparison}
 \end{figure}

We validate the effectiveness of our method through comprehensive experimental results on both widely used public datasets and a self-collected dataset. 
As illustrated in \cref{fig:Visualized comparison}, our method reaches a new balance between rendering quality, model size, and training time while remaining memory-friendly. 
Compared with most prevailing offline baselines, our method requires significantly less host memory during training, maintains a smaller model size at inference, and achieves comparable rendering quality by quantitative metrics. In contrast to online baselines, it simultaneously attains compact model size, high rendering fidelity, and fast convergence. The contributions of this paper are three-fold as follows:

% contribution 写清楚一点，在清楚的前提下简练。
\begin{itemize}
  \item We propose a novel method that combines separate spatial and temporal hash encodings to efficiently encode both scene geometry and motion patterns. This design improves scalability and rendering quality in dynamic scenes. 
  \item  We propose a fast and highly scalable continual learning framework for dynamic scenes. It enables reconstructing long videos with low memory usage and low storage consumption. Our method also outperforms online alternatives and achieves comparable results to offline methods. 
  \item We provide a dataset of multi-view exceptionally long video sequences ($>3$ minutes) with large non-rigid motion, which is rarely found in existing datasets. 

\end{itemize}

\section{Related Work}

% "大家都做得好。我们在大家的牛逼工作基础上，又做了一点点东西。“
% "我们就是做了一点incremental的工作，然后这个工作可能会有什么什么积极影响。”
% reviewer攻击你的novelty，你就说”我们和别人不一样。“ %不要说指标，你就说和别人不一样就行。方法不一样，任务不一样都行。不一样带来了什么优势。
% 
\subsection{Fast novel view synthesis in static scenes} 

 % NeRF
%Improvements of vanilla NeRF mainly focus on the following aspects:
%acceleration, quality, and robustness. 
%Training acceleration

The acceleration of NeRFs falls into two categories: training acceleration and rendering acceleration.  
Plenoxels \cite{plenoxels} uses voxel grids to represent the scene and achieves magnitudes of acceleration over MLP backbones. 
TensoRF \cite{tensorf} decomposes the 3D scene as the product of tensors and achieves fast novel view synthesis. 
Plane-based methods \cite{kplanes,trimiprf} project  3D points onto 3 orthogonal 2D planes to ensure sparsity and achieve fast convergence. 
Instant-NGP \cite{ingp} leverages an occupancy grid and a multi-resolution hash encoding along with optimized tiny MLPs to enable fast novel view synthesis within a minute. 
The recent 3D Gaussian method \cite{3dgs} represents the scenes as a union of explicit Gaussian point clouds and achieves high fidelity, fast convergence, and real-time rendering on high-end hardware by eliminating the querying burden of spatial representations. 
% Rendering acceleration
KiloNeRF \cite{kilonerf} distills the large MLP in NeRF \cite{nerfeccv} into tiny MLPs to accelerate rendering by thousands of times. 
Hedman \etal \cite{snerg_baking} propose the \emph{baking} method (converting the MLP representation into explicit meshes) to accelerate the rendering process. 
They alpha-compose the cached features along the ray and use MLPs only once per ray to overcome the computation load of the per-point MLP querying in NeRF. 
BakedSDF \cite{bakedsdf} stores the view-dependent features into Gaussian spheres along the reconstructed surfaces, and it achieves real-time rendering on ordinary laptops. MobileNeRF\cite{mobilenerf} represents scenes into meshes and enables real-time rendering on mobile devices.
Light field methods \cite{LightFieldNetworks,mobileR2L,lightspeed} directly represent the scene as a set of rays to eliminate volume rendering and enable real-time applications on mobile devices. 
% quality 

\subsection{Representations of dynamic scenes}

%你可以说自己好，不要说别人坏。
%“我们的方法可以scale到长视频，但是之前的方法没有在这方面做过尝试。”
%“我们在往前走，不要说别人走得慢”。
% 你可以说别人的方法没有做过长视频，但是不要说别人的方法不好。
% 在related works里面应该是顺着前人的研究进一步做。
% 

There are mainly 3 variants of NeRF for dynamic scenes.  The first class is pure MLP-based representation. 
%MLP
D-NeRF \cite{Dnerf} leverages a large \emph{deformation} MLP to model the transition and the trajectories of the objects and extends NeRF to simulated dynamic scenes. 
% point out that naively 
%applying the  Fourier spatial encoding in NeRF \cite{nerfeccv} to 4D is not desirable. 
DyNeRF  \cite{Dynerf} uses long latent vectors to encode temporal information along with large MLPs to achieve high-fidelity novel view synthesis in real dynamic scenes. 
 Park \etal \cite{park2023temporal} propose an interpolation-based MLP representation for dynamic scenes and achieve significant acceleration over DyNeRF. 
The second class is tensor factorization-based representation. 
HyperReel \cite{hyperreel} proposes a network to predict the sample locations and represent dynamic scenes by key-frame-based TensoRF \cite{tensorf}  representation to ensure high fidelity and real-time rendering without any custom CUDA kernels. 
%Tensor : tensor4D, kplanes, HexPlane
%Tensor factorization 
%enables a 4D field $(x,y,z,t)$ to be learned quickly. 
K-Planes \cite{kplanes} and HexPlane \cite{HexPlane_} factorize the whole 4D field into the product of 2D planes. 
 Though differently implemented, they both achieve high-fidelity results with intermediate model size (200MB) and training time (100 minutes) on the DyNeRF dataset. 
 %Voxel: TineuVox, MixVoxels
The third class is the explicit voxel-based representations. 
TineuVox \cite{tineuvox} leverages the deformation MLP in D-NeRF and implements a 3D voxel grid with multi-scale interpolation, and it achieves high fidelity and fast convergence on the D-NeRF dataset. 
MixVoxels \cite{mixvoxels}  leverages the standard deviation across the time of all pixels to estimate a variation field to decompose the dynamic scene into static and dynamic voxels. 
Equipped with different representations for static and dynamic objects,  MixVoxels achieves high-fidelity results and super-fast convergence (15 minutes) on the DyNeRF dataset.  Song  \etal \cite{nerfplayer}   train Instant-NGP \cite{ingp} and TensoRF \cite{tensorf} offline and stream the models along the feature channels for dynamic scenes. 
4K4D \cite{4k4d} uses foreground masks to separate the dynamic objects from the static background, and leverages a point cloud-based method to achieve real-time rendering. 
Concurrent works are leveraging 4D Gaussian point clouds \cite{4Dgaussiansplatting,gs4dhust} or 3D Gaussians with deformation networks \cite{deformable3Dgaussians}  achieving real-time rendering and high fidelity simultaneously.

\subsection{Continual learning of NeRF}

Continual learning (online learning) aims to adapt to new tasks while mitigating catastrophic forgetting (resisting performance downgrade on the previous tasks) \cite{CLsurvey}.  
There are mainly 3 approaches to achieving continual learning: replay, regularization, and parameter isolation. 
The replay-based methods either store the previous data in a buffer and replay them \cite{prioritized_replay}, or use the model to generate the pseudo-ground truth of previous tasks while training on new tasks \cite{generative_replay}.  
The regularization-based methods hold the parameters close to the solution of previous tasks \cite{cl_parameter_regularization}. The parameter-isolation methods use disjoint parameters to cope with different tasks. 

Depending on the application scenarios, the continual learning methods can be divided into two categories: static and dynamic. 
On the one hand, several recent works leverage MLP \cite{MEIL-nerf} or hash grid representations \cite{clnerf,instantCLstatic} and replay to enable continual learning of NeRF in static scenes. On the other hand, for the dynamic scenes, the continual learning is mainly based on parameter isolation:  
StreamRF \cite{streamRF} uses parameter isolation of explicit voxel grids to extend continual NeRF to dynamic scenes. 
Their representation enables fast reconstruction  (75 minutes for 300 frames) and can be streamed frame by frame.  
INV \cite{invnerf} uses a shared color MLP and trains different structure MLPs frame by frame to form a continual NeRF representation. It delivers high rendering quality in dynamic scenes. 

\section{Method}
Previous online methods for dynamic scenes \cite{streamRF,invnerf} do not consider the scalability of the model: their model size scales almost linearly with the number of frames, which is of limited use in long video sequences. 
Our method combines hash representation and parameter isolation for continual learning and achieves high scalability: it not only supports continually reconstructing the dynamic scene but also scales with a significantly lower model size complexity.

\subsection{Problem definition and preliminaries} 
\label{method-preliminaries}

% continual learning: problem definition
Let $I_i$ denote the $i$-th input image observed from direction $d_i$,   $\mathbf{V}= \{\nu_i~|~\nu_i=(I_i, d_i), 1\leq i \leq N_{view}\}$ is the set of the training views, where $i, N_{view} \in \mathbb{Z^+}$. Let  $\mathbf{D_{a}}$ denote the readily accessible data at one step.  
Denoting all the time stamps as $\mathbf{T}=\{\tau_i~|~1\leq i \leq N_{time}\}, i, N_{time} \in \mathbb{Z^+}$, we define the continual learning problem as synthesizing a video sequence when the model could only have access to a limited number of frames $\mathbf{T_{a}} \subsetneq \mathbf{T} $ from all views $\mathbf{V}$  at each step,  \ie $\mathbf{D_{a}} = \{ \mathbf{V} \times \mathbf{T_{a}} \}$ where $\times$ is the Cartesian product of sets. 
To simplify the notations, let the whole sequence of frames be evenly divided into $N_{chunk}$ \emph{chunks} where a {chunk} $\mathbf{T_k}= \{\tau_i~|~ \lfloor \frac{i}{T_{chunk}} \rfloor  = k, k = 0, 1, \dots N_{chunk}-1\}$ is a sequential subset of length $T_{chunk}$.  The continual learning problem is thus defined as learning a representation where only the frames in the current chunk are available at each optimization step, \ie. $\mathbf{T_a}=\mathbf{T_k}$.  
Specifically, $T_{chunk}=1$ denotes training the model online in a frame-by-frame manner, and $T_{chunk}=10$ denotes training the model online with a buffer of 10 frames. If the video contains 300 frames in total, the  $T_{chunk}=300$ denotes training the entire 300-frame sequence offline. 
The definition is close to real-world applications since caching all training frames in memory or storing them on the disk is limited in scalability for long videos. 

Inspired by the previous NeRF variants \cite{Dnerf, Dynerf,kplanes,HexPlane_,mixvoxels,ingp,merf} for dynamic novel view synthesis and continual learning methods \cite{clnerf,instantCLstatic,MEIL-nerf,streamRF,nerfplayer,invnerf} in the static scenes, we focus on the dynamic novel view synthesis problem in a multi-view setting. We solve the problem using continual learning to alleviate the huge memory consumption and bandwidth costs. 
Unlike the existing replay-based continual learning methods for static scenarios \cite{clnerf,instantCLstatic}, we solve the continual learning problem of dynamic novel view synthesis via parameter isolation. This is because the current offline state-of-the-art methods by tensor factorization \cite{HexPlane_, kplanes} or explicit voxels \cite{mixvoxels} are regressions of the optical attributes in 3D space within a limited time instead of the object trajectory and inherent dynamics. 
Therefore, the replay method is of limited use in our defined scenario: Since new transient processes are constantly arriving, the model of a limited size is not capable of reconstructing unlimited transient effects. Thus, the replay-based method suffers from catastrophic forgetting. 
The design of our model follows the parameter isolation approach: Let the time stamps be divided into $N_{chunk}$ chunks, and the model consists of $N_{chunk}$ \emph{branches} to represent different \emph{chunks} of the dynamic scene accordingly. 

Following previous methods \cite{Dnerf, Dynerf,kplanes,HexPlane_}, we consider the dynamic scene as a set of 3D points $\mathbf{x}=(x,y,z)$ with optical attributes $(\sigma, \mathbf{c})$,  where $\sigma$ is the volume density of the point and $\mathbf{c}$ is the color of the point. To reconstruct the scene, we train our model along camera rays which are defined as:

%---------------------------Equation-----------------------
\begin{equation}
  \mathbf{r}(u)=\mathbf{o}+u \mathbf{d} ,
   \label{cameraray}
\end{equation}
%-----------------------------------------------------------
where $\mathbf{o}$ is the optical center of the camera, $\mathbf{d}$ is the direction from the camera to the scenes. 
For each point along the camera ray with coordinates $\mathbf{x}=(x,y,z)$ observed at direction $\mathbf{d}=(\phi,\psi)$,  a field function $F$ is queried to obtain the optical attributes $(\sigma, \mathbf{c})$. Conditioned on time $t$, the entire model is defined as:
%---------------------------Equation-----------------------
\begin{equation}
   (\sigma, \mathbf{c})=F_{\Theta}(x,y,z, t ,\phi,\psi) ,
   \label{nerfquery}
\end{equation}
%-----------------------------------------------------------
where the function $F$ denotes the dynamic NeRF model and $\Theta$ denotes trainable model parameters.

\subsection{Representation of short dynamic intervals}
To reconstruct the individual chunks of the dynamic scenes, we propose a representation of short dynamic intervals, which serve as the structures of the different branches of our model and handle the dynamic scenes in one chunk efficiently. 
As shown in \cite{rerf}, naive 4D hash grids for dynamic scenes are short in reconstruction quality. 
To simultaneously achieve fast reconstruction and compact model size, we use hash encodings to represent temporal and spatial features separately, then we concatenate them to alleviate feature interference. 
The spatial hash encodings can be implemented as hash tables of 3D voxels \cite{ingp}, orthogonal 2D planes \cite{kplanes,trimiprf}, or both \cite{merf}. 
The time feature is encoded by short hash tables in a per-frame manner. 
Following previous state-of-the-art methods  \cite{nerfeccv,ingp,tineuvox}, we apply a two-fold field function for geometry and color individually.
The architecture of the field function is thus defined as:
%---------------------------Equation-----------------------
\begin{equation}
  (\sigma,\mathbf{H})=f_{\theta_1}( \Psi(\mathbf{x}) \oplus   \Gamma(t) ) ,
  \label{our_dnerf_sigma}
\end{equation}
\begin{equation}
    \mathbf{c}=f_{\theta_2}(\mathbf{H},\phi,\psi) , 
    \label{nerfcolor}
\end{equation}
%-----------------------------------------------------------
where $\mathbf{x}=(x,y,z)$ is the spatial coordinates, $\Psi (\mathbf{x})$ denotes configurable spatial hash encodings,
 $f_{{\theta_1}}$ is a tiny MLP with trainable parameter ${\theta_1}$,
 $\Gamma(t)$ is the hash encoding for time axis alone, and $\oplus$ is the  concatenation of vectors.  
 The MLP $f_{{\theta_1}}$ in \cref{our_dnerf_sigma} fuses features across the spatial domain and time domain and obtains the volume density $\sigma$.
View-dependent colors $\mathbf{c}$ are regressed by another tiny MLP with parameter $\theta_2$ in \cref{nerfcolor}. 
The spatial hash encoders $\Psi(x)$ follow a multi-resolution concatenated format, defined as: 
%---------------------------Equation-----------------------
\begin{equation}
  \Psi(\mathbf{x})= \bigoplus_{m=1}^{L} ~ P_m(x,y,z) , 
  \label{hash encoder config}
\end{equation}
%-------------------------------------------------
where the $P_m(x,y,z) \in \mathbf{R}^{F}$ denotes the tri-linearly interpolated feature from the nearby 8 grid vertices \wrt 3D coordinates $(x,y,z)$ at the $m$-th resolution, $L$ is the number of the resolutions, and $F$ is the dimensions of features  $P_m(x,y,z)$, which are similar to those in Instant-NGP \cite{ingp}. 
The voxel-based hash encoding structure is flexible and could be replaced by other spatial hash encodings, which will be discussed in the experiments.

\label{method: representation of short dynamic intervals}

\subsection{Continual dynamic neural graphics primitives (CD-NGP)} 
\label{method: cd-ngp}
As defined in \cref{method-preliminaries}, we split the model into $N_{chunk}$ individual branches to represent the $N_{chunk}$ chunks of the dynamic scene. Consequently, the spatial features encoded by different model branches could be fused to reduce the number of parameters and improve scalability.
This approach, defined as continual dynamic neural graphics primitives (CD-NGP), is designed to handle the dynamic nature of scenes in the online setting. 
Equipped with the hash encoders and MLP regressors defined in \cref{method: representation of short dynamic intervals}, the $k$-th branch of the model contains spatial hash encoder $\Psi_k(\mathbf{x})$, temporal hash encoder $\Gamma_{k}(t)$, and MLPs $f_{\theta_{1k}},f_{\theta_{2k}}$.  Please note the dimensions of the features are independent of the sizes of the hash tables: a larger hash table alleviates hash collision at the cost of more storage consumption, but the features encoded by different sizes of hash tables share the same size, i.e. $\Psi_k(x) \in \mathbf{R}^{LF}$ holds for all $k$. To alleviate hash collision and reduce model size simultaneously, we apply an asymmetric hash table design: the hash table in the first branch (\emph{base branch}) is set as $2^{P_1}$, and the hash tables in the subsequent branches  (\emph{auxiliary branches}) are set as $2^{P_2}$, where $P_1$ is much larger than $P_2$. The base branch is designed to reconstruct the scenes with initial movements and static objects, and the auxiliary branches are designed to compensate for residuals. The re-use of the features in the base branch ensures high scalability and alleviates the catastrophic forgetting problem.  When ($P_1,P_2)=(19,14)$, the hash table in the second branch consumes only $1/32$ of the parameters in the first branch. 
We substitute the $\Psi(\mathbf{x})$ in  \cref{our_dnerf_sigma} with  the fused feature to obtain the volume density $\sigma$ at the $k$-th model branch under the continual learning setting, defined as:
%---------------------------Equation-----------------------
\begin{equation}
  (\sigma,\mathbf{H})=f_{\theta_{1k}}( (\Psi_0(\mathbf{x}) + \Psi_k(\mathbf{x})) \oplus   \Gamma_{k}(t) ),
  \label{continual_dnerf_sigma}
\end{equation}
%-------------------------------------------------
where $\Psi_0(\mathbf{x})$ denotes the spatial encoder of the base branch, 
$\Psi_k(\mathbf{x})$ denotes the spatial encoder of the current branch. The view-dependent colors are regressed as in \cref{nerfcolor} with the color MLP $f_{\theta_{2k}}$. 

We further illustrate the entire process of our method in \cref{alg:continual-training}, where $\eta_{init}$ and $\eta_{aux}$ denote the numbers of iterations of the first branch and the auxiliary branches, respectively. $\Psi_i, \Gamma_i, \theta_1^{(i)}, \theta_2^{(i)}$ denote the model parameters in the branches. The model is trained in a continual learning manner: the first branch is trained from scratch to reconstruct the initial chunk of the dynamic scene, while the MLP decoders of the auxiliary branches are initialized from the previous branch to reconstruct subsequent chunks, facilitating faster convergence. To cope with large motions and complex transient effects in long time intervals, we assign $T_{episode}$ chunks as an \emph{episode}. The MLP regressors are re-trained from initialization in the starting branch of each episode to adapt to recent changes.

\renewcommand{\algorithmicrequire}{\textbf{Input:}}
\renewcommand{\algorithmicensure}{\textbf{Output:}}
  %-----------------algorithm------------------
  \begin{algorithm}[!ht]
    \caption{Continual learning pipeline}
    \label{alg:continual-training}
    \begin{algorithmic}[1]
        \REQUIRE  $N_{chunk}, T_{chunk}, T_{episode}, \eta_{init}, \eta_{aux}, \Psi_i, \Gamma_i, \theta_1^{(i)}, \theta_2^{(i)},  i \in \mathbb{N}$
        \ENSURE The whole representation for dynamic scenes. 
        \STATE  $ k \leftarrow 0$,  \hfill $\%$ initialize
        \WHILE {$k$<$N_{chunk}$}
          \STATE  $\mathbf{T_a} \leftarrow \mathbf{T_k}$   \hfill $\%$ assign the current chunk
          \STATE  $\mathbf{D_a} \leftarrow \mathbf{V} \times \mathbf{T_a}$   \hfill $\%$ assign accessible data
          \IF {$k ~ \mathbf{mod} ~ T_{episode} ~ = 0$ }
          \STATE $\eta \leftarrow \eta_{init}$ \hfill $\%$  The first branch in the episode.
        \ELSE 
          \STATE $\eta \leftarrow \eta_{aux}$  \hfill $\%$  The other branches in the episode.
          \STATE $\theta_1^{(k)}, \theta_2^{(k)}, \Gamma_k  \leftarrow \theta_1^{(k-1)}, \theta_2^{(k-1)}, \Gamma_{k-1} $   \hfill $\%$ Initialize  with the previous branch.
        \ENDIF
        \STATE  Train the $k$-th model branch   $(\Psi_k,\Gamma_k,\theta_1^{(k)}, \theta_2^{(k)})$ for $\eta$ steps.
        \STATE $k \leftarrow k + 1    $                           \hfill $\%$ update the chunk index $k$
        \ENDWHILE 
    \end{algorithmic}
  \end{algorithm}

The scalability of our method derives from the redundancy of dynamic scenes and the flexibility of hash table sizes. 
As pointed out by \cite{mixvoxels}, most points ($> 90\%$ in the DyNeRF dataset by their observation) in the dynamic scenes are static. We thus leverage the shared static points across time and represent the first chunk with a large hash table and the subsequent chunks with small hash tables to ensure scalability. 
Since the spatial hash tables occupy the most parameters in the model, and most chunks are encoded by the subsequent branches, our asymmetric hash table design enables continual learning of dynamic scenes and high scalability in streaming scenarios simultaneously. The total bandwidth consumption of our method is $ \mathcal{O} (N_{chunk} \cdot 2^{P_2}LF + 2^{P_1}LF) $, 
while the previous method \cite{nerfplayer} requires offline training and consumes $ \mathcal{O}(\mathbf{K}\cdot2^{P_1}LF) $ bandwidth,  where $\mathbf{K}$ ranging from $0.5$ to $16$ is their bandwidth budget controlling the bitrates in streaming scenarios. 
If the bandwidth is extremely limited, the hash tables in our model could be arranged as a combination of 3D representations and 2D representations like \cite{merf} to reduce the bandwidth cost additionally.

\subsection{Rendering and optimization}

Given the sampled optical attributes on the rays in \cref{cameraray}, we adopt volume rendering to get the 2D pixels, defined as:
%---------------------------Equation-----------------------
\begin{align}
    \hat{C}(r) &= \sum_{i=1}^N T_i\alpha_i \mathbf{c_i}= \sum_{i=1}^N T_i (1-e^{-\sigma_i\delta_i}) \mathbf{c_i} ,
\label{vren_discrete}
\end{align}
%-------------------------------------------------
where $T_i=\exp(-\sum_{j=1}^{i-1} \sigma_j\delta_j)$ denotes the transmittance, $\delta_i$ is the interval between neighboring samples, $N$ is the number of samples along the ray, $\mathbf{c_i}$ is the output of \cref{nerfcolor}. 
The model is optimized by L2 photometric loss with three regularization terms, defined as: 
%---------------------------Equation-----------------------
\begin{equation}
    L = \frac{1}{|\mathbf{R}|} \sum_{r \in \mathbf{R}} \|C(r) - \hat{C}(r)\|_2^2 + \lambda_d L_d + \lambda_o L_o + \lambda_r L_r ,
    \label{eq:optim}
\end{equation}
%--------------------------------------------------
where $\mathbf{R}$ denotes the set of the camera rays. $L_d$ is the distortion loss  in  Mip-NeRF 360  \cite{mipnerf360}, defined as:
\begin{align}L_{d} &= \sum_{i,j} w_i w_j  \left\lvert \frac{s_i + s_{i+1}}{2} - \frac{s_j + s_{j+1}}{2} \right\rvert \notag  \\
&+ \frac{1}{3} \sum_i w_i^2 (s_{i+1} - s_i) , 
\end{align}
\noindent where $s_i$  denotes the normalized ray distance along the  ray, $w_i=T_i \alpha_i$ is the weight of the $i$-th sample. 

 $L_o$ in \cref{eq:optim}  is the entropy on the opacity  \cite{ngppl}, defined as:  
%---------------------------Equation-----------------------
\begin{equation}\label{opac}%\label{opac_loss}
  L_{o} = -o\log(o), \quad o = \sum_{j=1}^{N}T_j\alpha_j ,
\end{equation}
%--------------------------------------------------
$L_r$ in \cref{eq:optim} is the L1 regularization on the  spatial features in the current branch as: 
%---------------------------Equation-----------------------
\begin{equation}
  L_{r} =  ||\Psi_k(\mathbf{x})||_1 . 
  \label{eq:spatial_regularization}
\end{equation} 
We set $\lambda_r$  as 0.001 to encourage using the features encoded by the base branch. 
Based on \cite{ngppl}, we set a large value of  $\lambda_d, \lambda_o$ as 0.005 to alleviate foggy effects. 
% 更换为tabular环境以后，必须手动设定列之间间距才能接近subfigure的vfill hfill的效果
\begin{figure*}[!ht]
  \begingroup
  \renewcommand{\arraystretch}{0.8}  % 缩短行间距
  \setlength{\tabcolsep}{1pt}        % 缩短列间距
  \begin{tabular}{cccc} % 
  \includegraphics[width=0.242\textwidth]{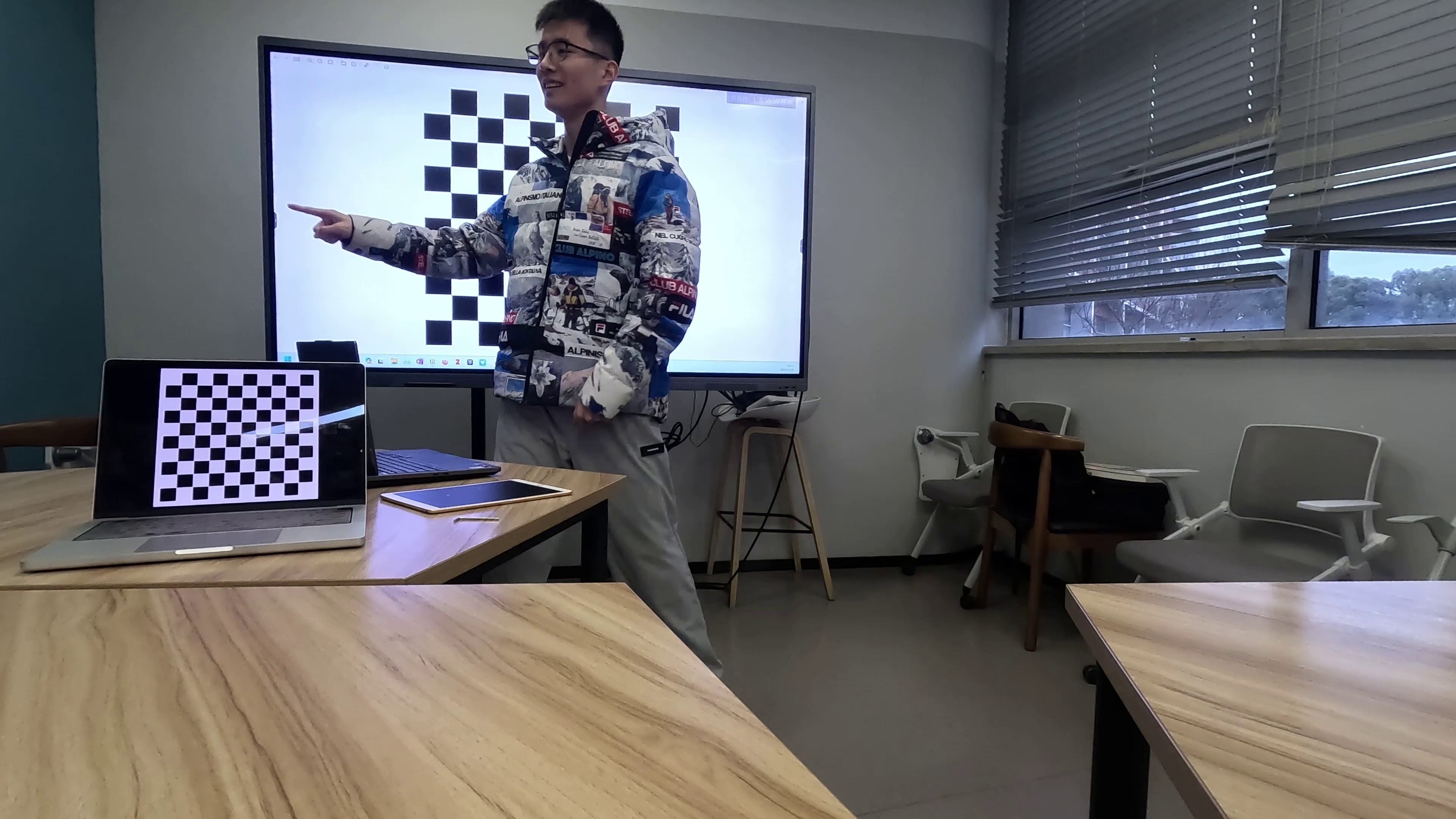}&
  \includegraphics[width=0.242\textwidth]{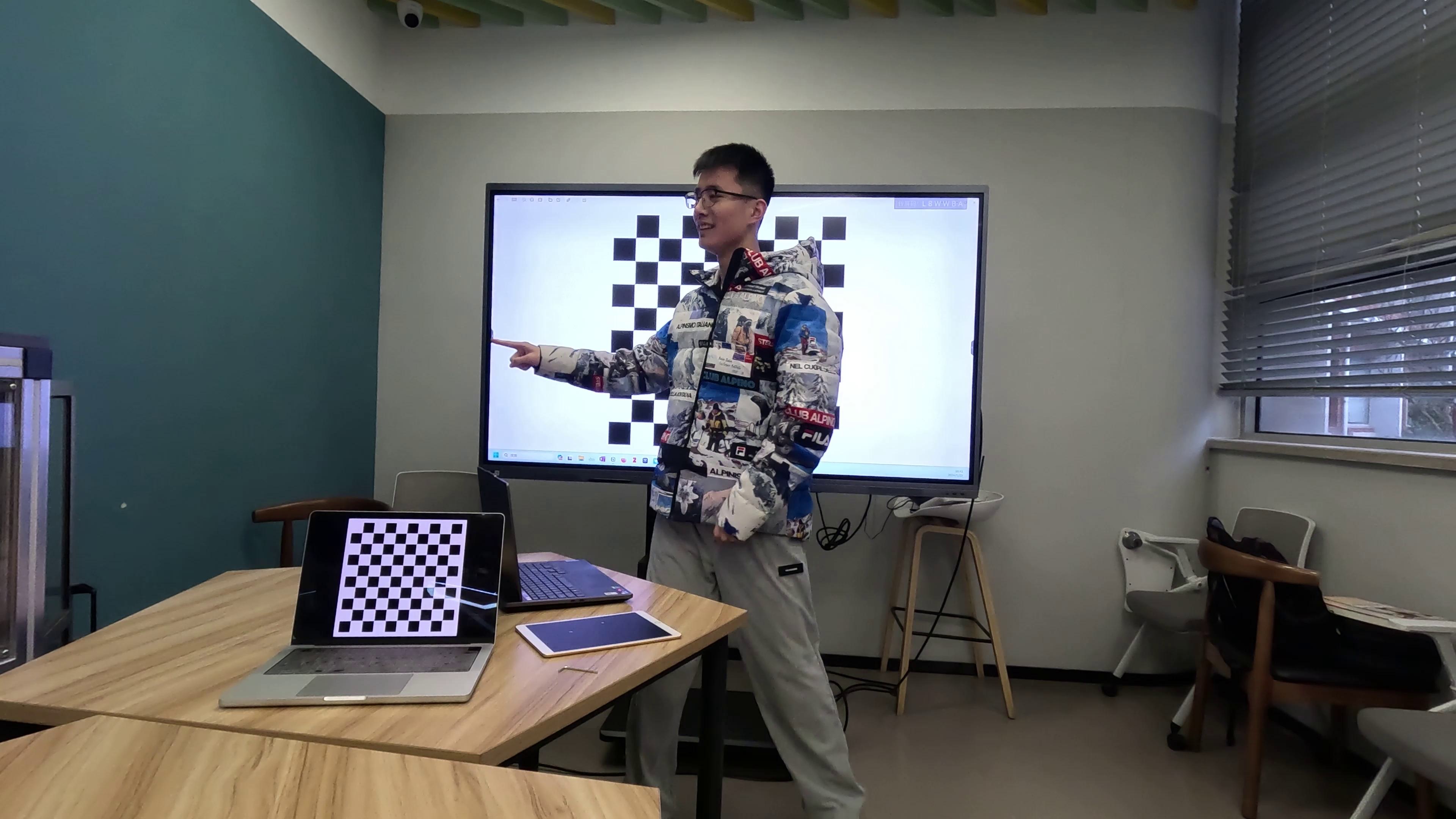}&
  \includegraphics[width=0.242\textwidth]{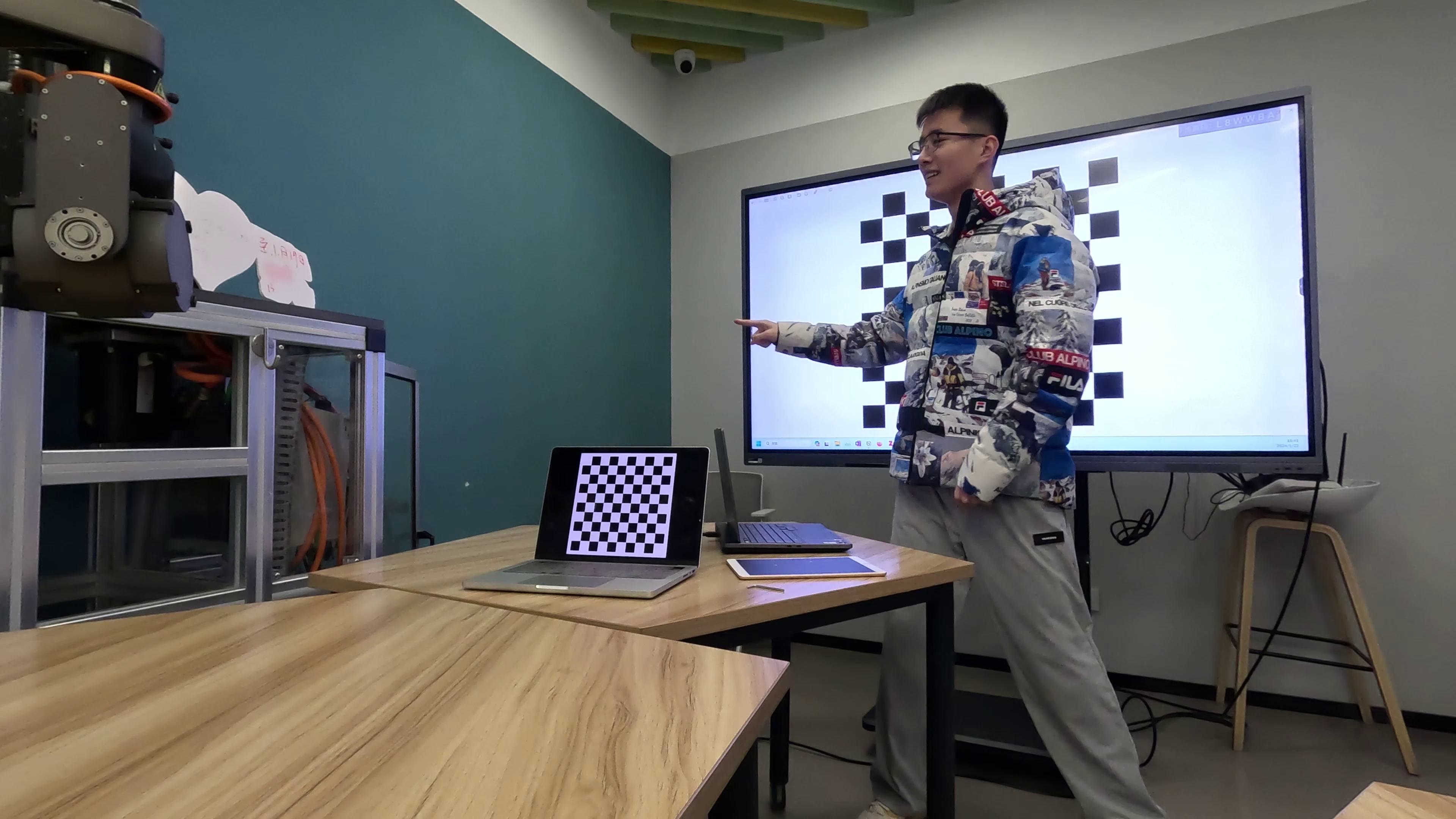}&
  \includegraphics[width=0.242\textwidth]{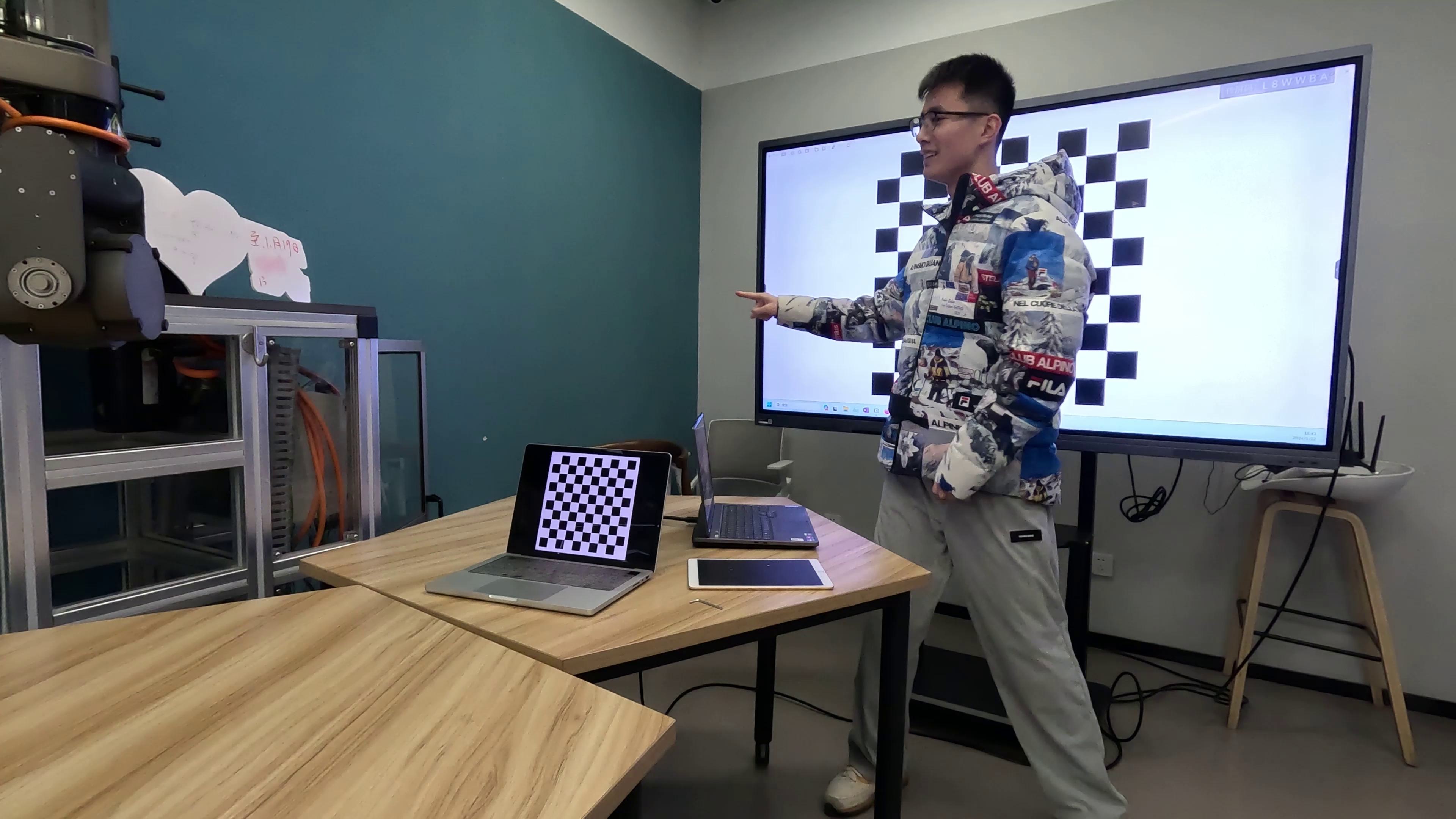}\\
  \includegraphics[width=0.242\textwidth]{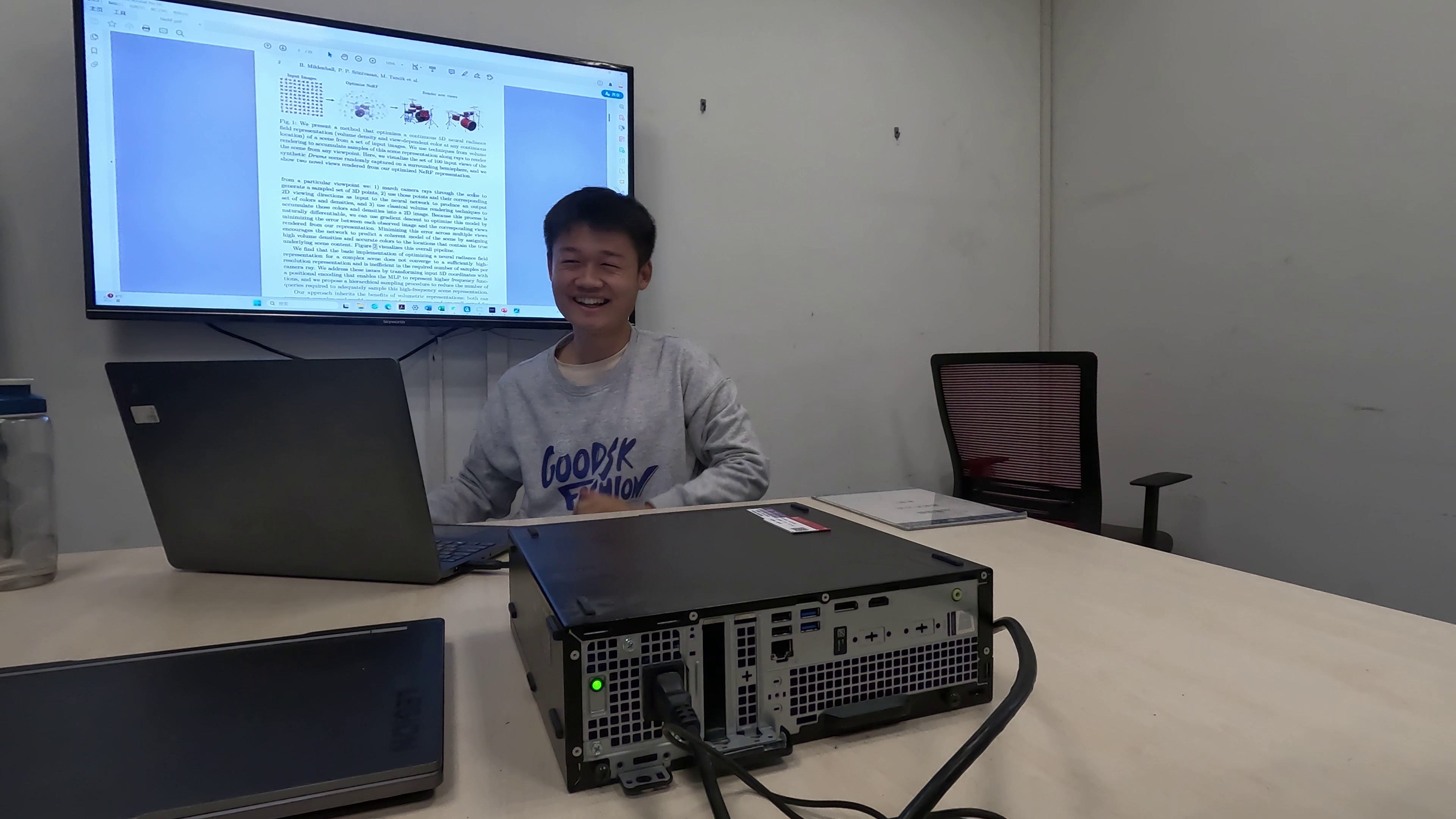}&
  \includegraphics[width=0.242\textwidth]{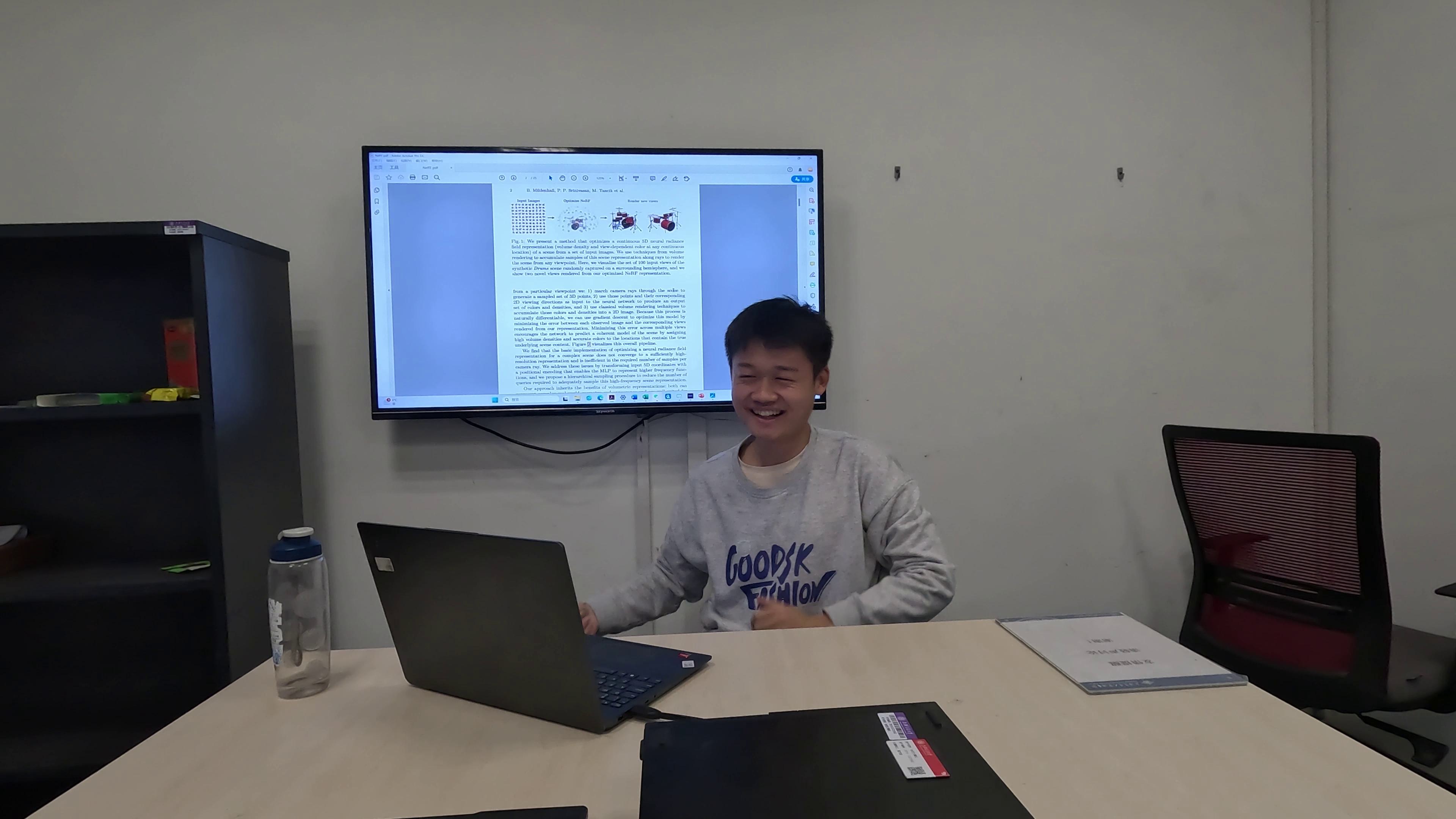}&
  \includegraphics[width=0.242\textwidth]{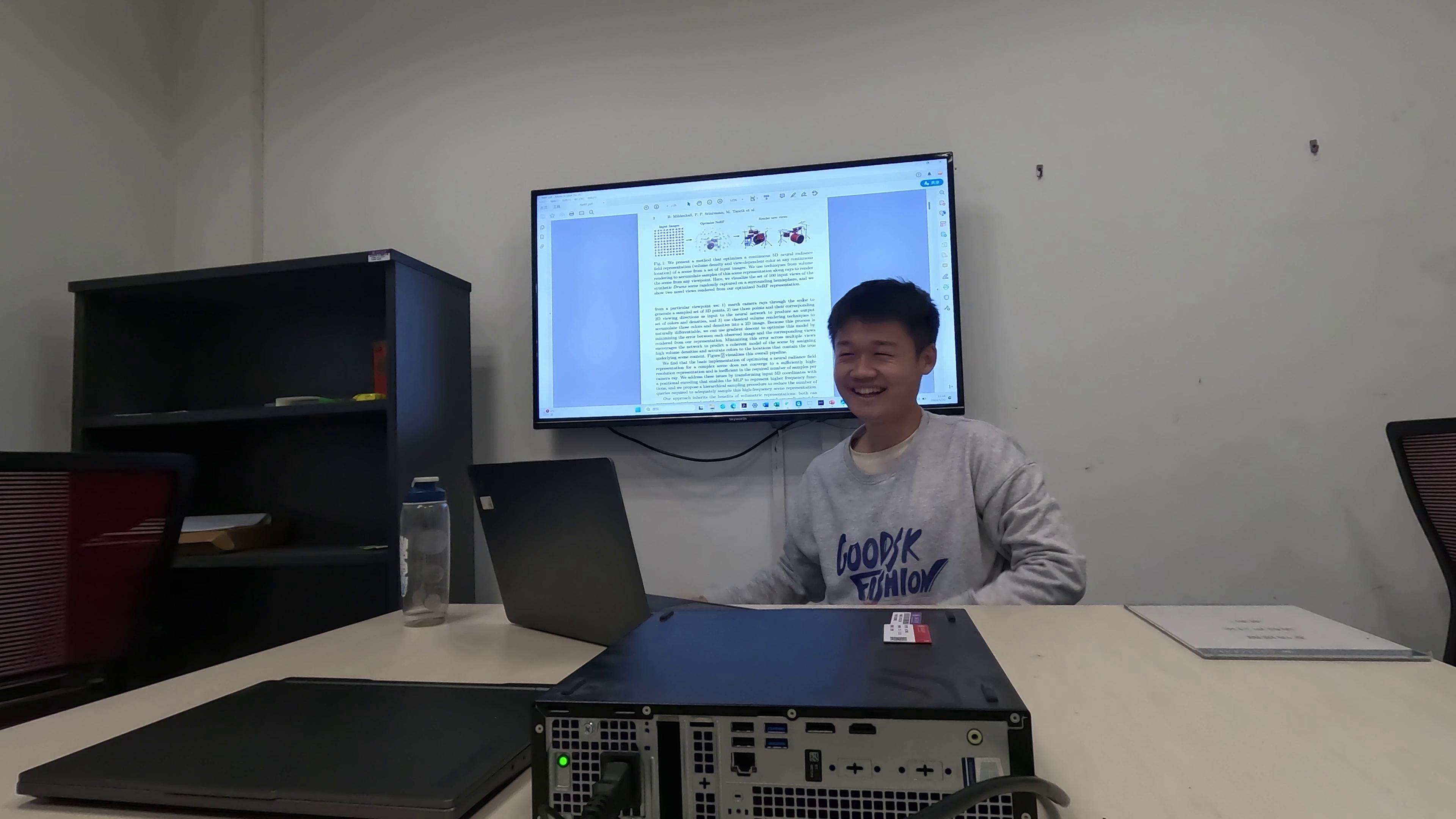}&
  \includegraphics[width=0.242\textwidth]{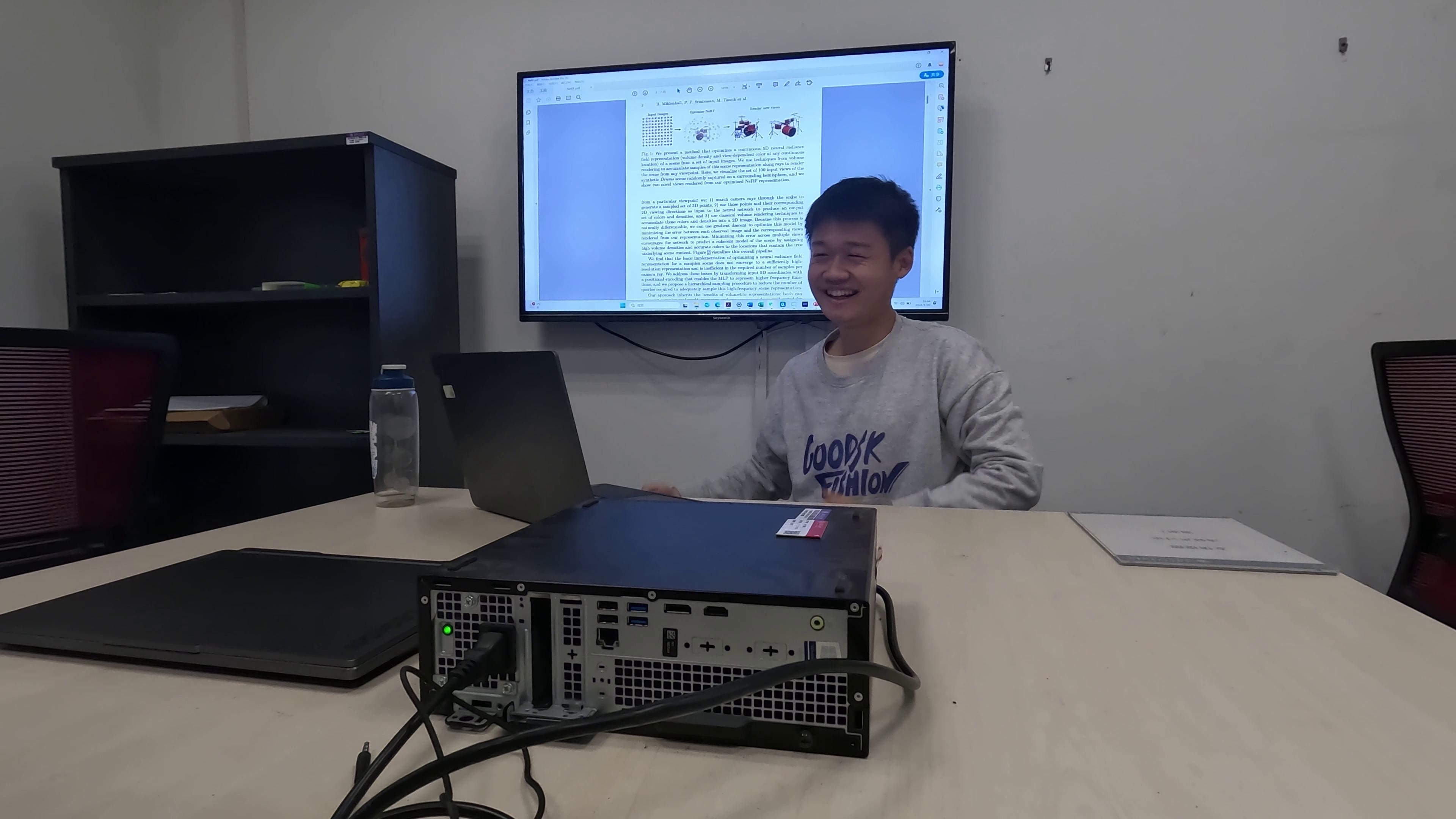}\\
  \includegraphics[width=0.242\textwidth]{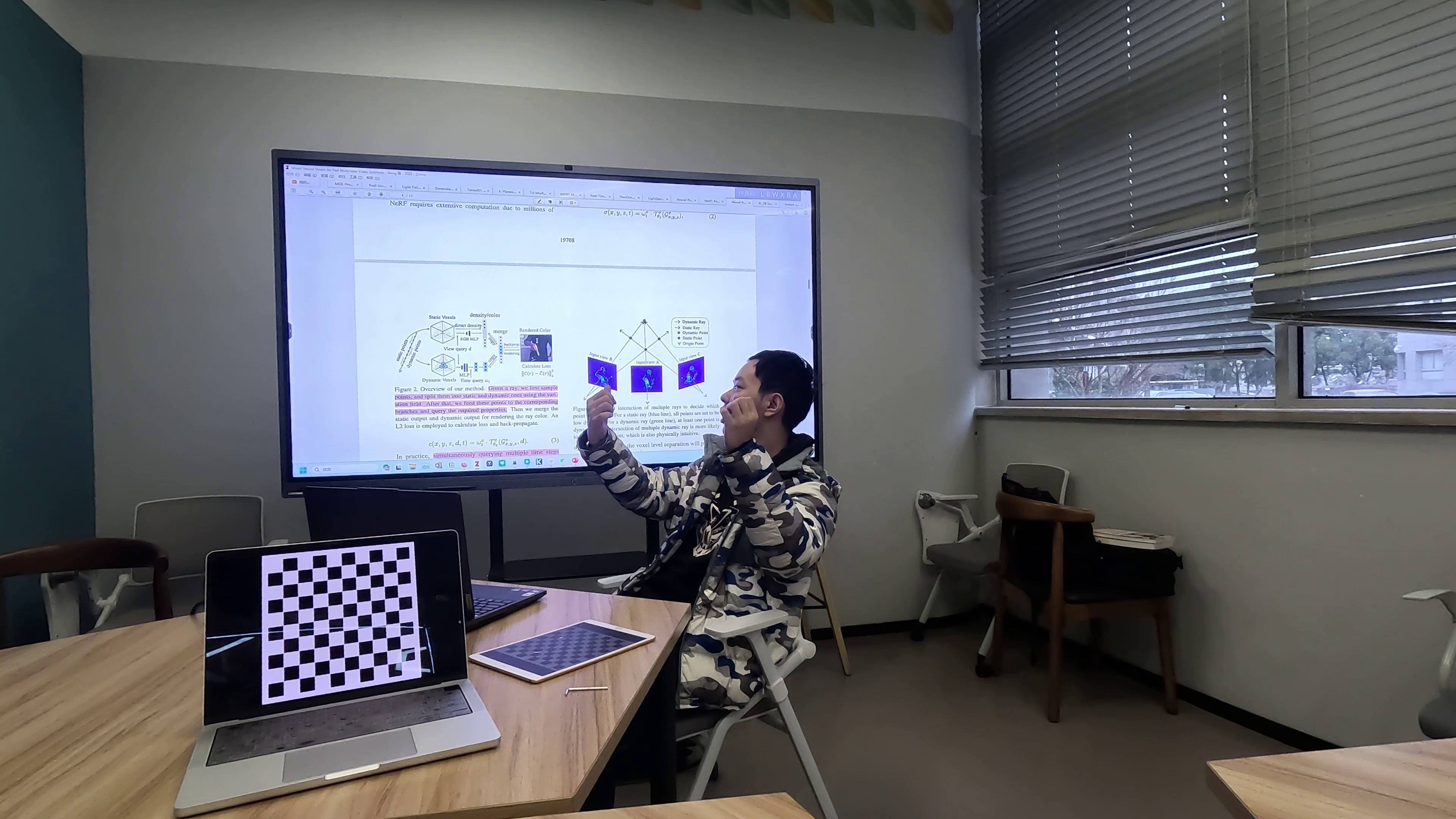}&
  \includegraphics[width=0.242\textwidth]{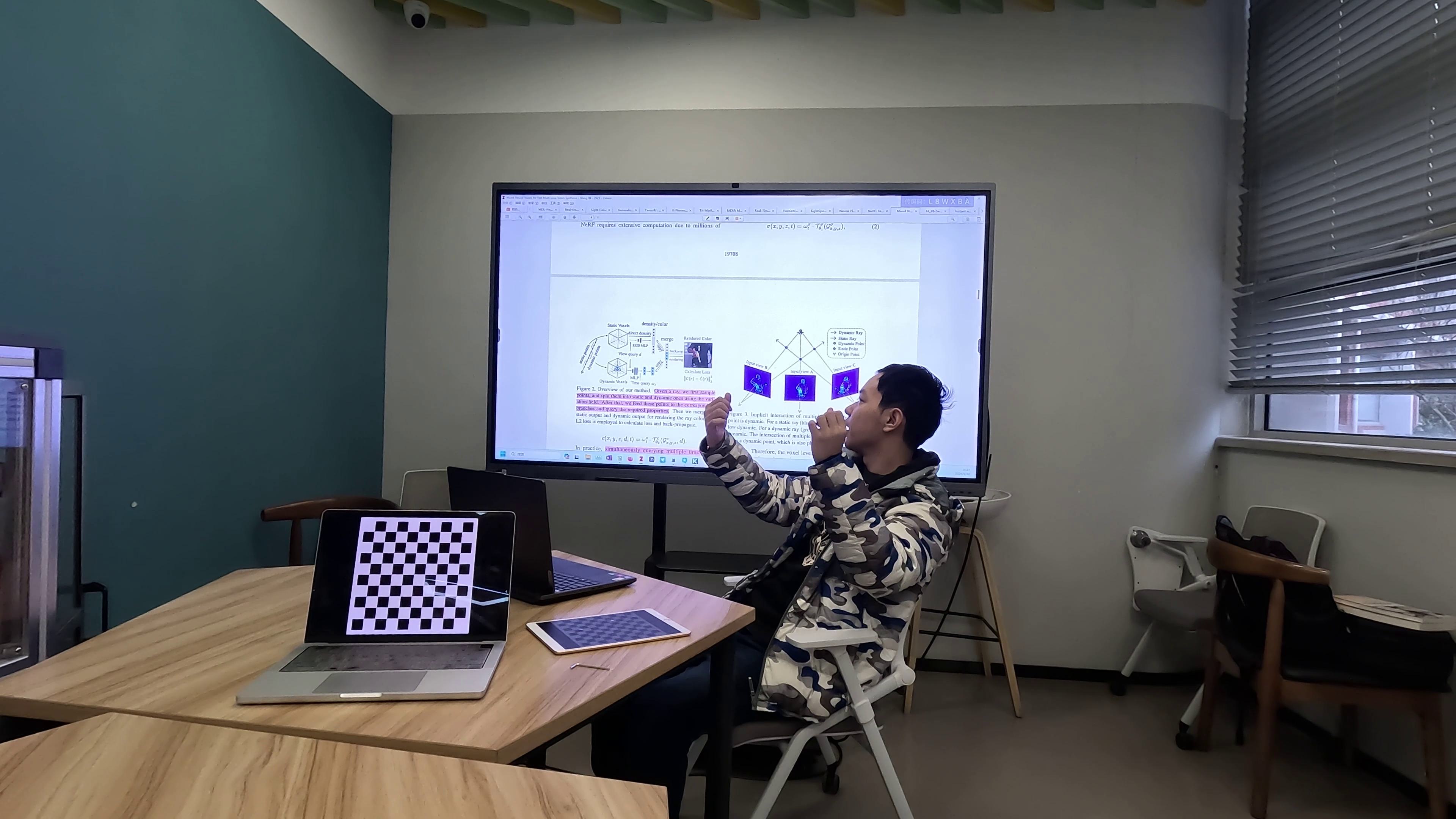}&
  \includegraphics[width=0.242\textwidth]{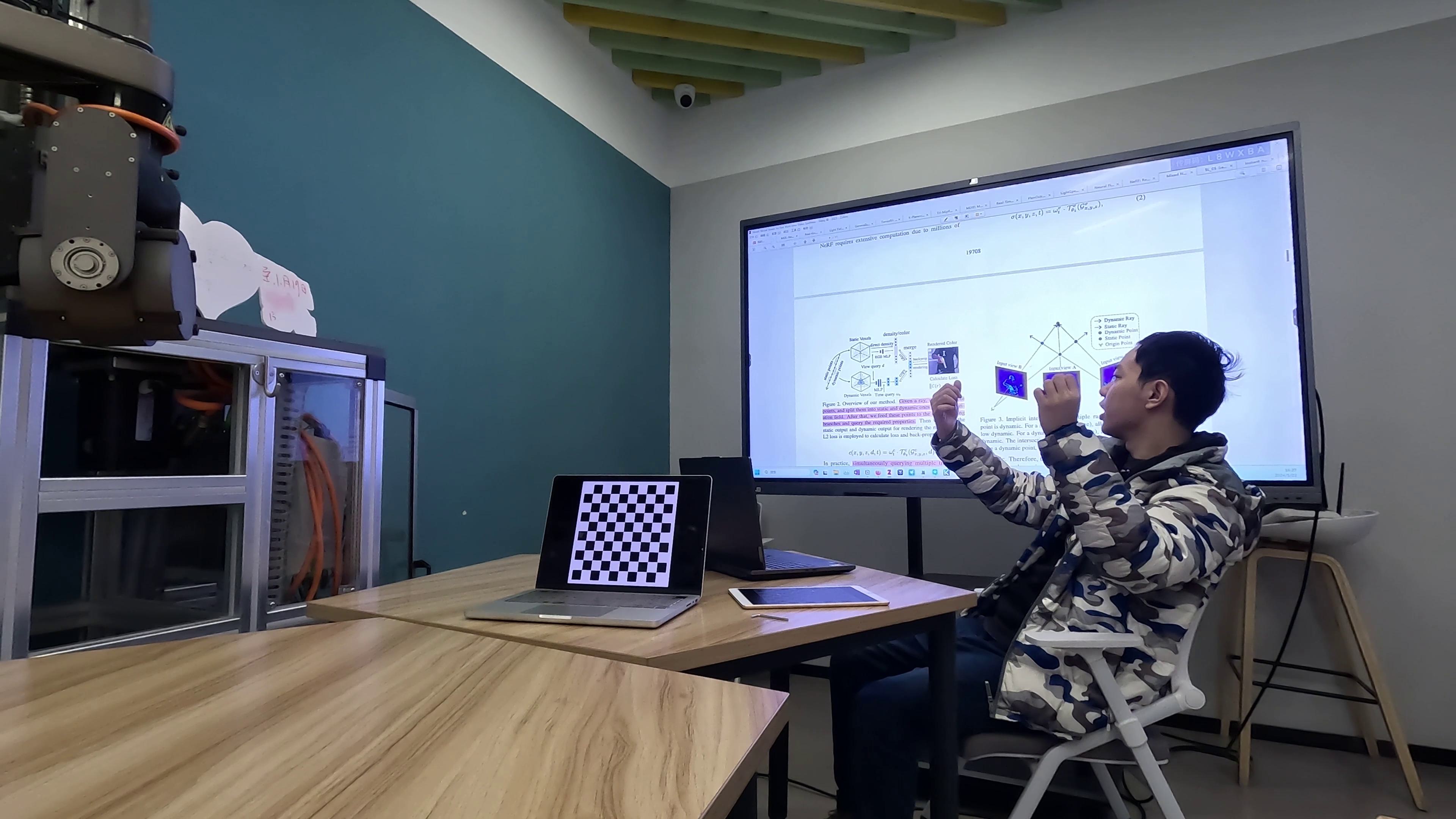}&
  \includegraphics[width=0.242\textwidth]{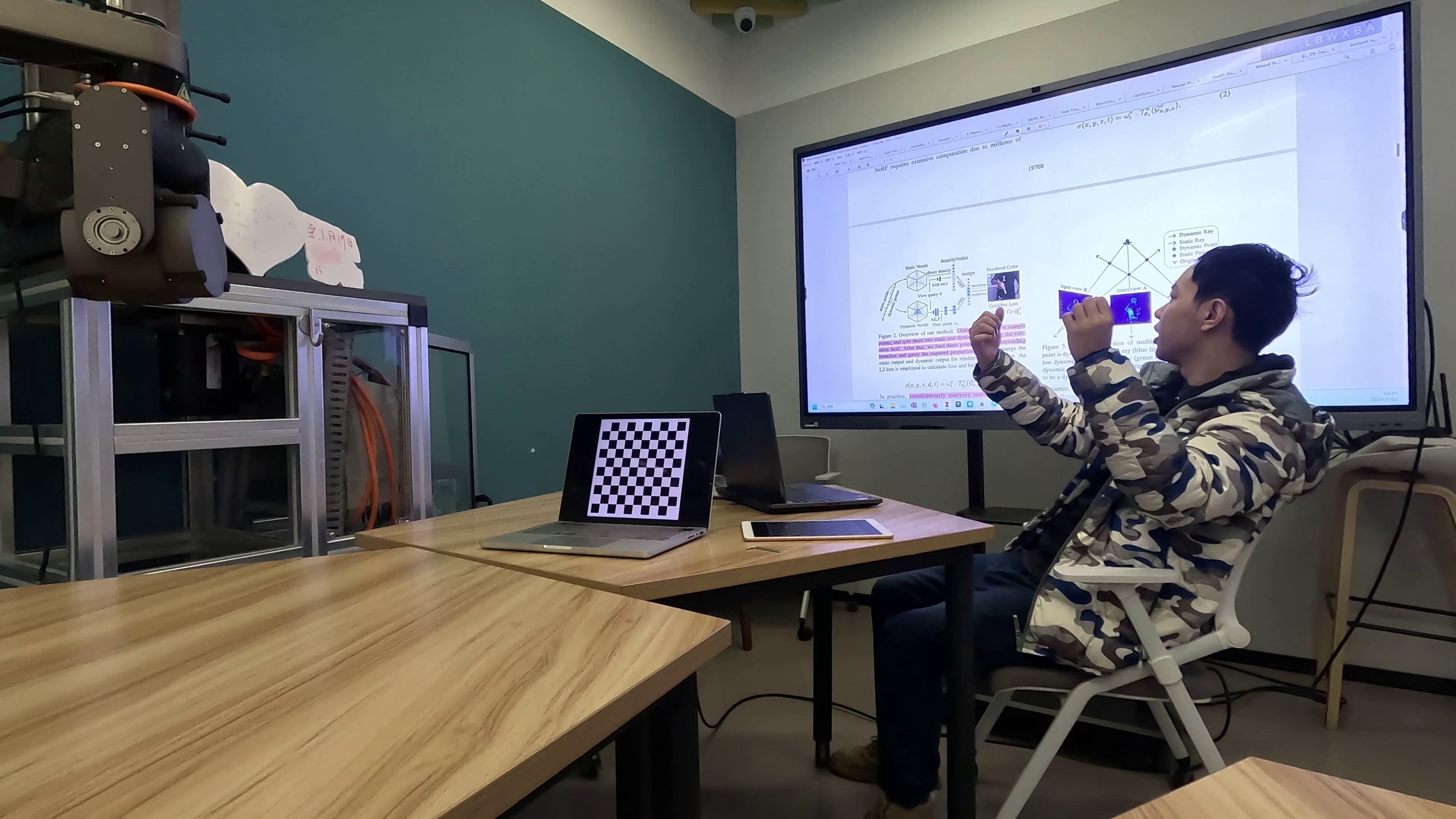}\\
\end{tabular}
\endgroup
  \caption{Example frames sampled from the proposed long multi-view video dataset.}
  \label{fig:our dataset gt demon}
\end{figure*}

\section{Experiments}
\subsection{Datasets}

The DyNeRF dataset \cite{Dynerf} is the most prevalent benchmark for dynamic view synthesis, providing high-quality multi-camera videos on which current state-of-the-art methods are primarily evaluated. This dataset includes six unique cooking scenes, each featuring varying illumination, food topological changes, and transient effects such as flames. It consists of five videos, each containing 300 frames, and one extended video with 1200 frames, captured at a resolution of $2704\times2028$ with up to 21 camera views at 30 frames per second (FPS). Following the approach of HexPlane \cite{HexPlane_}, we excluded the unsynchronized scene \emph{coffee martini} and conducted experiments on the remaining scenes, using the first 300 frames of the 1200-frame video \emph{flame salmon} for evaluation. 

The Meetroom dataset \cite{streamRF} is another dataset for dynamic NVS that  recorded by 13 synchronized cameras with  $1280 \times 720$ resolution at 30 FPS. Each of the 3 scenes contains 300 frames. 

Although existing datasets have been widely used for dynamic NVS, their limited lengths (300 frames) prevent the evaluation of the scalability of the methods on long video sequences. 
To further validate the effectiveness of our method on long video sequences, we introduce a new dataset comprising 5 dynamic scenes at a resolution of  $3840 \times 2160$ as shown in \cref{fig:our dataset gt demon}. 
Similar to those in the DyNeRF dataset \cite{Dynerf}, these scenes are captured by 20 GoPro cameras operating at 30 FPS. The cameras are arranged in two rows and placed in an arc facing the scenes. The camera in the center of the upper row is designated for validation. 
We synchronize the multi-view videos and incorporate chessboard images within the scenes to facilitate camera calibration and pose estimation. 
Our dataset features more challenging dynamics, including significant non-rigid human motions and transient effects like screen slides, compared to the DyNeRF dataset. To comprehensively assess the scalability of our method, we provide 5 scenes with a duration of 6000 frames. 
We hope this dataset will serve as a valuable benchmark for future research in dynamic novel view synthesis, especially in long video sequences. 
We have carefully reviewed the videos to ensure they do not contain any controversial or offensive content before using them for research purposes. While the faces in the videos may contain identifiable information, this information will be used solely for research and will not be disclosed to any third party. All videos were collected with the explicit consent of all participants.

\subsection{Implementation details}
We adopt the hash-encoding and tiny MLP implementation of tiny-cuda-nn \cite{tcnn} by Instant-NGP \cite{ingp}, and customize the CUDA-based training and rendering pipelines for the dynamic scenes based on \cite{ngppl}. The data preparation process is based on HexPlane \cite{HexPlane_} and LLFF \cite{llff}. Please see the supplementary material for more detailed explanations and analysis.  
All the metrics are measured on a single RTX 3090 GPU if not otherwise specified. 
Similar to the previous work  \cite{HexPlane_,Dynerf},  we select the view in the center with index $0$ for evaluation on all datasets. Following most of the state-of-the-art methods \cite{Dynerf,streamRF,kplanes,mixvoxels,nerfplayer}, we train and evaluate the model at resolution $1352 \times 1014$ on the DyNeRF dataset. We follow \cite{streamRF} and use  $1280 \times 720$ resolution on the Meetroom dataset. We use $1280 \times 720$ resolution for our dataset to fit into the host memory. 
 We use the importance sampling approach described in \cite{Dynerf}  at a batch size of 1024 rays.
The model parameters are optimized by Adam \cite{adamoptim}, with hyper parameters $\epsilon=10^{-15}, \beta=(0.9,0.96)$ for stability. 
 The learning rate is set as 0.001 and scheduled by a cosine function. 
 We set the hash table sizes as $2^{P_1}=2^{19}$ in the first branch and  $2^{P_2}=2^{14}$ in the subsequent branches. The structural parameters $(L,F)$ in the spatial hash encoders are set as $(12,2)$ if not otherwise specified. 
 At the base branch, we initialize the models and train for $\eta_{init}=18k$ iterations, and for each of the subsequent branches we train for $\eta_{aux}=3k$ iterations using the previous branch for initialization for the DyNeRF dataset and our dataset. We use $\eta_{init}=33k$ and $\eta_{aux}=3k$ for the Meetroom dataset. 
  We use LeakyReLU activation in the MLP regressors. 
Strategies for updating the occupancy grids are identical with \cite{ingp}. 
   To save storage, we use only one shared occupancy grid for all the branches. 
 We report the model sizes with the occupancy grid under {bfloat16} precision and parameters in {float32} precision.  The episode length $T_{episode}$ is set as $30$ only for videos longer than 300 frames in our dataset.  
We set the dimensions of the latent feature $\mathbf{H}$ in \cref{our_dnerf_sigma} as 48. We use MLPs with two hidden layers and the dimensions of the hidden layers in \cref{our_dnerf_sigma,nerfcolor} are set as 128, and 64 respectively. We use MLPs with identical structures for different branches. We set the resolutions of the hash encoders of the CD-NGP within $(16, 2048)$ times the scale of the scene.

\subsection{Results and scalability analysis}
  % 对比 host memory, time, model size.
  
  % 和 HexPlane-continual 的对比
  % 和 现有离线baseline的对比
  % 和 整个field的composition进行对比 （ 两种composition方法）
  % 自己的方法如果以全部离线的方式学习进行对比
  
  % 定义一个command，画箭头。
  \newcommand{\uaa}{$\uparrow$}
  \newcommand{\daa}{$\downarrow$}

  %--------------------------------------------- Table ------------------------------------
  \begin{table*}[!htbp]
    \centering
    \caption{Comparisons between our CD-NGP and state-of-the-art methods on the DyNeRF dataset. Memory denotes the maximum host memory required during training. $\dagger$ denotes the methods evaluated only on the \emph{flame salmon} scene. 
     \emph{-} denotes the metrics not reported in the original publications. }
     \begin{tabular}{l|ccllllllr}  
      \toprule
    Method    & Online & Offline   & $T_{chunk}$     & PSNR\uaa   & DSSIM\daa & LPIPS\daa  & Time\daa             & Size\daa  & Memory\daa \\ 
  \midrule
  DyNeRF$^{\dag}$ \cite{Dynerf} & & $\checkmark$ & 300                       & 29.58  &  0.020 &   0.083  & 7 days  &  28MB     &  >100GB      \\
  NeRF-T$^{\dag}$ \cite{Dynerf}  & & $\checkmark$  & 300                   & 28.45  &   0.023 & 0.10   & -                & -        &  >100GB       \\
  Interp-NN \cite{park2023temporal}  & & $\checkmark$  & 300            & 29.88  & -     & 0.096 & 2 days      & \textbf{20MB}     &  >100GB       \\
  HexPlane \cite{HexPlane_}     & & $\checkmark$  & 300               &   31.57  & \textbf{0.016} & 0.089 &   96 min           & 200MB    &  62GB       \\
  K-Planes \cite{kplanes}       & & $\checkmark$  & 300             &  31.63  & -     & -      & 110 min          & 200MB    &  >100GB       \\
  MixVoxels \cite{mixvoxels}  & & $\checkmark$  & 300              & 30.71  & 0.024 & 0.162 & \textbf{15 min}            & 500MB    &  >100GB      \\        
  4DGS \cite{4Dgaussiansplatting}  & & $\checkmark$  & 300              & \textbf{32.01}   & - & \textbf{0.055} & 6 hours            & 4000MB    &   \textbf{<10GB}    \\   
  HyperReel \cite{hyperreel}  & & $\checkmark$  & 50          & 31.1  & - & 0.096 & 9 hours            & 360MB    &      18GB    \\  
  NeRFplayer \cite{nerfplayer}  & & $\checkmark$  & 300               & 30.29  & - & 0.152  & 5.5 hours            & -     &  >100GB   \\ \midrule 
  StreamRF \cite{streamRF} & $\checkmark$ & & 1              & 28.85  & 0.042     & 0.253      &  75 min           & 5000MB   & \textbf{<10GB}       \\
  HexPlane-PR & $\checkmark$ & & 10  & 24.03 & 0.081 &  0.244  & 100 min            & 318MB    &    14GB     \\
  HexPlane-GR & $\checkmark$ & & 10  & 25.17 & 0.050  & 0.272    & 125 min            & 318MB    &    14GB     \\                          
  INV \cite{invnerf} &  $\checkmark$ & & 1             & 29.64  &   0.023 &  0.078 & 40 hours            & 336MB     &  \textbf{<10GB}     \\
  CD-NGP (ours) & $\checkmark$ & & 10        & 30.23  & 0.027& 0.198 &  75 min            &    113MB     &    14GB      \\
  
   \bottomrule  
  \end{tabular}
     \label{tab:comparison_dynerf_dataset}
  \end{table*}

  %---------------------------------------------------------------------------------

  \subsubsection{Comparisons on DyNeRF dataset.}
  \Cref{tab:comparison_dynerf_dataset} shows the quantitative comparisons between CD-NGP and other state-of-the-art methods on DyNeRF dataset. The methods in \cite{Dynerf,park2023temporal} are trained on 8 V100 GPUs or equivalently 4 A100 GPUs. The methods in \cite{Dynerf,park2023temporal,nerfplayer} have no official implementations available. 
  We are unable to obtain the missing quality metrics in \cite{kplanes} because of their huge memory occupation in the default resolution. We report the metrics in the publication if not otherwise specified. 
  HexPlane \cite{HexPlane_} uses resolution $1024 \times 768$ in their implementation, so the memory occupation is much smaller than other offline methods. 
   We use the official implementation of StreamRF \cite{streamRF} and INV \cite{invnerf} and report the average metrics on the 5 synchronized scenes of the dataset. StreamRF \cite{streamRF} and INV \cite{invnerf} are trained per frame. 
   NeRFplayer \cite{nerfplayer} is trained offline but streamable upon inference. 4DGS \cite{4Dgaussiansplatting} caches the frames onto disk for training to alleviate memory occupation with a trade-off on training speed. 
   The reconstruction quality is measured by PSNR for pixel-wise similarities, DSSIM for structural similarities, and LPIPS \cite{LPIPS} for perceptual similarities (by AlexNet \cite{Alexnet}) following common practices \cite{Dynerf,HexPlane_,mixvoxels}. 
   To prove the viability of our parameter-isolation-based method, we apply the previous state-of-the-art method \emph{HexPlane} \cite{HexPlane_} to continual learning setting with parameter-regularization (\emph{HexPlane-PR}) and generative replay (\emph{HexPlane-GR}) as additional baselines. 
   
  %--------------------------------------------- Figure ------------------------------------

  %---------------------------------------------------------------------------------
   We compare our method with both online and offline baselines. Our method shows significantly better efficiency and quality than the other online methods.  
   As demonstrated in \cref{tab:comparison_dynerf_dataset}, our CD-NGP is $32\times$ faster than the MLP-based continual representation INV \cite{invnerf}, $44\times$ smaller than the voxel-based continual representation StreamRF \cite{streamRF} and significantly outperforms these baselines in quality metrics. 
  The undesirable results from HexPlane-PR and HexPlane-GR show the limitations of the regularization-based and replay-based methods in dynamic scenes: unlike the static scenes, the objects are not stationary across time and the models with limited complexity are not capable of reconstructing unlimited transient effects. 
  Benefiting from parameter isolation like StreamRF \cite{streamRF} and INV \cite{invnerf}, our method achieves fast convergence and compact model size simultaneously. It leverages the hash tables to ensure fast convergence and reuses the features from differently configured spatial hash tables to ensure high scalability. 

  Compared with most of the prevailing offline methods \cite{Dynerf,park2023temporal,HexPlane_,kplanes,mixvoxels}, our method only requires caching $T_{chunk}=10$ frames into the buffer by continual learning and thus reduces memory occupation significantly. Although the offline baseline 4DGS \cite{4Dgaussiansplatting} consumes lower memory by using a lazy load strategy with a trade-off on training speed, their implementation consumes another 27 GB hard disk space on 300-frame DyNeRF dataset, and the space requirement also scales at a complexity of O($N_{frames}$) as the number of frames increases. On the contrary, our method leverages the continual learning strategy and enables directly decoding the frames from the videos \emph{without any additional storage consumption} for training (scales at  $O(1)$ complexity). 
  It also balances training speed and model size by leveraging the feature fusion of the \emph{base} and the \emph{auxiliary} branches and delivers high quality.  

  % 使用minipage环境，强制两张图贴在一起
  \begin{figure*}[!ht]
    \centering
    \captionsetup{skip=0pt}
    \captionsetup[sub]{font=normalsize}  % 设置subcaption的字体大小
    \begin{minipage}{\textwidth}
        \centering
        \begin{subfigure}[b]{0.3\textwidth}
          \centering
          \includegraphics[width=\textwidth]{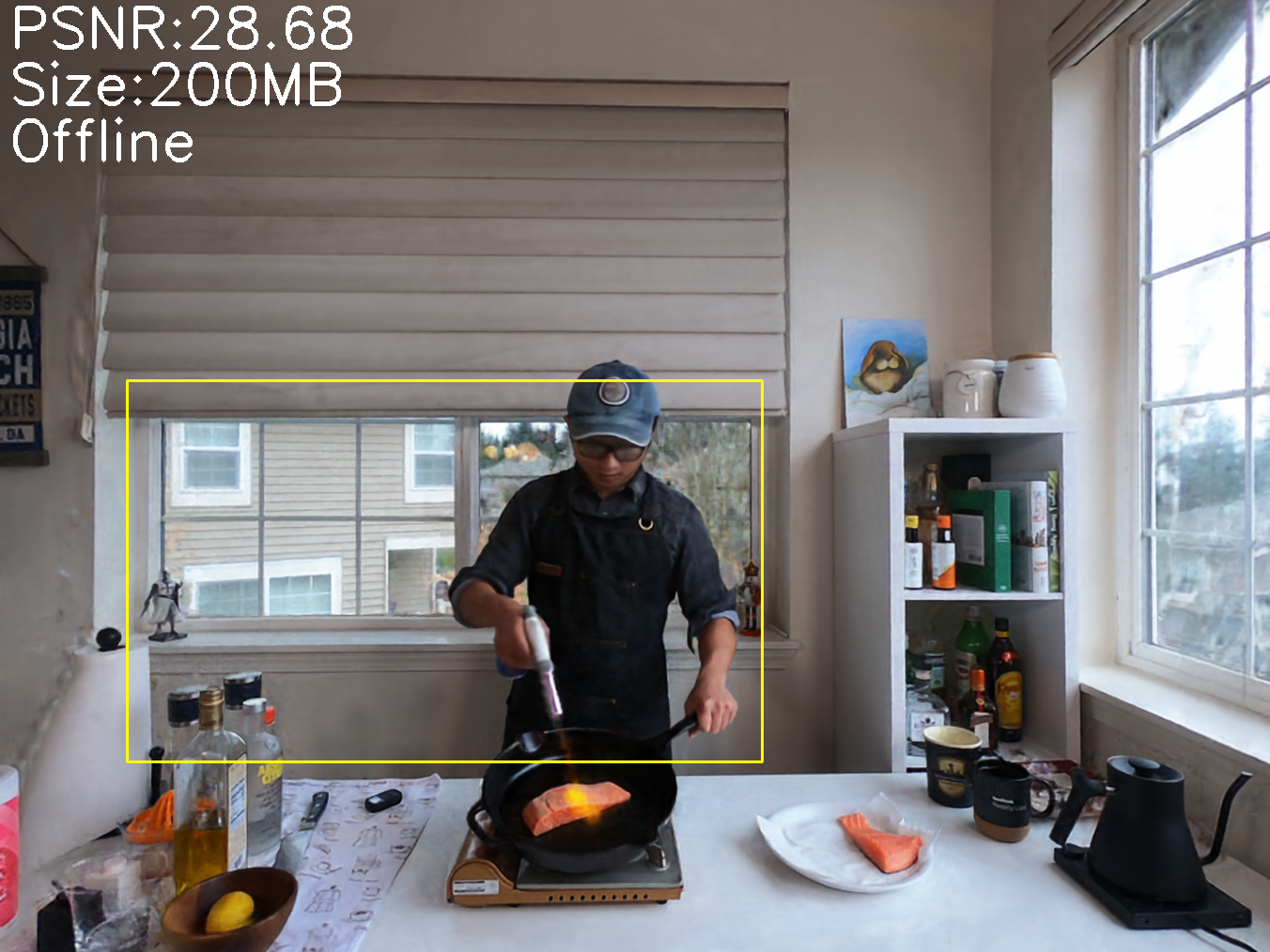}
          \caption{HexPlane-offline}
          \label{fig:cmp hexplaneoffline}
      \end{subfigure}
      \begin{subfigure}[b]{0.3\textwidth}
          \centering
          \includegraphics[width=\textwidth]{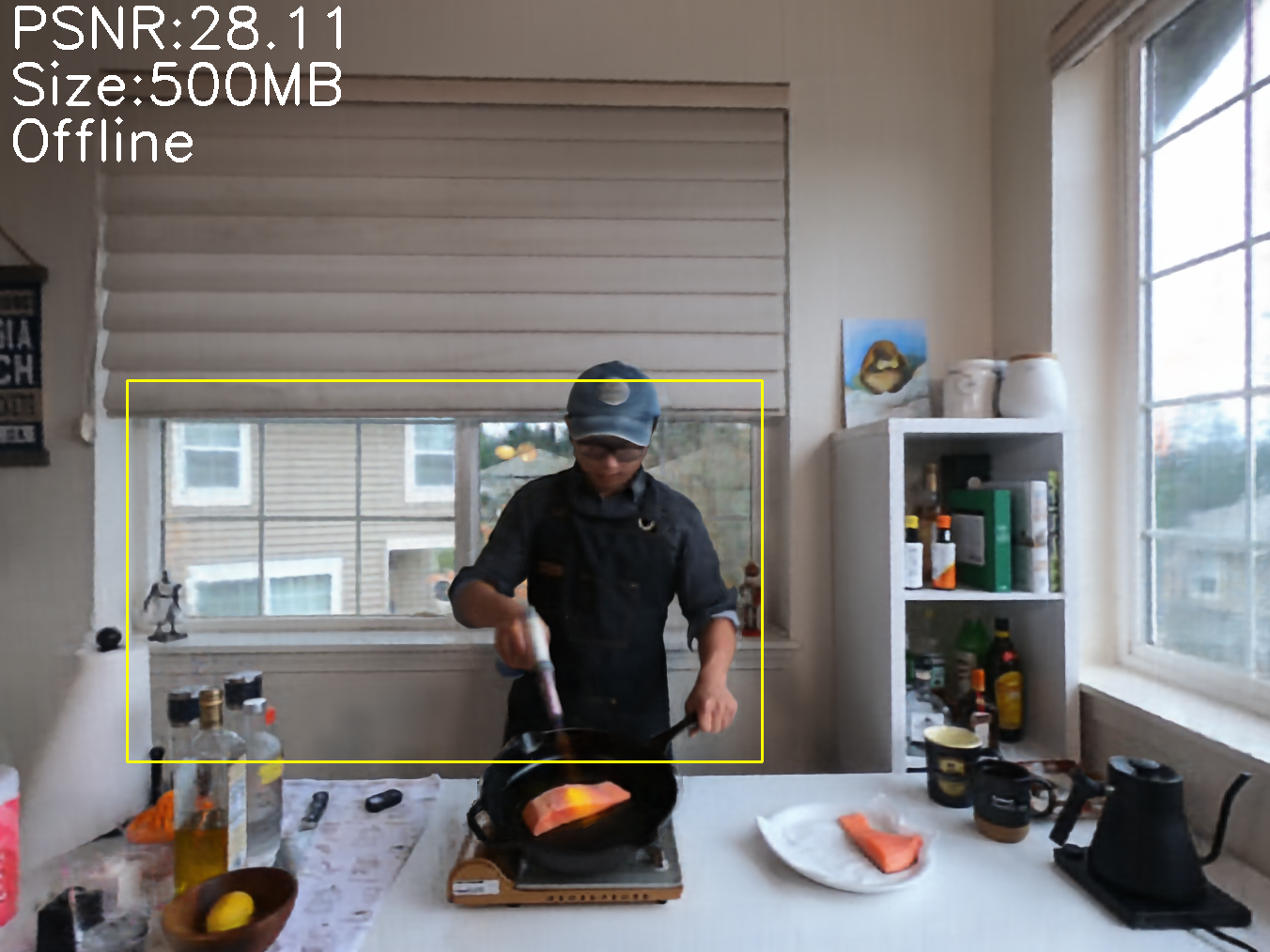}
          \caption{MixVoxels}
          \label{fig:cmp mixvoxels}
      \end{subfigure}
      \begin{subfigure}[b]{0.3\textwidth}
          \centering
          \includegraphics[width=\textwidth]{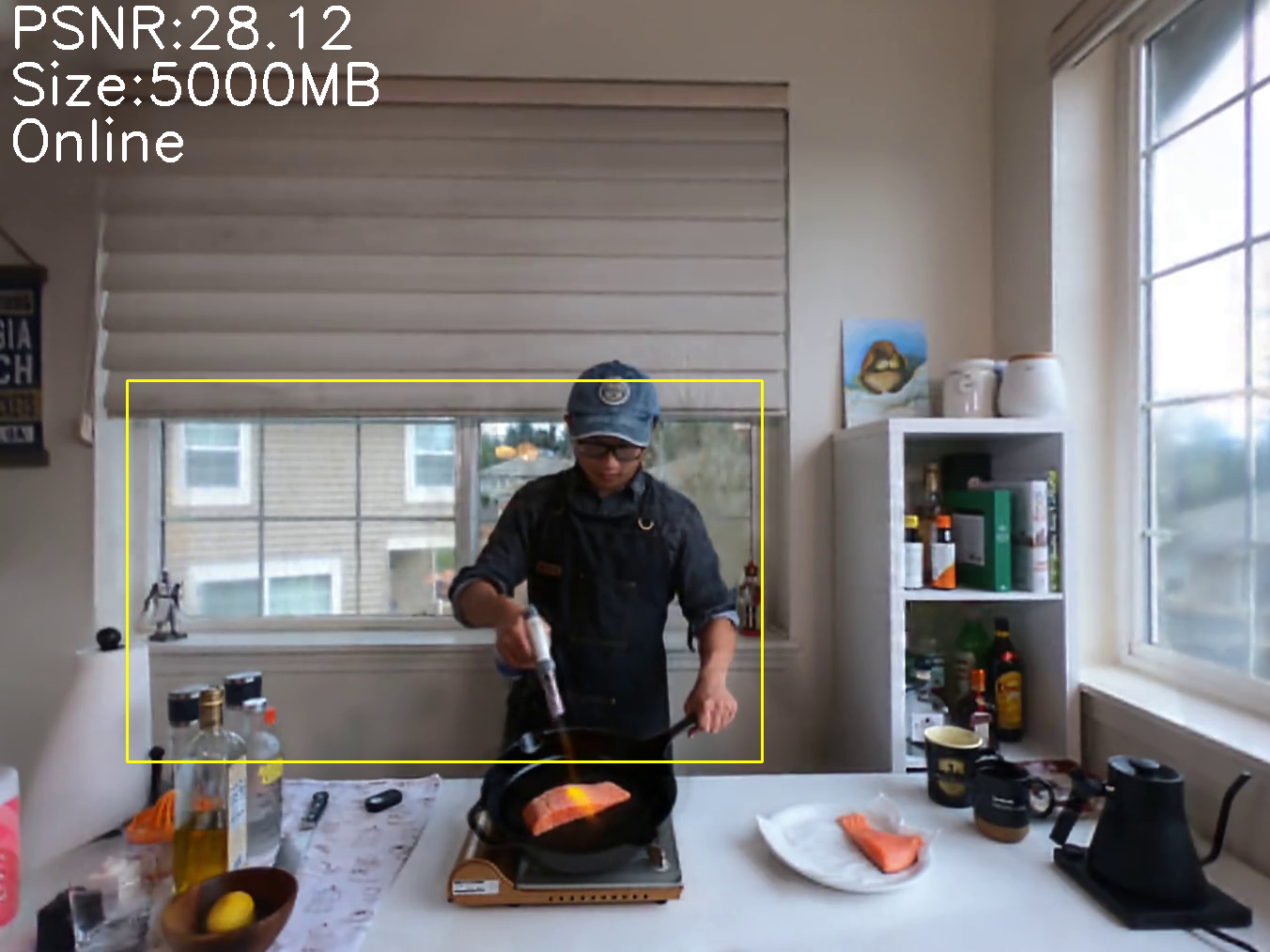}
          \caption{StreamRF}
          \label{fig:cmp streamRF}
      \end{subfigure}
      \begin{subfigure}[b]{0.3\textwidth}
      \centering
      \includegraphics[width=\textwidth]{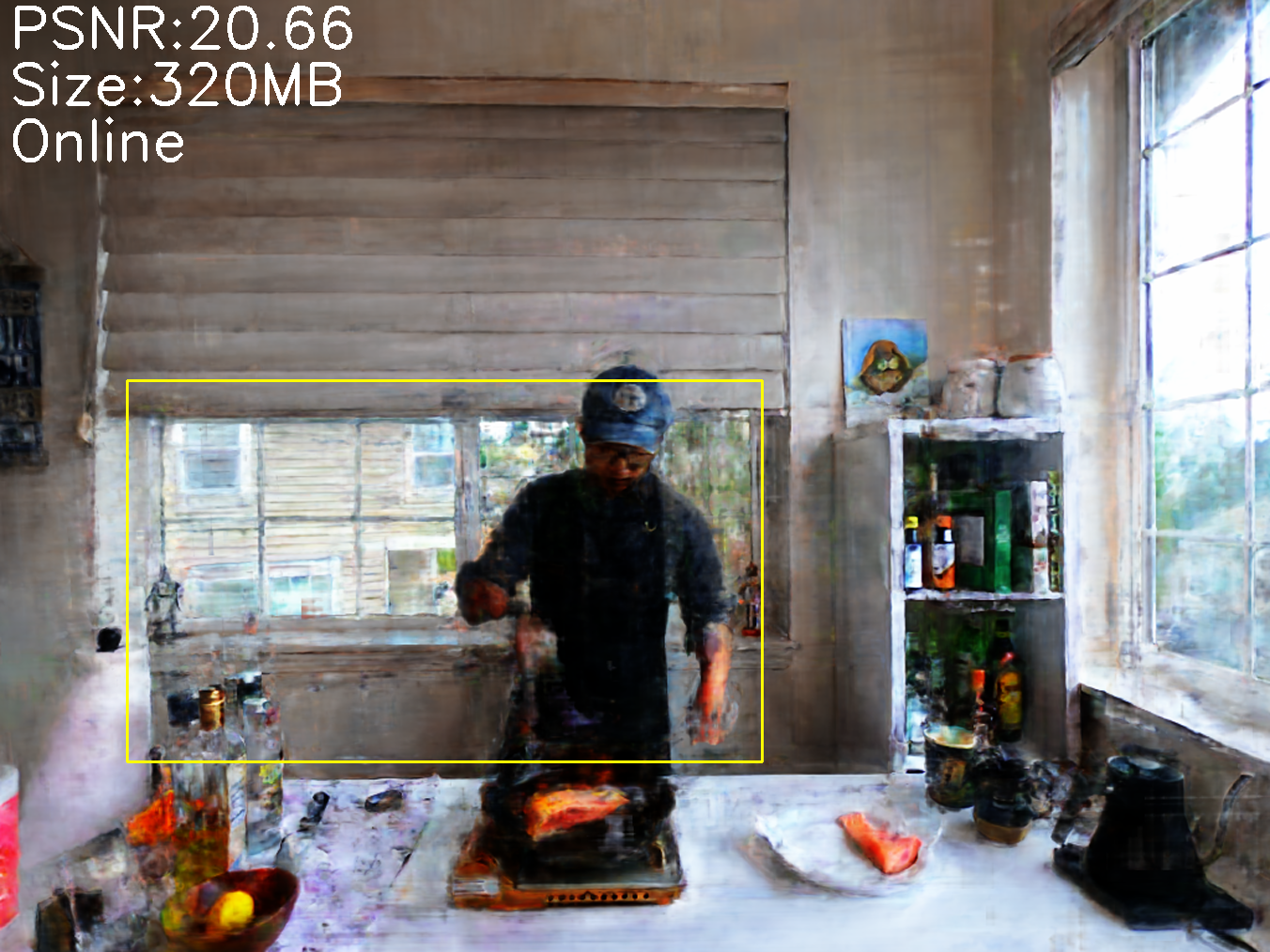}
      \caption{HexPlane-PR}
      \label{fig:cmp HexPlane-PR}
      \end{subfigure}
      \begin{subfigure}[b]{0.3\textwidth}
      \centering
      \includegraphics[width=\textwidth]{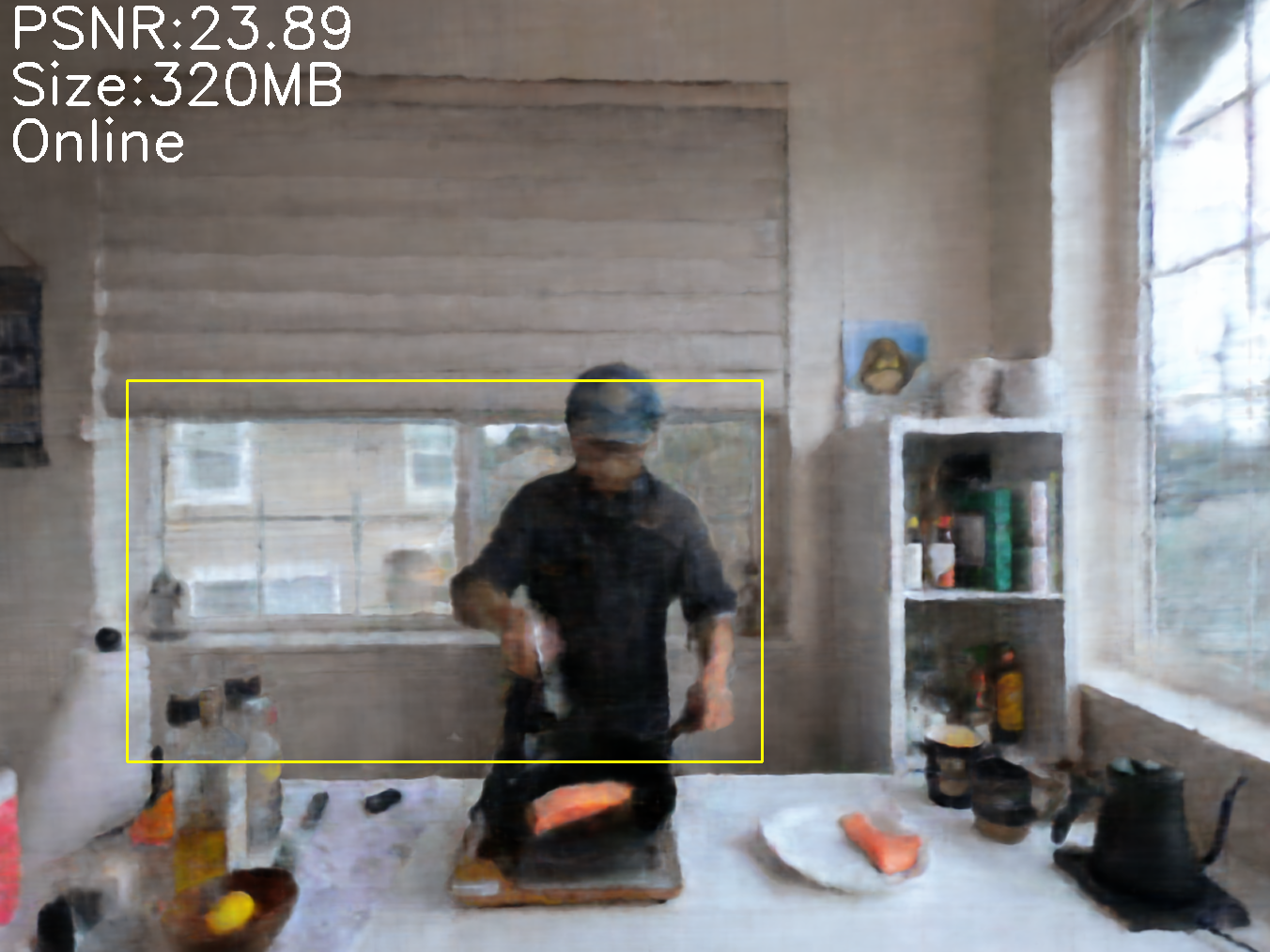}
      \caption{HexPlane-GR}
      \label{fig:HexPlane-GR}
      \end{subfigure}
      \begin{subfigure}[b]{0.3\textwidth}
      \centering
      \includegraphics[width=\textwidth]{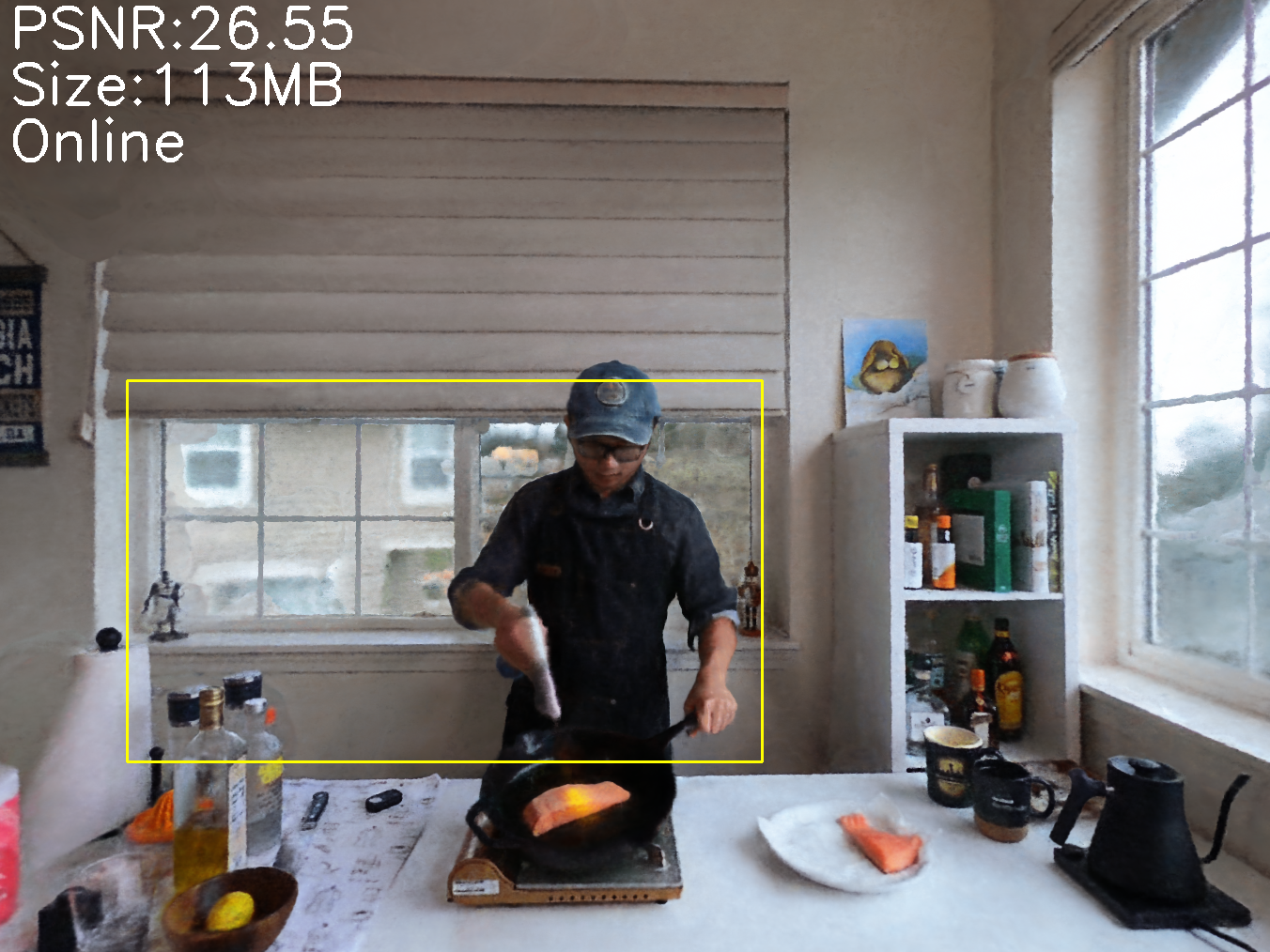}
      \caption{CD-NGP (ours)}
      \label{fig:cmp:CD-NGP}
      \end{subfigure}
      \vspace{0pt}
      \caption{Comparison of reconstruction quality on the most challenging \emph{flame salmon} scene in DyNeRF dataset. 
      PSNRs are computed under the resolution $1352\times1014$.}
      \label{fig:compatison_dynerf_dataset}
    \end{minipage}
    %\vspace{3pt}  % 确保两张图紧挨在一起 % 太丑了，算了
    \begin{minipage}{\textwidth}
        \centering
        \begingroup
        \renewcommand{\arraystretch}{0.8}  % 缩短行间距
        \setlength{\tabcolsep}{1pt}        % 缩短列间距
        \begin{tabular}{cccc}
            \includegraphics[width=0.24\textwidth]{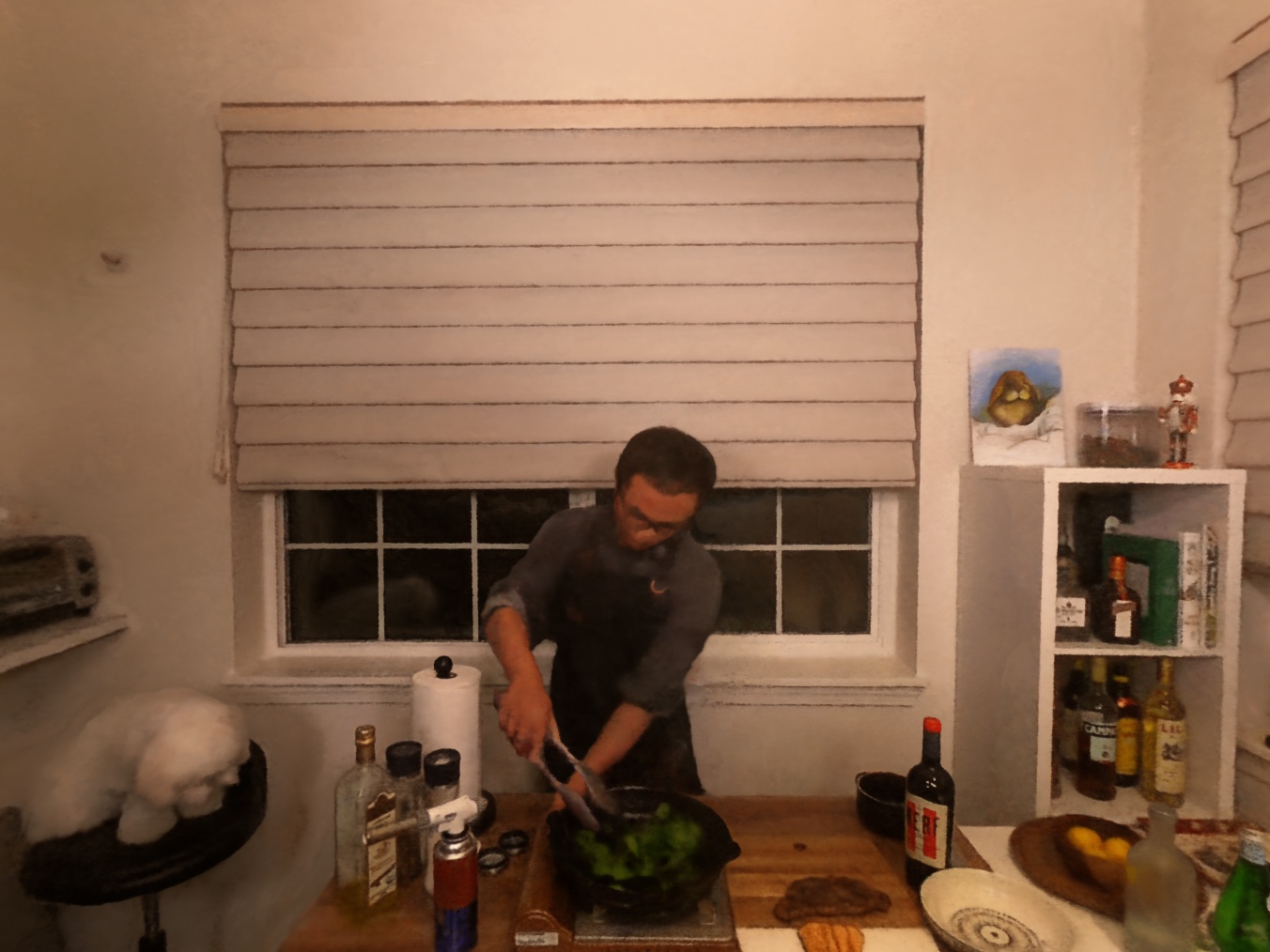} &
            \includegraphics[width=0.24\textwidth]{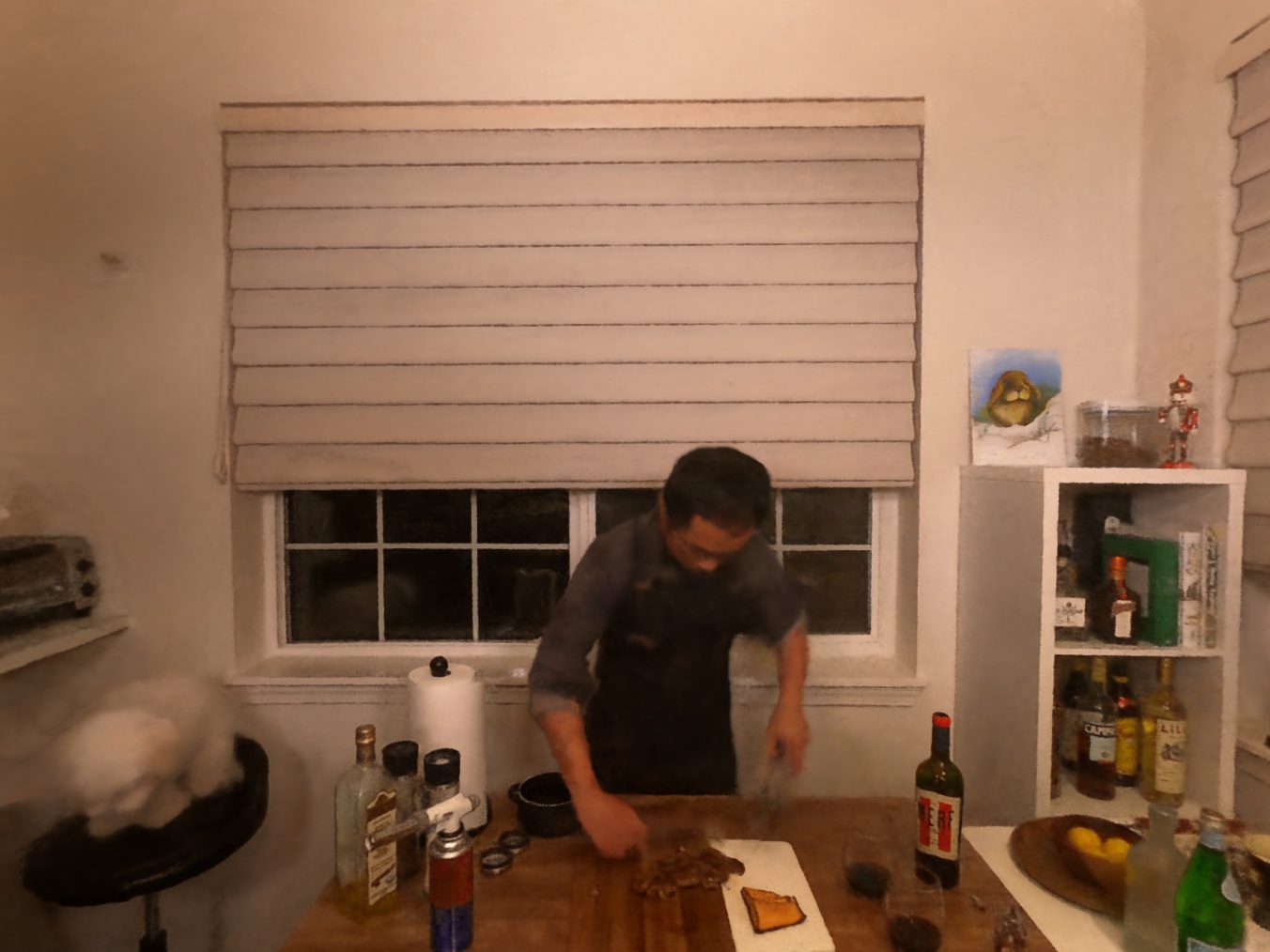} &
            \includegraphics[width=0.24\textwidth]{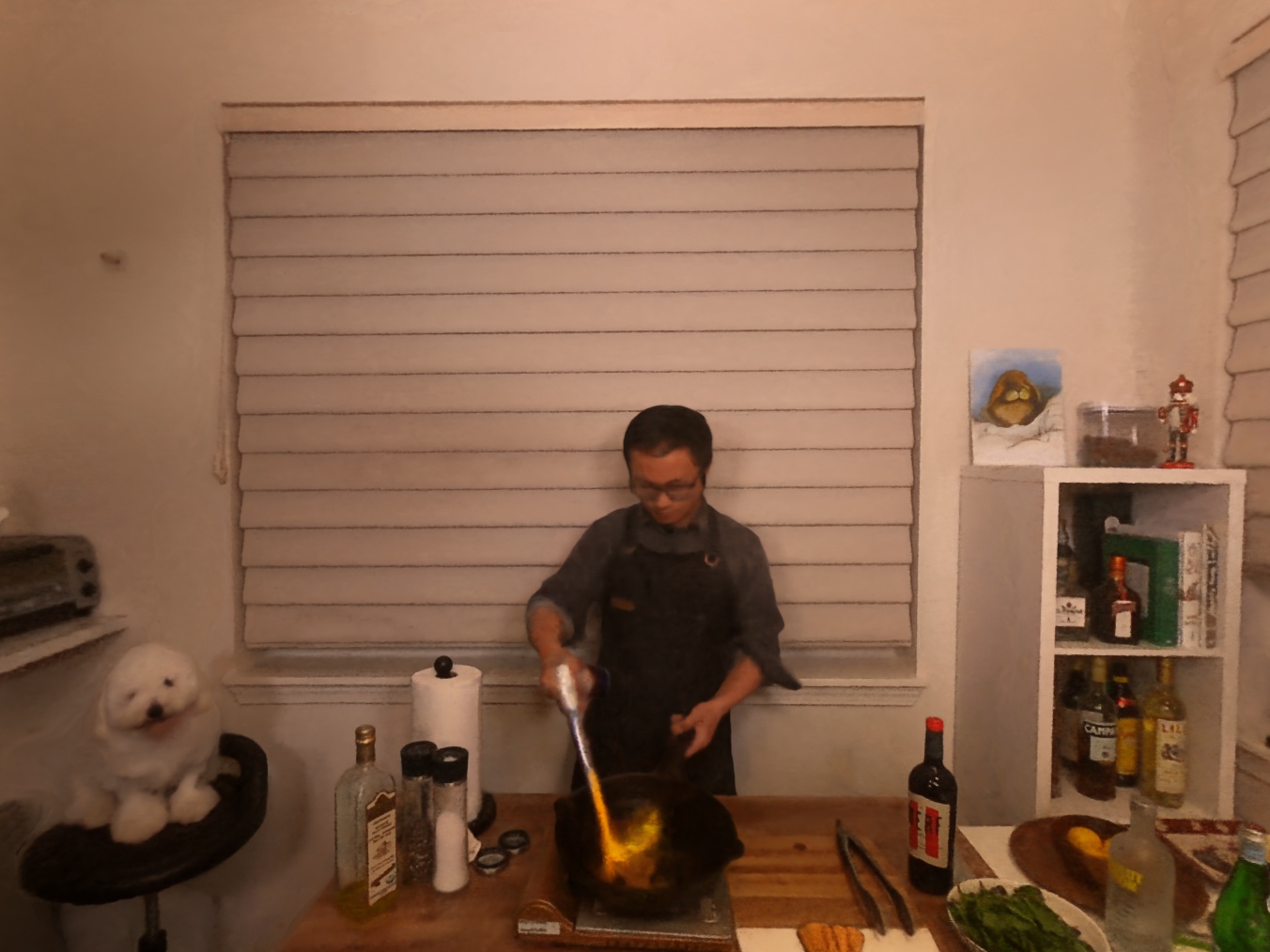} &
            \includegraphics[width=0.24\textwidth]{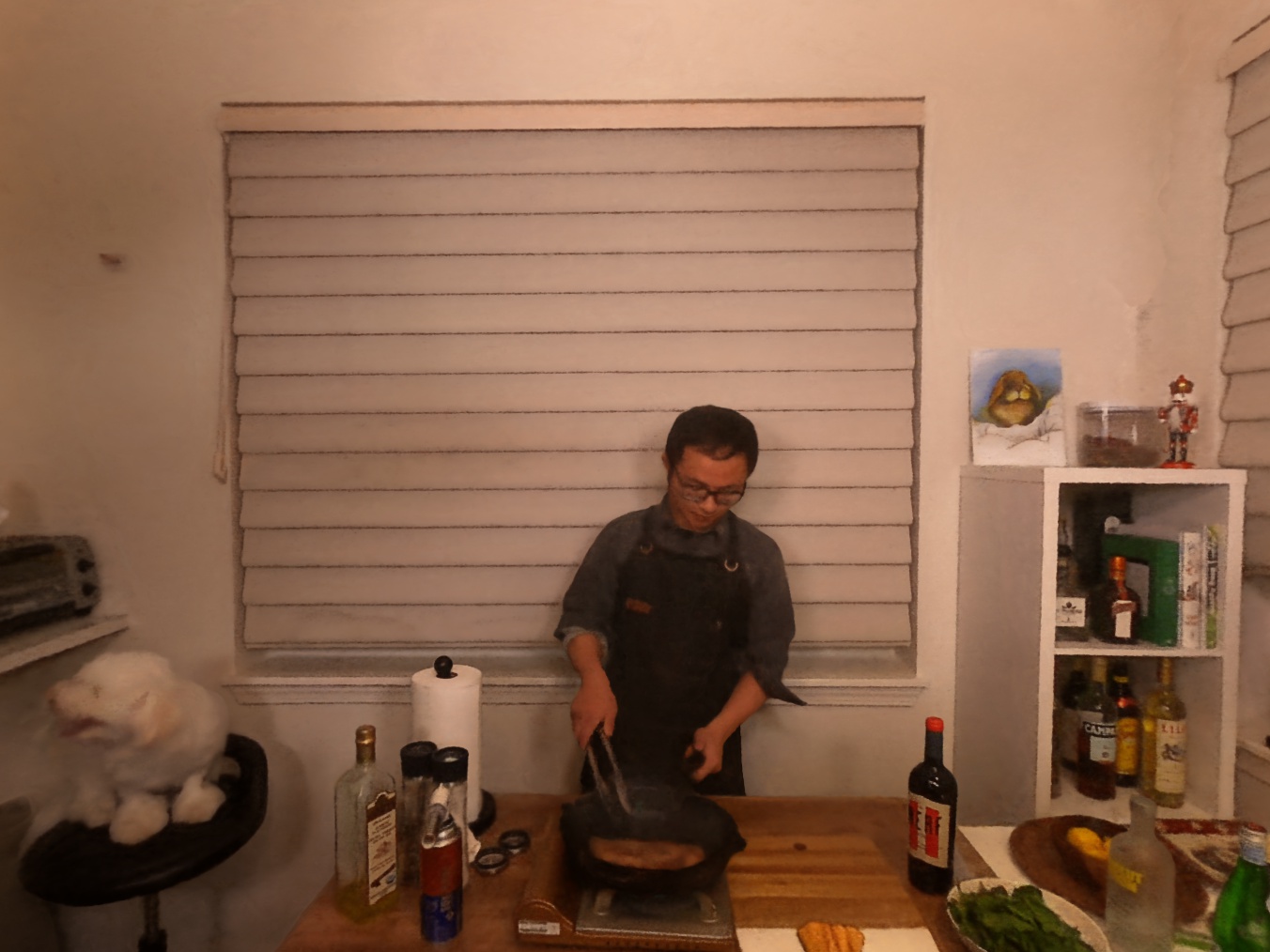} \\
            \includegraphics[width=0.24\textwidth]{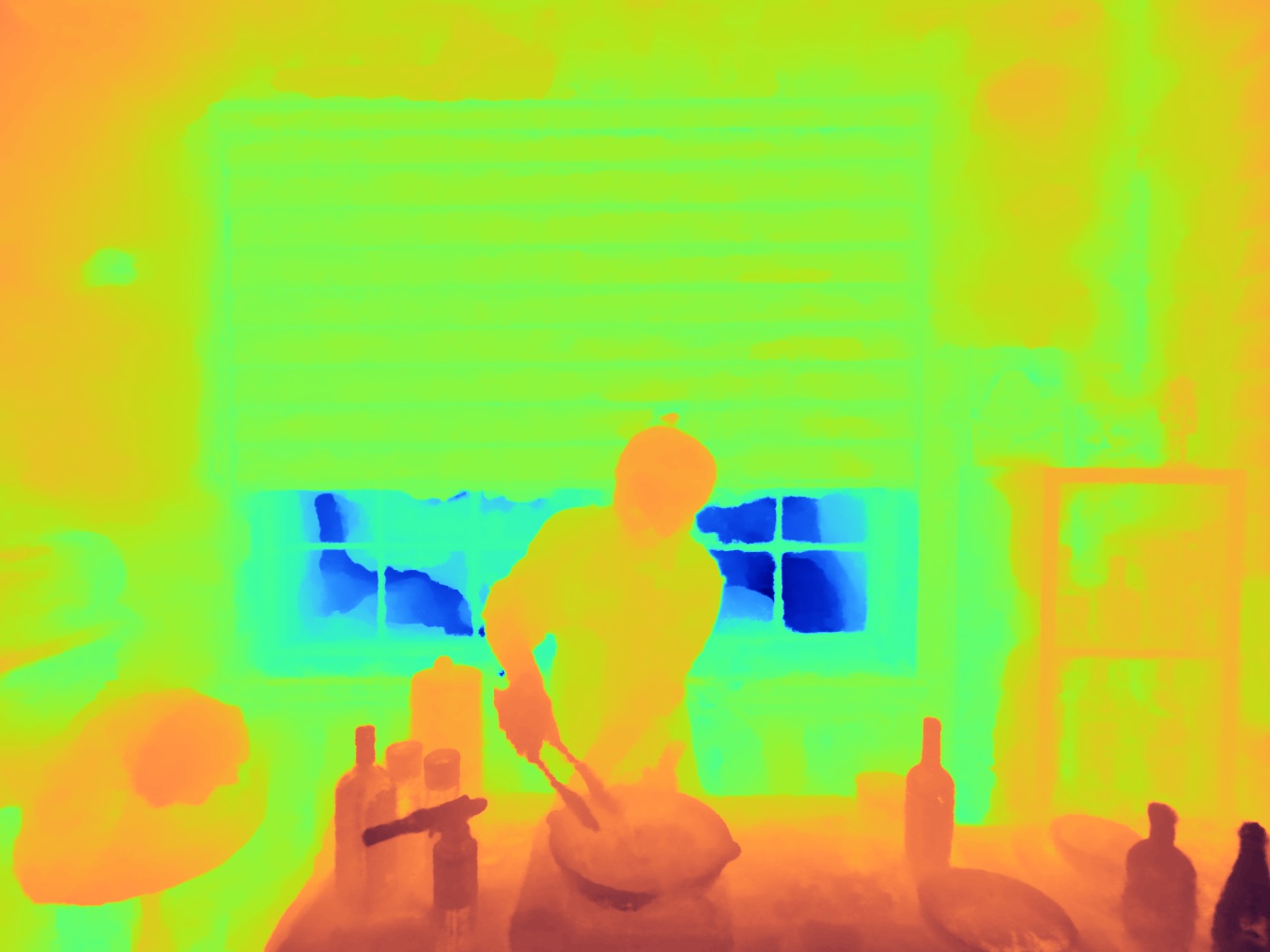} &
            \includegraphics[width=0.24\textwidth]{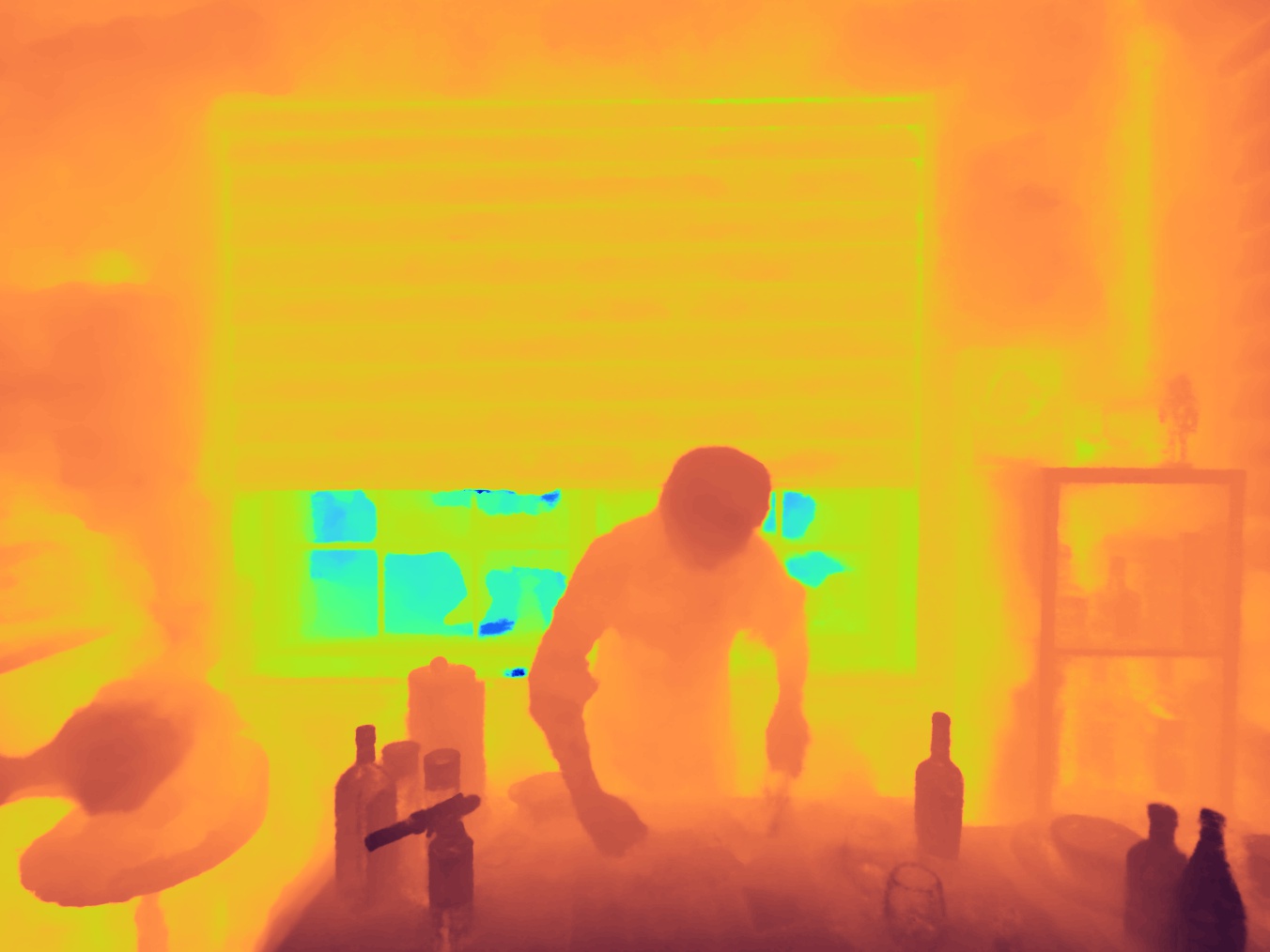} &
            \includegraphics[width=0.24\textwidth]{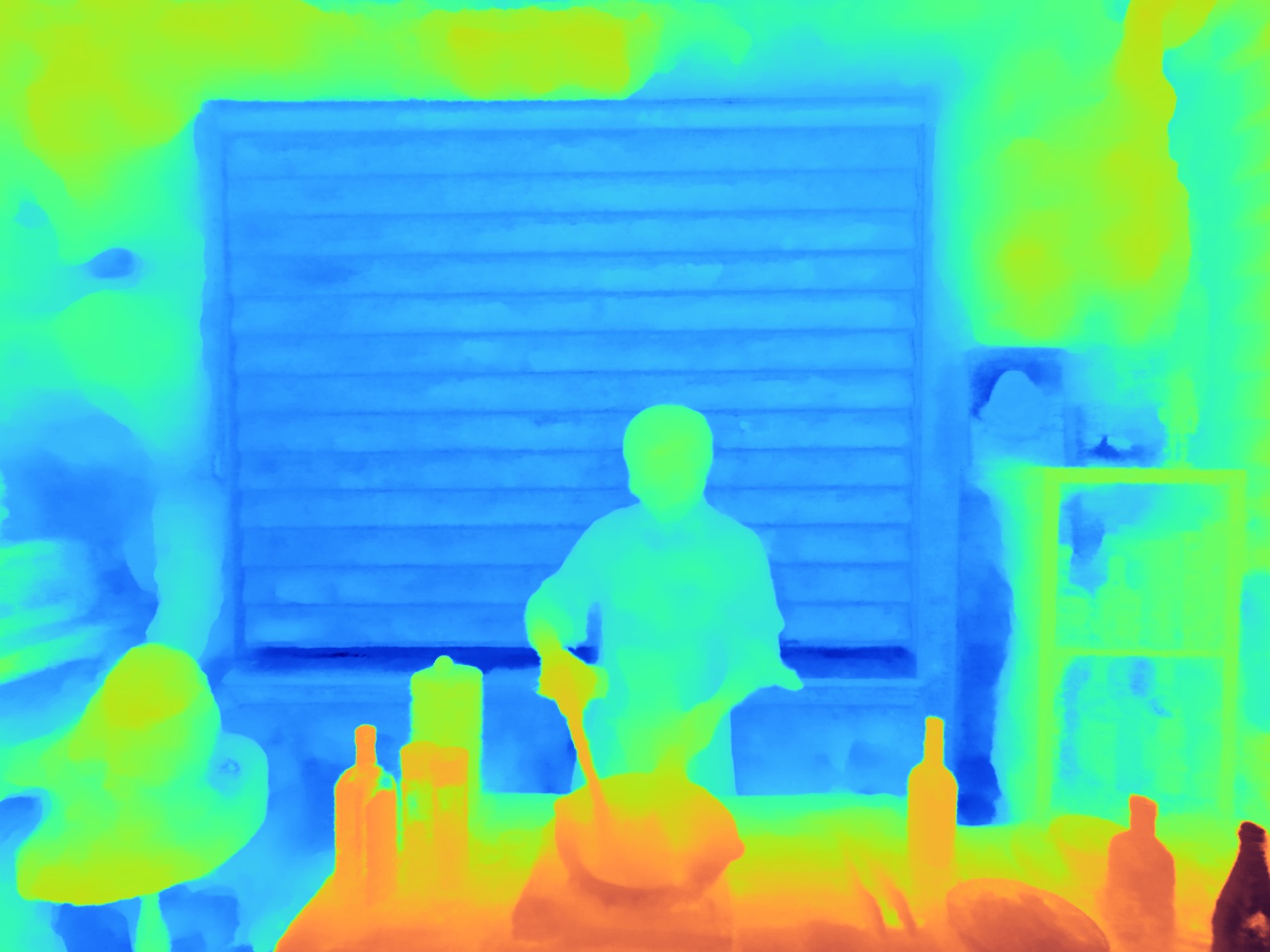} &
            \includegraphics[width=0.24\textwidth]{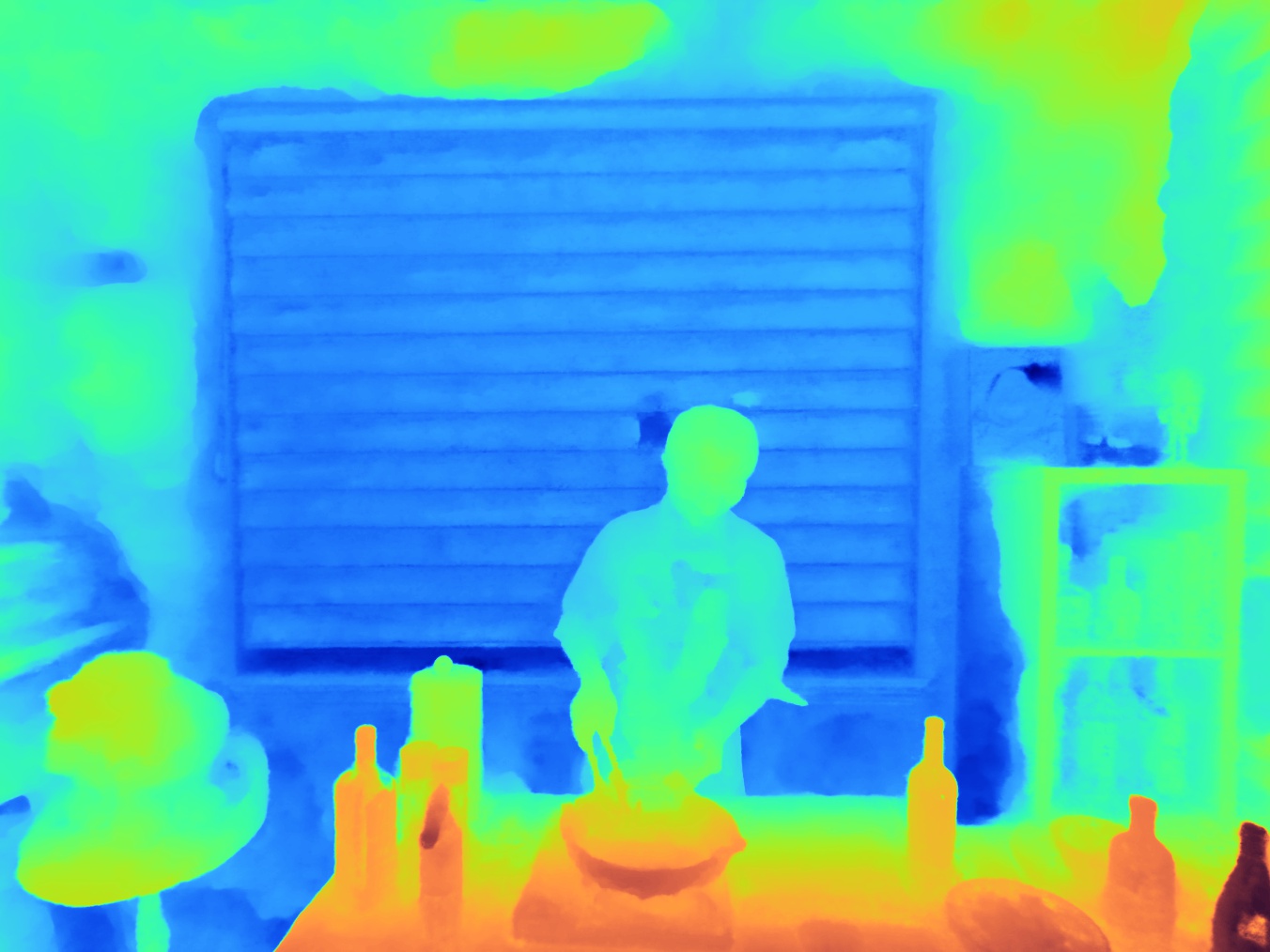} \\
        \end{tabular}
        \endgroup
        \caption{The results of the proposed CD-NGP on the DyNeRF dataset.}
        \label{fig:mainexp_results_demon}
    \end{minipage}
\end{figure*}

  \Cref{fig:compatison_dynerf_dataset} provides qualitative comparisons. 
  Compared with offline methods in \cref{fig:cmp hexplaneoffline}, \cref{fig:cmp mixvoxels} and the online method in \cref{fig:cmp streamRF}, our model is much smaller without large quality loss. 
  Compared with regularization-based and replay-based online methods in  \cref{fig:cmp HexPlane-PR} and \cref{fig:HexPlane-GR} respectively, our method achieves significantly higher quality. 
  Results of other scenes including depth maps are shown in \cref{fig:mainexp_results_demon}. 
%--------------------------------------------- Figure ------------------------------------

\begin{figure*}[!ht]
  \centering
  \captionsetup{skip=0pt}
  \captionsetup[sub]{font=normalsize}  % 设置subcaption的字体大小
  \begin{subfigure}[b]{0.3\textwidth}
      \centering
      \includegraphics[width=\linewidth]{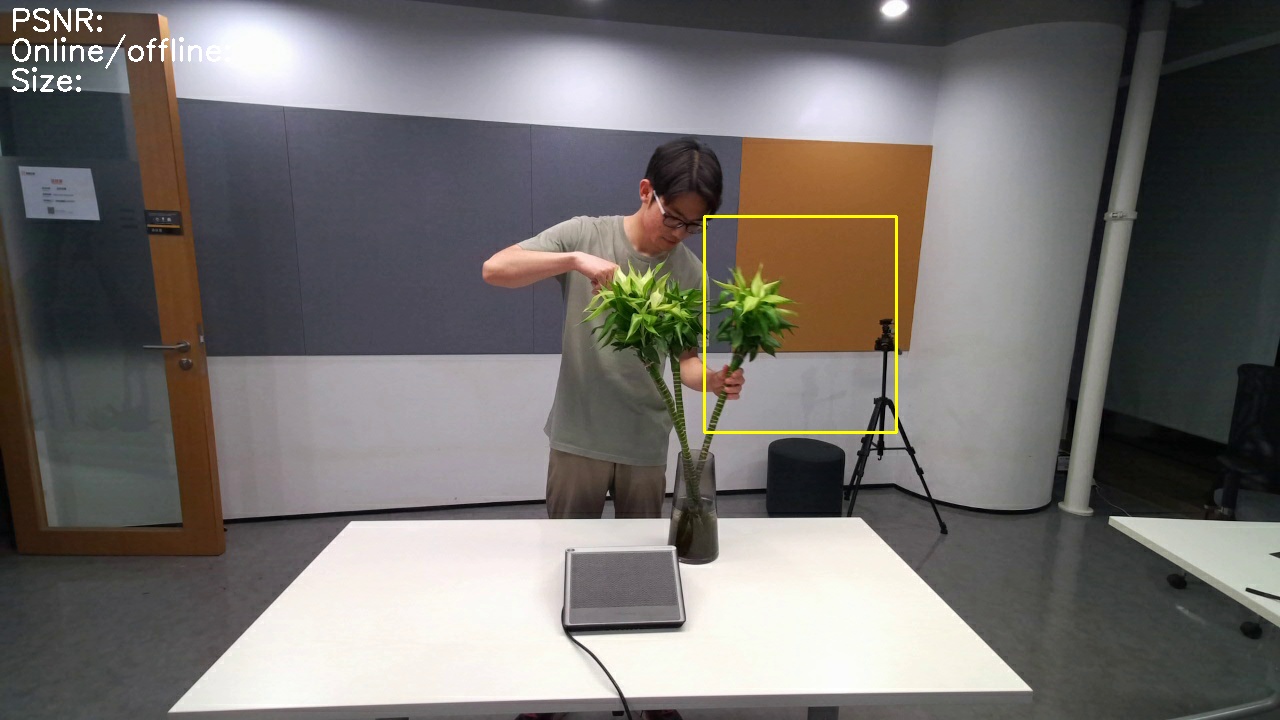}
      \caption{GT}
      \label{fig:meetroom_gt}
  \end{subfigure}
  \begin{subfigure}[b]{0.3\textwidth}
      \centering
      \includegraphics[width=\linewidth]{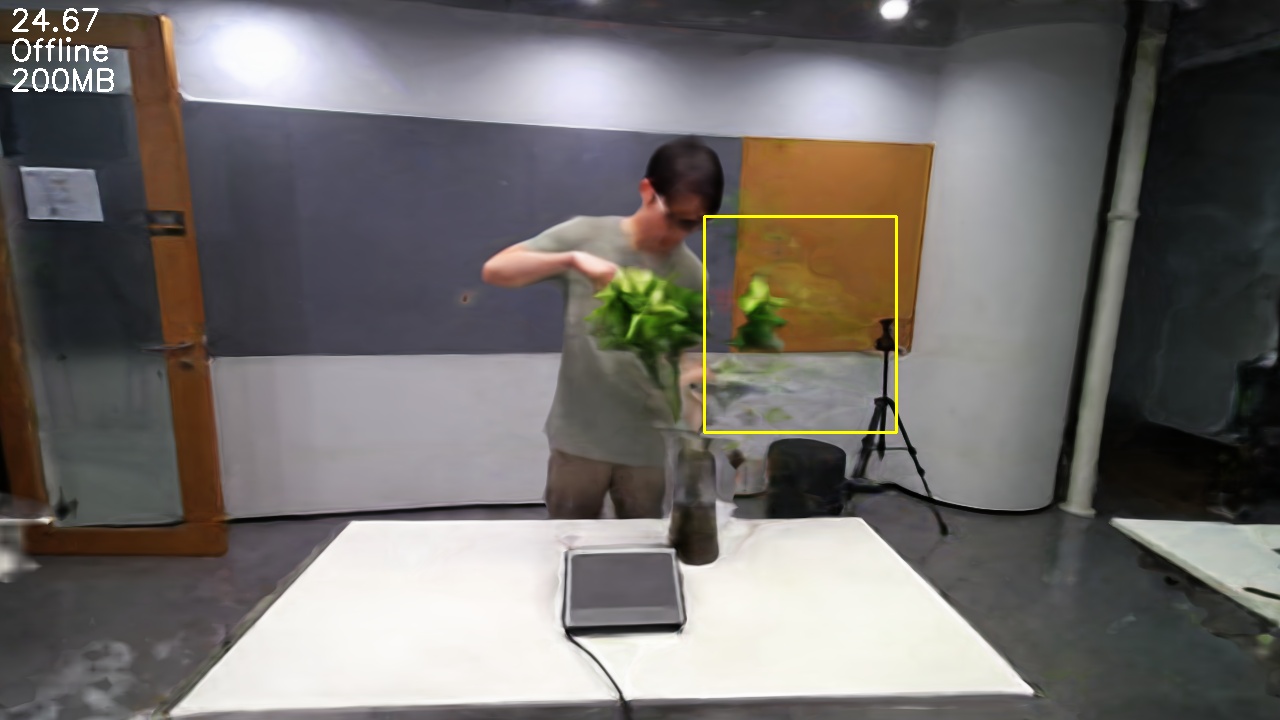}
      \caption{HexPlane}
      \label{fig:meetroom_hexplane}
  \end{subfigure}
  \begin{subfigure}[b]{0.3\textwidth}
      \centering
      \includegraphics[width=\linewidth]{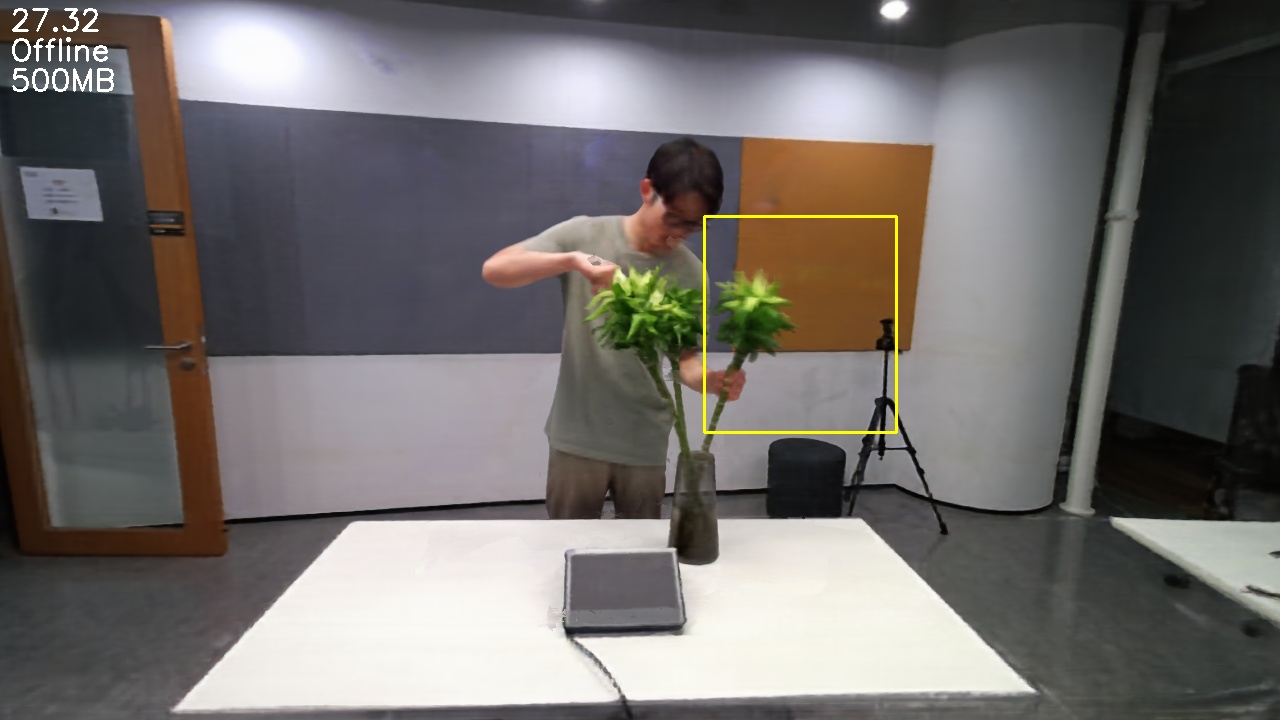}
      \caption{MixVoxels}
      \label{fig:meetroom_mixvoxels}
  \end{subfigure}
  \begin{subfigure}[b]{0.3\textwidth}
      \centering
      \includegraphics[width=\linewidth]{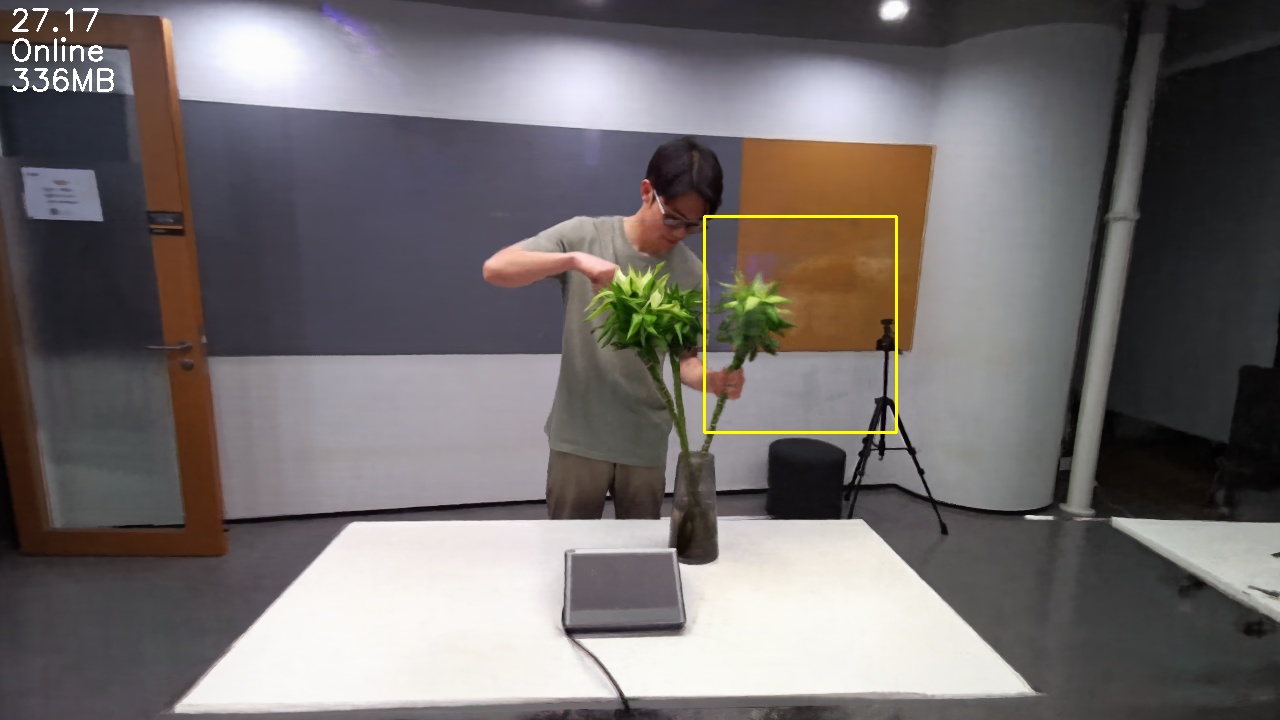}
      \caption{INV}
      \label{fig:meetroom_inv}
  \end{subfigure}
  \begin{subfigure}[b]{0.3\textwidth}
      \centering
      \includegraphics[width=\linewidth]{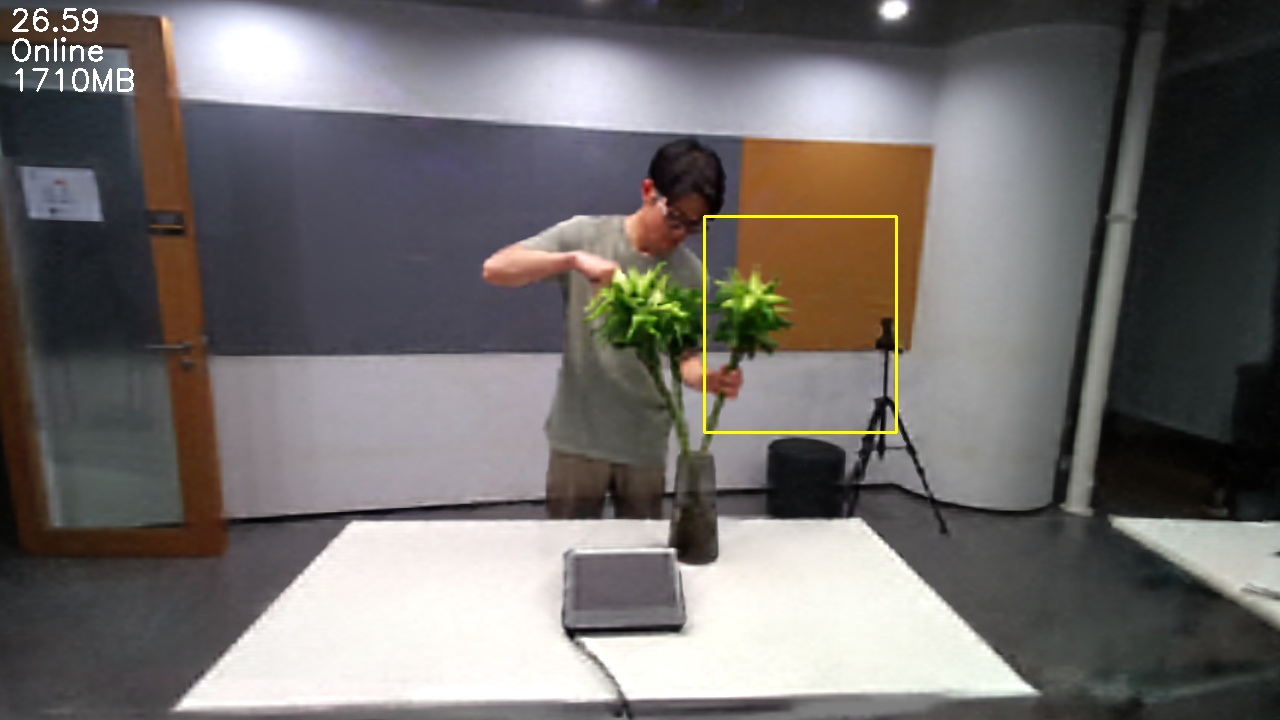}
      \caption{StreamRF}
      \label{fig:meetroom_streamRF}
  \end{subfigure}
  \begin{subfigure}[b]{0.3\textwidth}
      \centering
      \includegraphics[width=\linewidth]{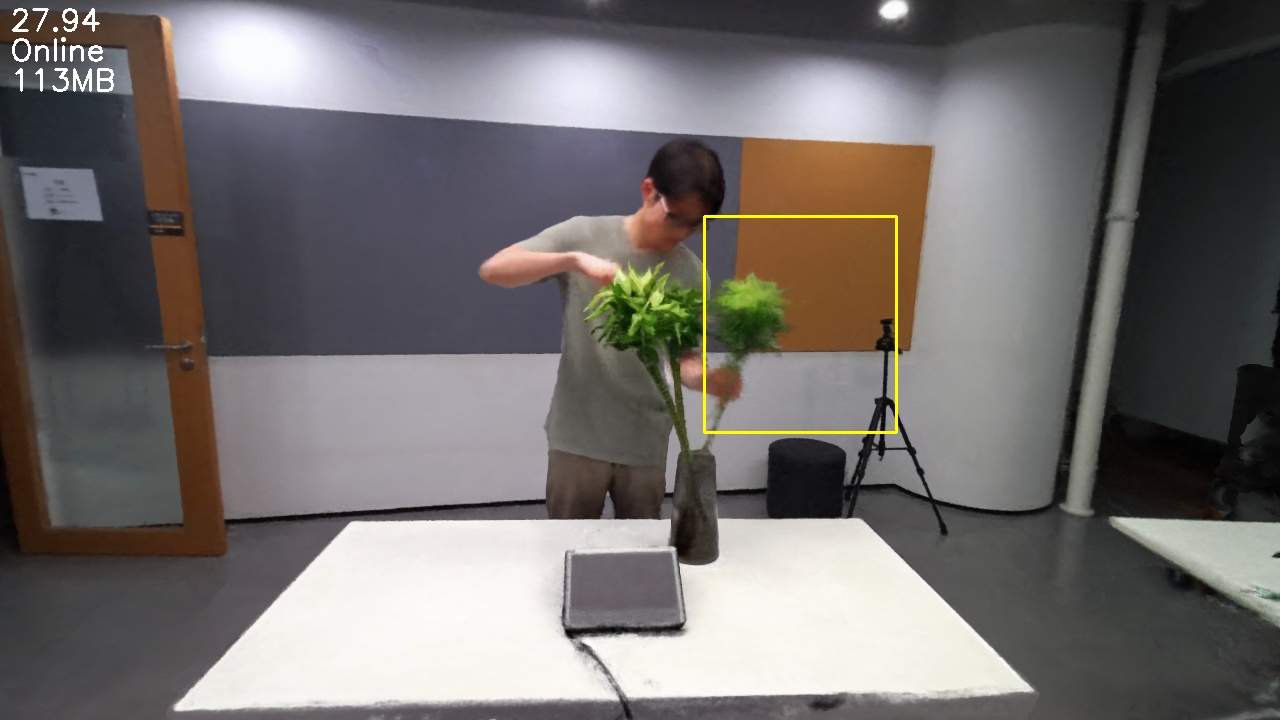}
      \caption{Ours}
      \label{fig:meetroom_ours}
  \end{subfigure}
  \caption{Comparison of reconstruction quality on the \emph{trimming} scene in the Meetroom dataset. Our method produces comparable quality to both offline and online state-of-the-art methods but with much less memory and storage cost.}
  \label{fig:comparison meetroom dataset}
\end{figure*}

%---------------------------------------------------------------------------------

\subsubsection{Comparisons on the Meetroom dataset.}
\begin{figure*}[!htbp]
    \centering
    \captionsetup{skip=0pt}
    \captionsetup[sub]{font=normalsize}  % 设置subcaption的字体大小
    \begin{subfigure}[b]{0.24\textwidth}
        \centering
        \includegraphics[width=\linewidth]{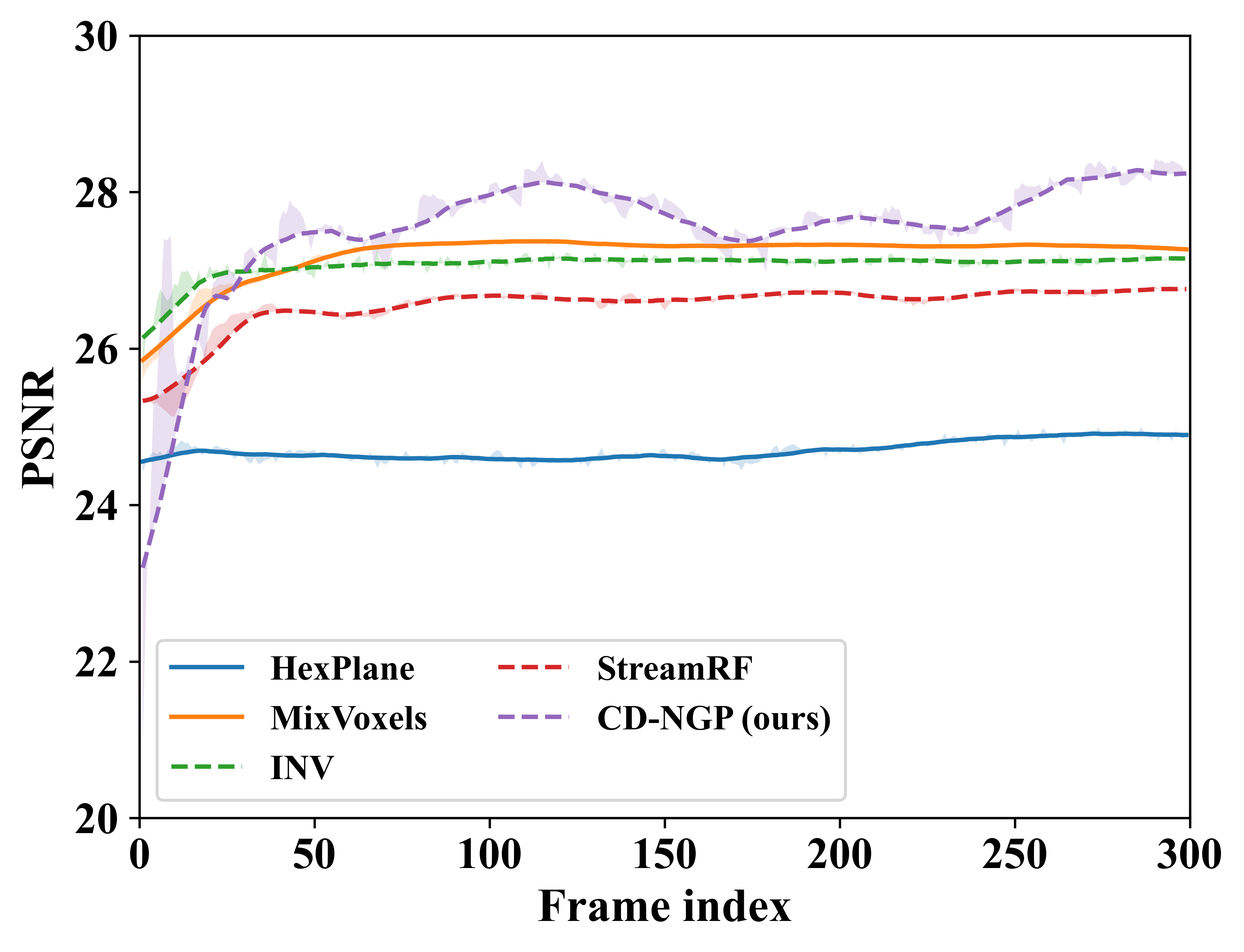}
        \caption{}
        \label{fig:psnr_curve_meetroom_trimming}
    \end{subfigure}
    \hfill
    \begin{subfigure}[b]{0.24\textwidth}
        \centering
        \includegraphics[width=\linewidth]{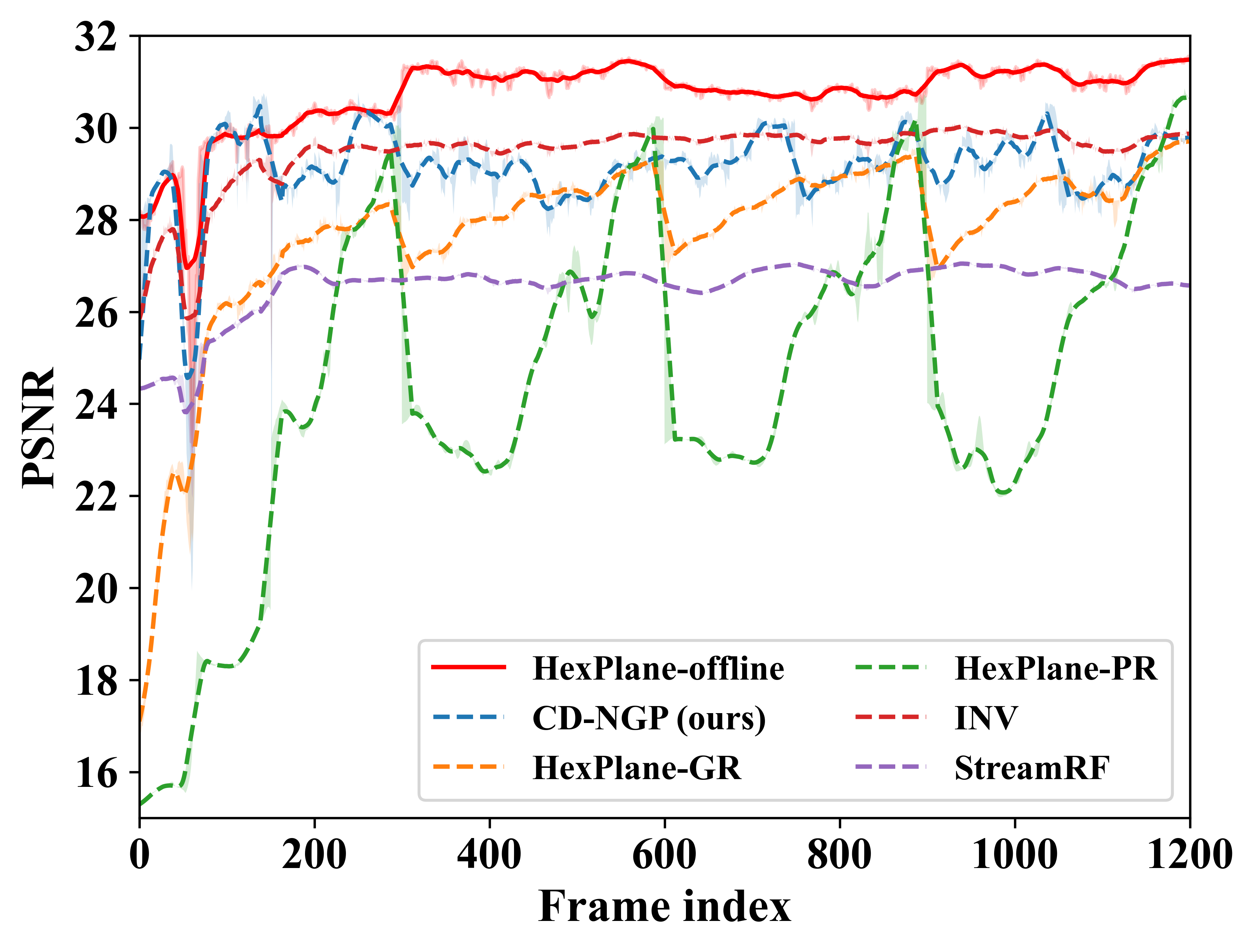}
        \caption{}
        \label{fig:PSNR_curve_long_video_sequence}
    \end{subfigure}
    \hfill
    \begin{subfigure}[b]{0.24\textwidth}
        \centering
        \includegraphics[width=\linewidth]{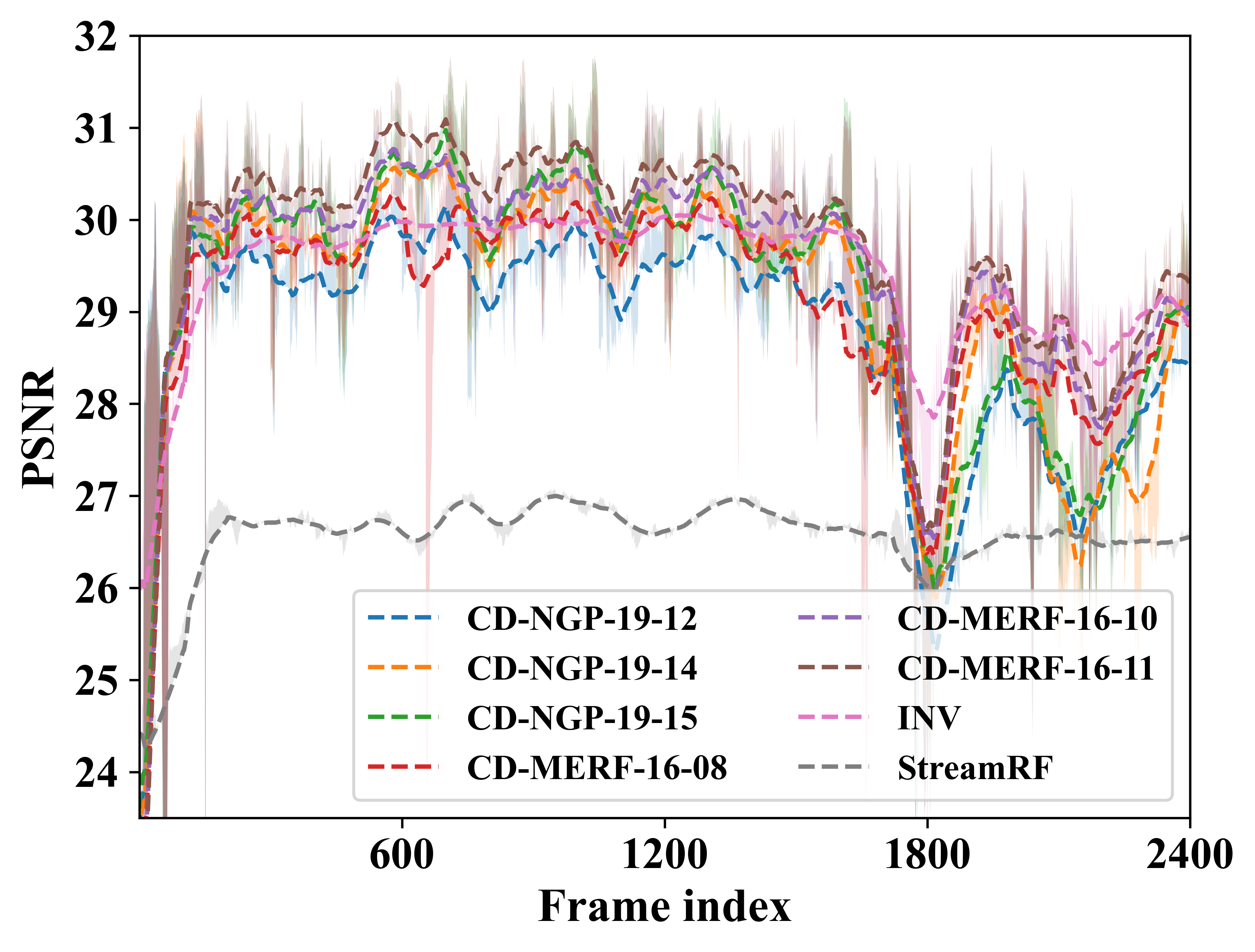}
        \caption{}
        \label{fig:PSNR_curve_2400_frame_long_video_sequence}
    \end{subfigure}
    \hfill
    \begin{subfigure}[b]{0.24\textwidth}
        \centering
        \includegraphics[width=\linewidth]{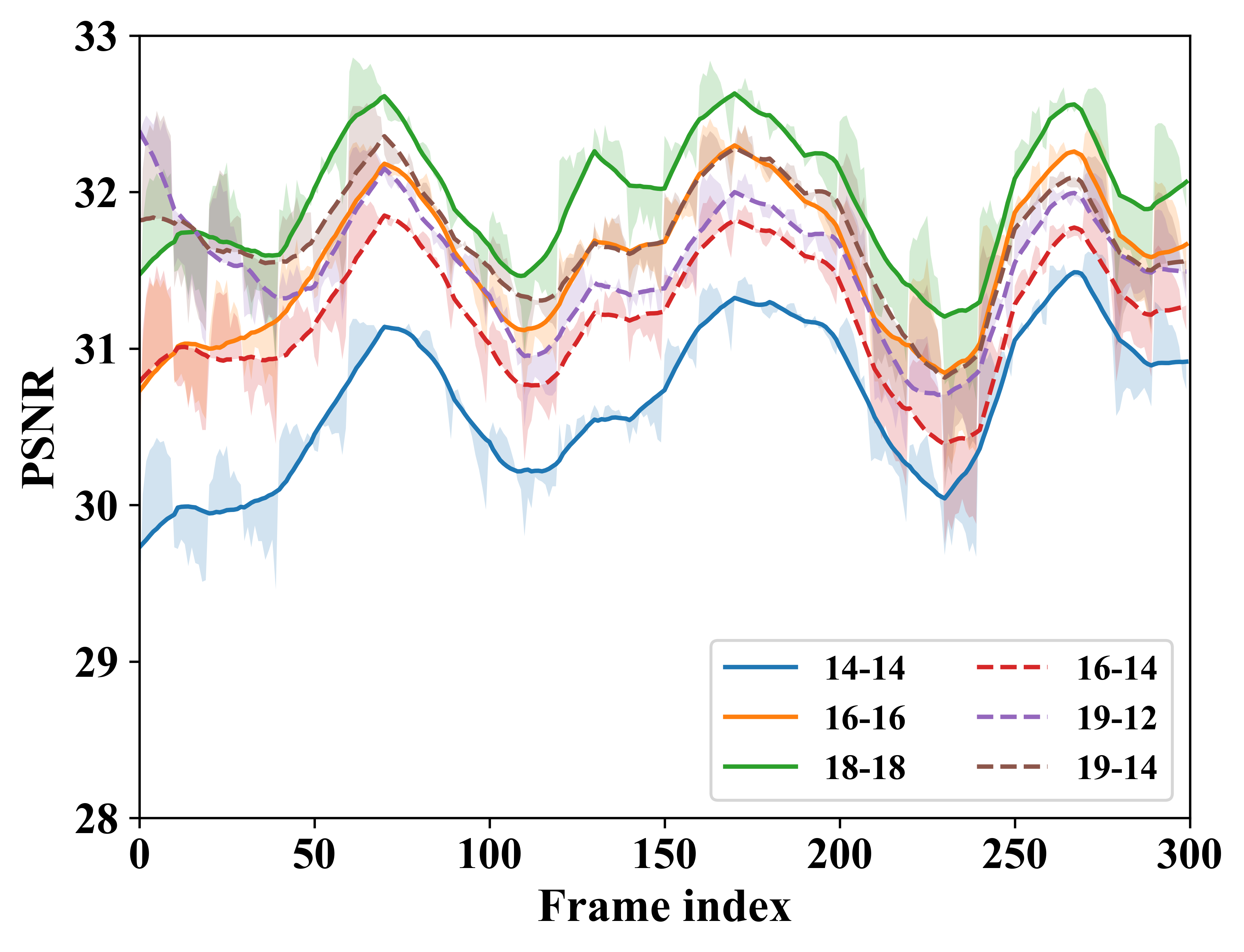}
        \caption{}
        \label{fig:ab_hash_structure_psnr_curve}
    \end{subfigure}
    \caption{Combined PSNR curves for various scenes and methods. (a) The \emph{trimming} scene in the Meetroom dataset. Solid and dotted lines denote offline and online methods respectively. (b)  The first 1200 frames of \emph{seminar} scene in our dataset. The solid lines and dotted lines denote offline and online methods respectively. (c) The first 2400 frames of \emph{seminar} scene in our long video dataset. (d) the \emph{sear steak} scene in the DyNeRF dataset. Please note solid and dotted lines denote symmetric and asymmetric settings, respectively. The CD-NGP/MERF-$P_1$-$P_2$ annotation denotes different hash table sizes in the base and auxiliary branches defined in \cref{method: cd-ngp}. }
    \label{fig:combined_psnr_curves}
\end{figure*}

\Cref{tab:comparison_meetroom_dataset} and \cref{fig:comparison meetroom dataset} show the quantitative and qualitative comparisons on the Meetroom dataset.  \cref{fig:psnr_curve_meetroom_trimming} is the PSNR curve of the compared methods on the \emph{trimming} scene. We use the default setup of StreamRF \cite{streamRF} for the Meetroom dataset to obtain the results. 
As shown in \cref{tab:comparison_meetroom_dataset}, HexPlane-GR and HexPlane-PR produce significantly lower quality than the other methods on the Meetroom dataset, we thus exclude them from \cref{fig:psnr_curve_meetroom_trimming} for better visualization. The metrics in \cref{tab:comparison_meetroom_dataset} and the curves in \cref{fig:psnr_curve_meetroom_trimming} show the advantage of our method over both online and offline baselines. 
We note that the Meetroom dataset provides only 13 views (much fewer than 20+ views in the DyNeRF dataset), which leads to limited coverage for multi-view stereo. As a result, COLMAP \cite{mvs,sfmotion} fails to generate high-quality dense point clouds, making it difficult for 4DGS \cite{4Dgaussiansplatting} to obtain reliable initialization. Consequently, 4DGS does not produce valid results on this dataset and is excluded from the comparison. 
On the contrary, our method does not require initial point clouds and is more robust. It outperforms the implicit MLP-based online method \cite{invnerf} regarding reconstruction quality, speed, and model size. 
It obtains slightly lower metrics than the explicit voxel-based online method \cite{streamRF}, which is likely due to the robustness advantage of explicit voxel representations under sparse-view conditions. 
When compared with offline baselines, our method produces comparable quality to MixVoxels \cite{mixvoxels} and outperforms HexPlane \cite{HexPlane_} in terms of quality and speed by much less memory cost. 

  %-----------figure-------------------

\vspace{0pt}
\begin{table*}[ht]
    \centering
  \caption{Comparisons of the metrics on the Meetroom dataset.}
  \begin{tabular}{l|ccllllr}
    \toprule
    Method        & Online  &  Offline        & PSNR\uaa        & DSSIM\daa    & LPIPS\daa & Time\daa & Size\daa \\ \midrule
    Mixvoxels   &  &  $\checkmark$         & \textbf{26.96}  & \textbf{0.027}   & \textbf{0.103} & \textbf{15min} & 500MB  \\
    HexPlane & & $\checkmark$   & 24.71   & 0.043 & 0.221 & 100min & 200MB  \\ \midrule
    HexPlane-PR      & $\checkmark$ & & 18.63 & 0.112 & 0.359 & 100min & 318MB  \\
    HexPlane-GR      &$\checkmark$ & & 19.49  & 0.099 & 0.400 & 100min & 318MB  \\
    StreamRF   & $\checkmark$       &  & 26.48  & 0.029 & 0.170 & 75min & 1710MB \\
    INV        & $\checkmark$     & & 23.87   & 0.041 & 0.120 & 40hours & 336MB  \\
    CD-NGP (ours)       & $\checkmark$  & & 26.01  & 0.039 & 0.191 & 70min & \textbf{113MB}  \\
  \bottomrule
  \end{tabular}
  \label{tab:comparison_meetroom_dataset}
  \end{table*}

\subsubsection{Spatial representations and scalability analysis. } 
\label{ablation_spatial_representation}
Our proposed continual learning framework is robust against changes in spatial backbones, including spatial hash encoder configurations and feature fusion methods. 
The hash encoder in \cref{hash encoder config} could be changed into the hybrid representation (MERF) \cite{merf} and the plane representation like those in \cite{kplanes,trimiprf}. The MERF representation is defined as: 
 \begin{equation}
  \Psi(\mathbf{x})= \bigoplus_{m=1}^{L} ~ P_m(x,y,z) + Q_{mxy}(x,y) + Q_{myz}(y,z) + Q_{mzx}(z,x) ,
  \label{hash encoder config-merf}
\end{equation}
where $Q_{mij}(i,j)$ denotes the bi-linearly interpolated features from the 4 nearby vertices in the orthogonally projected planes \wrt axes $(i,j)$ at the $m$-th resolution. 
The plane representation is defined as:  
\begin{equation}
  \Psi(\mathbf{x})= \bigoplus_{m=1}^{L} Q_{mxy}(x,y) + Q_{myz}(y,z) + Q_{mzx}(z,x).
  \label{hash encoder config-plane}
\end{equation}
% 再check一下merf和plane的P的大小，说清楚。这里好像没说清楚
For the simplicity of the notations, we denote the models adopting \cref{hash encoder config}, \cref{hash encoder config-merf}, and \cref{hash encoder config-plane} as CD-NGP, CD-MERF, and CD-Plane respectively. 

We compare the results of different spatial representations on the DyNeRF dataset and list the parameter numbers in \cref{tab:combined_ablation_parameter_stats}. The alternatives of spatial representations include CD-MERF and CD-Plane. 
The element-wise product for plane representation in \cite{kplanes} produces unsatisfactory results in the experiments. We thus remove the $P_m(x,y,z)$ in \cref{hash encoder config-merf} and fuse the features $Q_{mij}(i,j)$ by concatenation in the implementation. 
As shown in \cref{tab:combined_ablation_parameter_stats}, the hybrid representation CD-MERF achieves high compression and comparable model sizes to methods for static scenarios \cite{ingp,trimiprf} ($\sim$60MB)  without significant quality loss. 
We also adopt the 3D+4D representation similar to that in \cite{park2023temporal} for comparison. This representation is implemented  by concatenating the spatial $(x,y,z)$ features and spatial-temporal $(x,y,z,t)$ features encoded by a 4D extension of \cref{hash encoder config}, i.e. $\Gamma(x,t) = \bigoplus_{m=1}^{L} P_m(x,y,z,t)$. However, the asymmetric hash sizes (using $(P_1, P_2) = (19, 14)$) in the 3D+4D representation results in serious feature collisions and deliver bad results (PSNR < 16 dB on 3 scenes). We thus set $(P_1, P_2) = (14,14)$ for the 3D+4D representation in \cref{tab:combined_ablation_parameter_stats} to keep the size of the model close to our CD-NGP for fair comparison. The metrics show the advantage of our separated encoding of spatial-temporal $(x,y,z,t)$ features explained in \cref{our_dnerf_sigma,continual_dnerf_sigma}: representing the features of spatial coordinates $(x,y,z)$ and temporal coordinate $(t)$ separately avoids feature collisions and leads to better results.  

The results in \cref{tab:combined_ablation_parameter_stats} also show the flexibility in the feature fusion methods. 
We replace the element-wise sum in \cref{continual_dnerf_sigma,hash encoder config-merf}  with vector concatenation and denote them as CD-NGP-C and CD-MERF-C in \cref{tab:combined_ablation_parameter_stats}. 
Benefiting from larger feature dimensions, these methods lead to slightly better quality (especially when measured in LPIPS) at the cost of more training time, and the reasonable trade-off also shows the robustness of our method. 
  \begin{table*}[!ht]
    \centering
    \caption{Quality metrics, parameter numbers, and bandwidth costs of different spatial representations. The parameter numbers of each module of the 30-branch representations are also shown. $\# 3D Enc.$ and $\# 2D Enc.$ denote the parameter numbers of the 3D encoders and 2D encoders in the base branch and one auxiliary branch accordingly.  $\mathbf{B_{min}}$ and  $\mathbf{B_{avg}}$ are minimum and average bandwidth costs in MB/frame. $\#$ denotes the number of parameters per branch (10 frames). }
    \label{tab:combined_ablation_parameter_stats}
  \begin{tabular}{l|llllr|lllll}
      \toprule
      Model   & Params & {PSNR\uaa}       & {LPIPS\daa}   & {Time\daa} & {Size\daa} & $\#$3D Enc. & $\#$2D Enc. & $\#$Online & $\mathbf{B_{min}}$ & $\mathbf{B_{avg}}$ \\ 
      \midrule
      CD-NGP & 21M & 30.23 & 0.198 & \textbf{75min} & 113MB & $(9.9M,0.37M)$ & - & $0.39M$ & 0.16 & 0.38 \\ 
      3D+4D & 23M & 29.31 & 0.242 & 82min & 122MB & $(0.73M,0.73M)$ & - & $0.76M$ & 0.31 & 0.41 \\ 
      CD-NGP-C & 21M & \textbf{30.52} & \textbf{0.174} & 80min & 114MB & $(9.9M,0.37M)$ & - & $0.40M$ & 0.16 & 0.38 \\ 
      CD-Plane & 14M & 29.16 & 0.215 & 77min & 83MB & - & $(1.9M,0.082M)$ & $0.28M$ & 0.11 & 0.28 \\ 
      CD-MERF & 7.5M & 29.93 & 0.201 & 109min & \textbf{57MB} & $(1.3M,0.049M)$ & $(0.56M,0.023M)$ & $0.16M$ & 0.061 & 0.19 \\ 
      CD-MERF-C & 7.5M & 30.22 & \textbf{0.174} & 115min & \textbf{57MB} & $(1.3M,0.049M)$ & $(0.56M,0.023M)$ & $0.17M$ & 0.065 & 0.19 \\ 
      \bottomrule
    \end{tabular}
\end{table*}
  The parameter numbers and bandwidth consumption in \cref{tab:combined_ablation_parameter_stats} also prove the superior scalability of our method, where \emph{3D Enc.} and \emph{2D Enc.} denote the spatial encodings according to voxel and plane indices respectively. 
  Following MERF \cite{merf} for static scenes,  we use a smaller hash table for the hybrid representation in \cref{hash encoder config-merf} than pure voxel hash grid \cite{ingp} representation in \cref{tab:combined_ablation_parameter_stats} to ensure the compactness and scalability of our method. Denoting the lengths of hash tables along pure voxels in a branch of CD-NGP as $2^{P}$,  the lengths of hash tables in the {CD-MERF} are set as $2^{P-3}$ and $2^{P-4}$ for voxel and plane encoders respectively. The number of the parameters of the MERF representation is reduced to $1-2^{-3}-3\times 2^{-4}=31.25\%$ of the voxel hash grids. Since the hash tables comprise most of the parameters, the model size is reduced from $21M$ to $7.5M$ as listed in \cref{tab:combined_ablation_parameter_stats}. 
  Likewise, we use $2^{P-2}$-length hash tables for the Plane representation in \cref{tab:combined_ablation_parameter_stats} to balance the quality and the model size. The structural configurations $(L, F)$ of the plane representation in \cref{tab:combined_ablation_parameter_stats} are set as $(6,4)$ , and the resolution of the hash encoders of the CD-Plane are set within $(64, 2048)$ for better results. 
  Following \cite{streamRF,nerfplayer,hyperreel,invnerf} we also report the minimum per-frame bandwidth cost $\mathbf{B_{min}}$, which only depends on the parameter numbers of the auxiliary branches and is thus significantly lower than the average bandwidth consumption $\mathbf{B_{avg}}$. Since the base encoder and the density grid could be shared among different branches and transmitted only once in the beginning, the minimum bandwidth consumption is reduced to $0.065\sim 0.16$ MB per frame, which is $100\times$ less than the voxel-based StreamRF \cite{streamRF} and $7\times$ less than the pure MLP representation  INV \cite{invnerf}. 

\subsubsection{Long sequence dataset and scalability analysis.}
As detailed in \cref{method: cd-ngp}, our method exhibits superior scalability. The overall model size is $\mathcal{O}(N_{chunk} \cdot 2^{P_2}LF + 2^{P_1}LF)$, where $N_{chunk}$ denotes the number of chunks in the sequence. We provide quantitative comparisons on our proposed long multi-view video dataset in \cref{tab: comparison long video} and \cref{fig:PSNR_curve_long_video_sequence}. 
Following the default configurations on the DyNeRF dataset, the offline baselines HexPlane \cite{HexPlane_} and MixVoxels \cite{mixvoxels} are trained with a chunk length of $T_{chunk} = 300$. Consistent with its DyNeRF configuration, 4DGS \cite{4Dgaussiansplatting} adopts a lazy loading strategy, caching frames to disk in order to mitigate memory usage at the expense of training speed. 
 In CD-NGP, we set the hash table size in the auxiliary branches to $2^{12}$ and adopt a compact chunk length of $T_{chunk} = 5$. As shown in \cref{tab: comparison long video}, our method reduces storage consumption by at least $4\times$, delivers rendering quality comparable to HexPlane \cite{HexPlane_}, and maintains low memory usage and fast training speed. Although 4DGS produces better results than our method, it consumes approximately 100 times more storage space.  
 %和300帧对比，我们的优势更大，因为我们的可扩展性更好。
Compared with the results in \cref{tab:comparison_dynerf_dataset}, the performance gap in storage consumption is even larger in long video sequences. This is because the majority of the parameters in our method are used for the auxiliary branches, and the base branch only takes up a small fraction of the parameters. Consequently, applying smaller hash tables in the auxiliary branches and leveraging the reuse of the features in the base branch lead to a significant reduction in the model size. 
  
\begin{figure*}[!ht]
    \centering
    \captionsetup{skip=0pt}
    \captionsetup[sub]{font=normalsize}  % 设置subcaption的字体大小
    \begin{subfigure}[b]{0.32\linewidth}
        \centering
        \includegraphics[width=\linewidth]{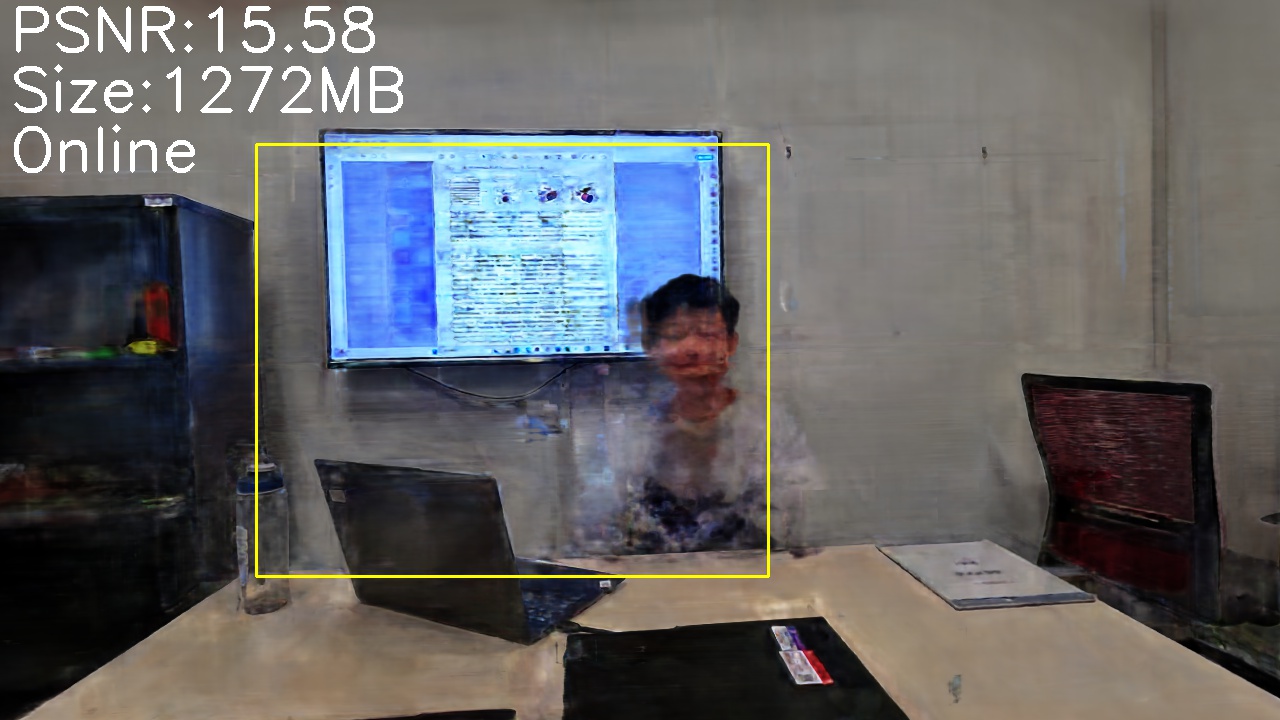}
        \caption{HexPlane-PR}
        \label{fig:cmp_HexPlane-PR_long}
    \end{subfigure}
    \begin{subfigure}[b]{0.32\linewidth}
        \centering
        \includegraphics[width=\linewidth]{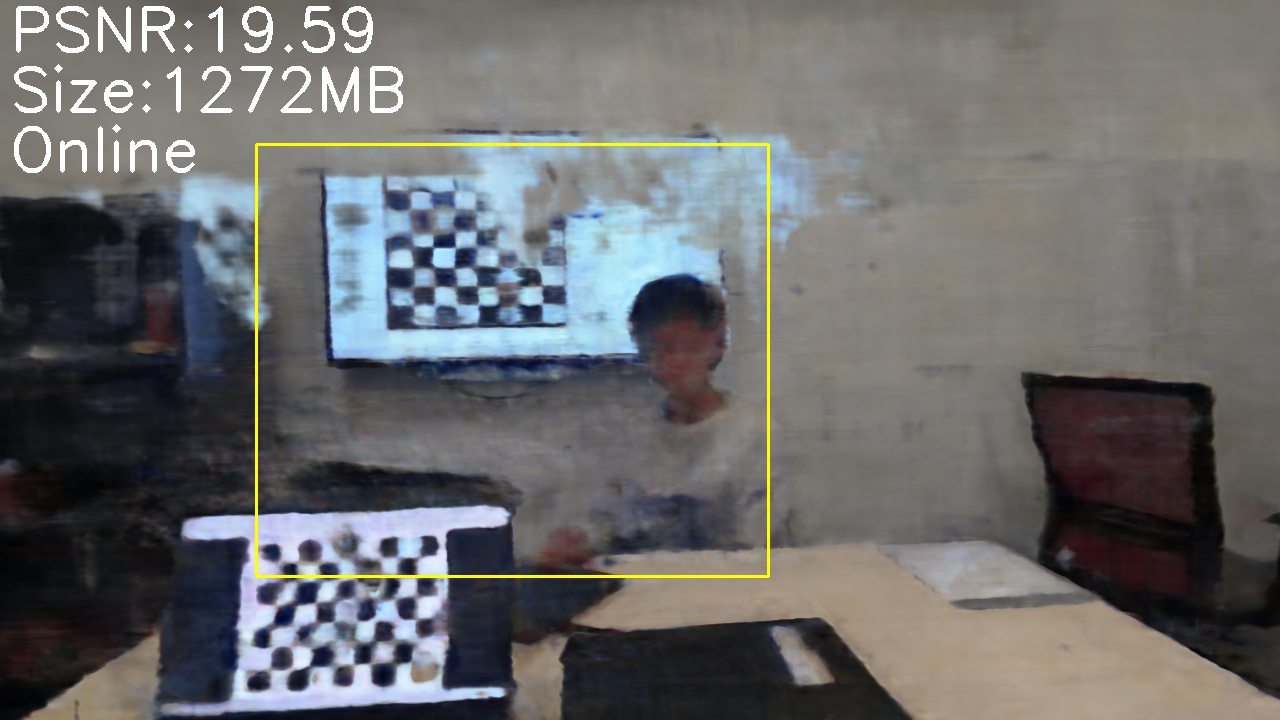}
        \caption{HexPlane-GR}
        \label{fig:cmp_HexPlane-GR_long}
    \end{subfigure}
    \begin{subfigure}[b]{0.32\linewidth}
        \centering
        \includegraphics[width=\linewidth]{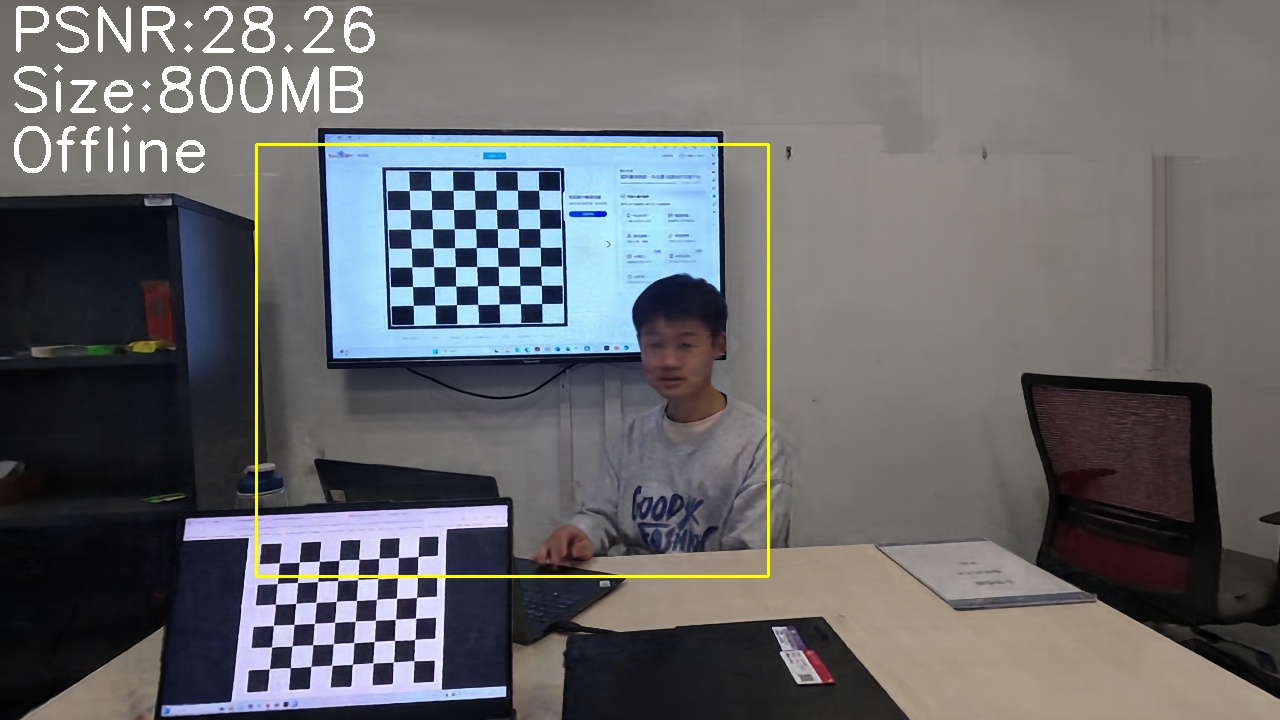}
        \caption{HexPlane}
        \label{fig:HexPlane-offline}
    \end{subfigure}
    \begin{subfigure}[b]{0.32\linewidth}
        \centering
        \includegraphics[width=\linewidth]{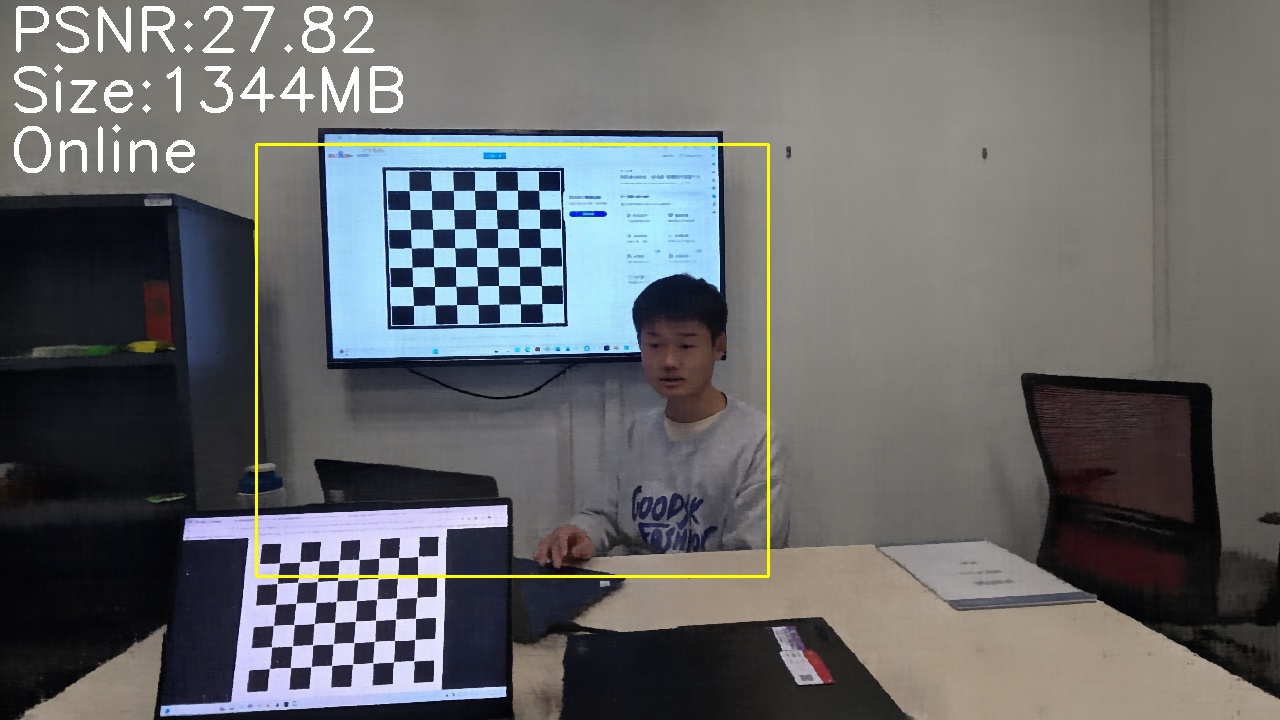}
        \caption{INV}
        \label{fig:cmp_inv_long}
    \end{subfigure}
    \begin{subfigure}[b]{0.32\linewidth}
        \centering
        \includegraphics[width=\linewidth]{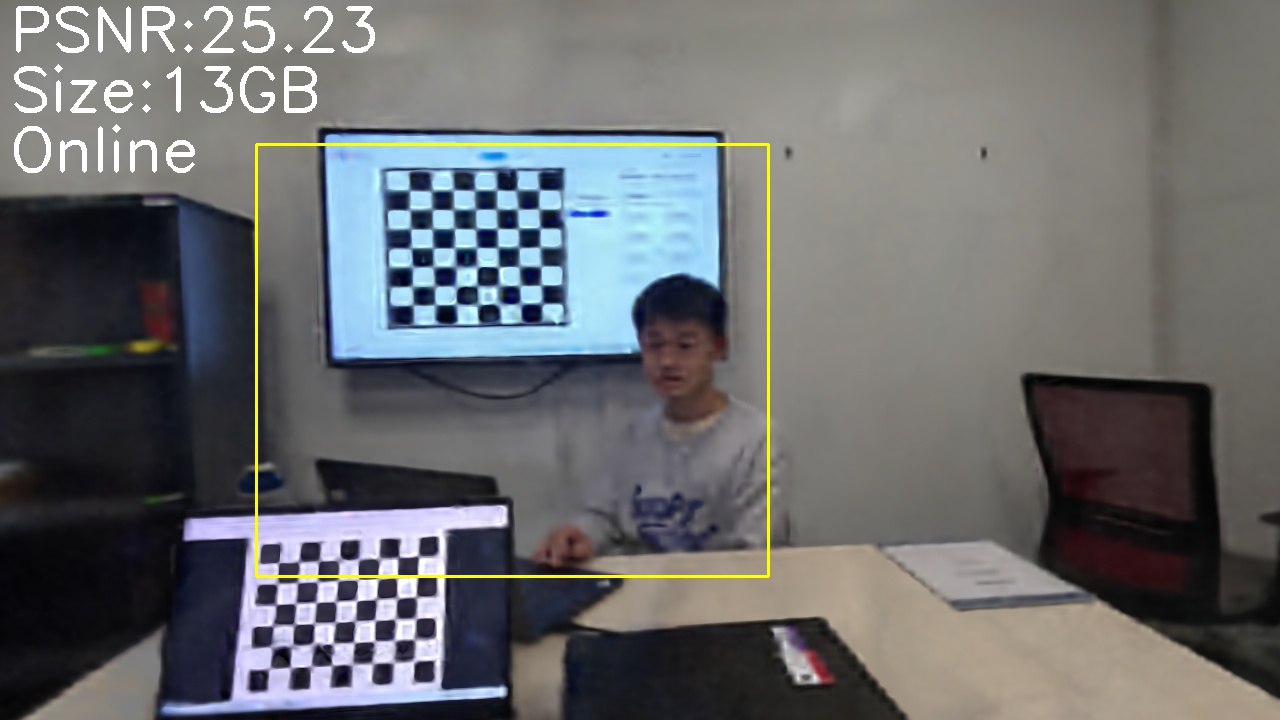}
        \caption{StreamRF}
        \label{fig:cmp_streamrf_long}
    \end{subfigure}
    \begin{subfigure}[b]{0.32\linewidth}
        \centering
        \includegraphics[width=\linewidth]{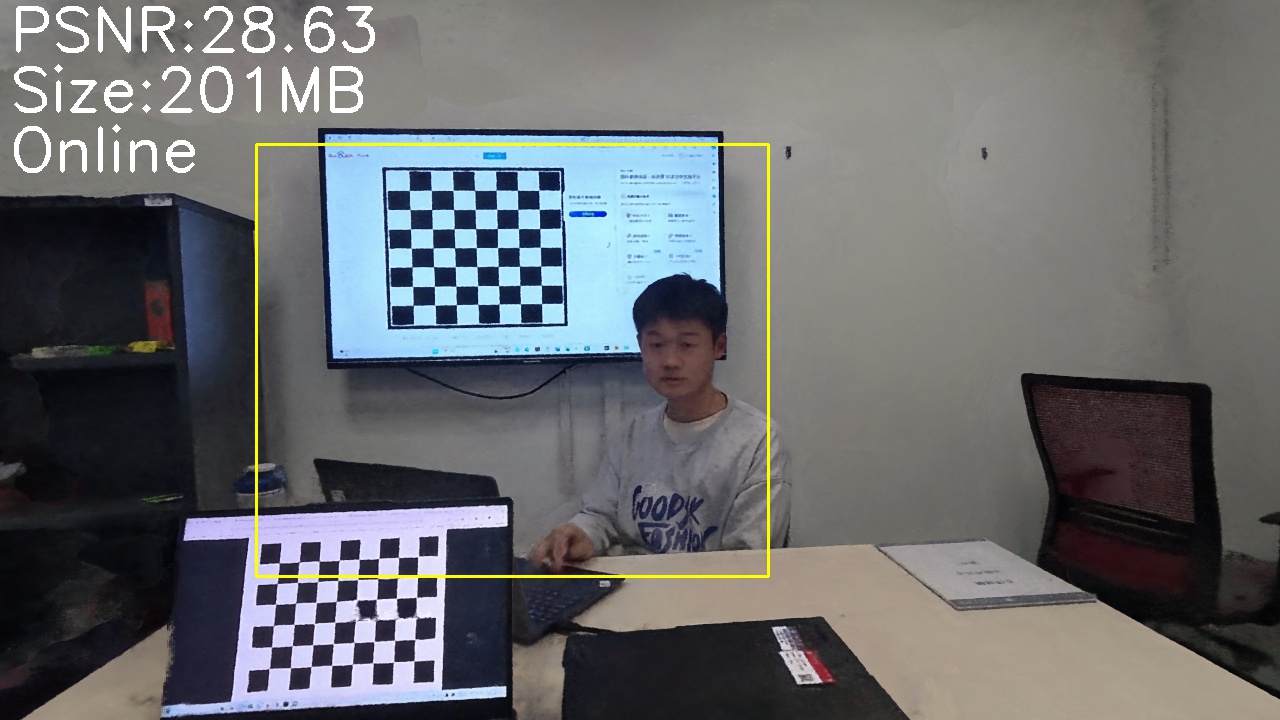}
        \caption{CD-NGP (ours)}
        \label{fig:cmp_CD-NGP_long}
    \end{subfigure}
    \caption{Comparison of reconstruction quality on the 19th frame of the \emph{seminar} scene in our long video dataset.}
    \label{fig:comparison_long_video_dataset}
\end{figure*}

  The PSNR curves in \cref{fig:PSNR_curve_long_video_sequence} further demonstrate the stability of our method. The compared regularization-based and replay-based methods suffer from \emph{catastrophic forgetting}: the PSNRs at the beginning of the chunks are much lower than those in the end. 
  The problem is properly eliminated by our method through parameter isolation, and our method continuously produces high-quality results for the whole sequence.   We also provide comparisons between the rendered results in \cref{fig:comparison_long_video_dataset}. 
    The regularization method displayed in \cref{fig:cmp_HexPlane-PR_long} reconstructs the TV screen in the background erroneously. 
    The replay method displayed in \cref{fig:cmp_HexPlane-GR_long} fails to recover the details. 
    Ours in \cref{fig:cmp_CD-NGP_long} produces comparable quality to the offline baseline in \cref{fig:HexPlane-offline}.

\begin{figure}[!htbp]
      \includegraphics[width=\linewidth]{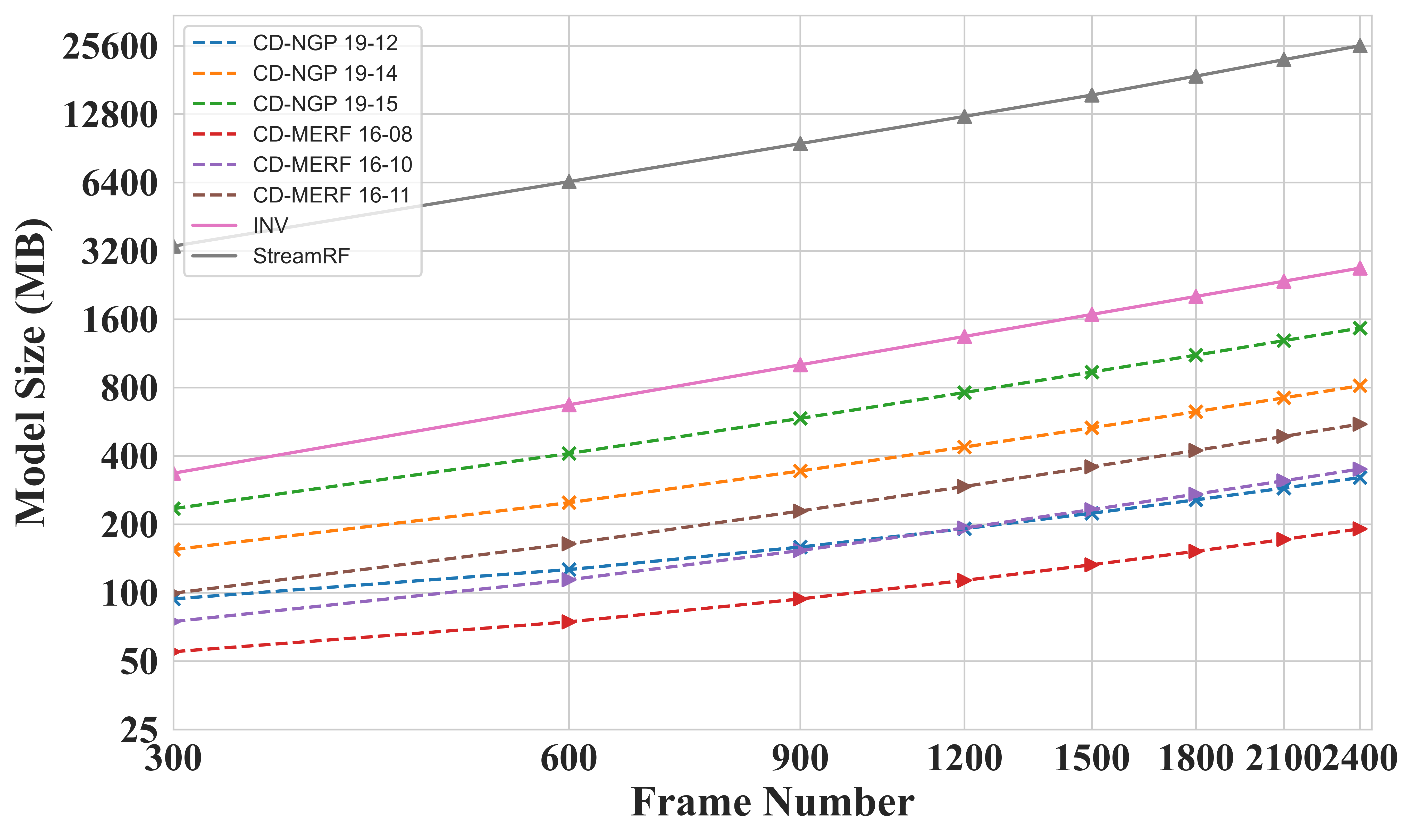}
        \caption{Comparison of the model sizes of our method and other online baselines. Our method shows superior scalability over others. } 
        \label{fig:2400frame model size comparison}
  \end{figure}

To further demonstrate the scalability of our method, we conduct experiments on the first 2400 frames of the \emph{seminar} scene in our long video dataset. We further show the PSNR curves in \cref{fig:PSNR_curve_2400_frame_long_video_sequence} and compare the model sizes in \cref{fig:2400frame model size comparison}. 
Our method produces high-quality results while consuming only a fraction of storage space compared with the other online baselines.

\subsection{Ablation studies}
  %----------------figure-------------------

  \subsubsection{Lengths of hash tables.} 
  
  Benefiting from the multi-branch hash table design,
   our model is highly efficient and configurable. 
   We change the lengths of the hash tables of the initial branch and the auxiliary branches and obtain the results on the DyNeRF dataset in \cref{tab:ablation hash size}. $(P_1, P_2)$ are the \emph{log2} values of the hash table lengths of the base branch and the auxiliary branches. 
   The whole model contains hash encoders, MLP regressors, and the occupancy grid, the sizes of the models are thus larger than the value deduced from the combinations of the hash table and MLP regressors. 
 Although larger hash tables lead to fewer hash collisions, the default configuration $(19,14)$ in CD-NGP produces better results than the naive $(16,16)$ configuration even if the storage consumption is $40.8\%$ lower. 
  Moreover, the $(19,12)$ configuration of CD-NGP significantly outperforms the naive $(14,14)$ configuration at a similar storage cost. 
  To further demonstrate the superiority of the asymmetric hash table design and prove the scalability of our method, we provide the PSNR curves of different hash table designs on the \emph{sear steak} scene in  \cref{fig:ab_hash_structure_psnr_curve}.  
   Equipped with a larger hash table in the base branch, the asymmetric setting $(19,14)$ and $(19,12)$ outperforms others in the beginning and produces comparable metrics to the naive $(16,16)$, $(14,14)$ configuration respectively in the subsequent chunks. 
   The $(18, 18)$ in \cref{tab:ablation hash size} configuration delivers the best results (30.55 dB in PSNR) on the DyNeRF dataset, but at the cost of an exceptionally large model size (638MB). However, the CD-NGP-C configuration with $(P_1, P_2) = (19, 14)$ listed in \cref{tab:combined_ablation_parameter_stats} achieves a comparable PSNR (30.52 dB) with a much smaller model size (114MB, $80\%$ compression). 
   The advantage comes from our re-use of the spatial features encoded by different branches: the base branch with a large hash table recovers more environment details, and the auxiliary branches could thus use fewer parameters to cope with recent changes only. The combination of the base encoder and the auxiliary encoders leads to a more efficient and scalable model.

\begin{table*}[!ht]
  \centering
      \caption{Quantitative comparisons between our CD-NGP and baselines on the long video sequences in our dataset. Metrics are measured on the first 1200 frames and averaged among the different scenes. The compared HexPlane baselines are trained with a chunk size of 300 frames. }
      \begin{tabular}{l|llllllll}
        \toprule
        Method   &   Online       &  Offline             & PSNR\uaa & DSSIM\daa       & LPIPS\daa    & Size\daa  & Time\daa   & Memory\daa \\ \midrule
        MixVoxels  &   & $\checkmark$ & 26.94 & {0.022}  & 0.118 & 2000MB & \textbf{80 min} & 80GB   \\ 
        HexPlane  &   & $\checkmark$ & {27.74} & {0.022}        & 0.117 & 800MB & 8 hours & 74GB   \\ 
        4DGS  &   & $\checkmark$ & \textbf{28.85} & \textbf{0.019}        & \textbf{0.078} & 19GB & 24 hours & \textbf{3GB}   \\ \midrule 
        HexPlane-PR  & $\checkmark$  &      & 20.90 & 0.095  & 0.286 & 1272MB & 8 hours & {12GB}   \\
        HexPlane-GR  & $\checkmark$  &     & 24.82 & 0.052 & 0.282   & 1272MB & 8 hours & {12GB}   \\
        StreamRF  & $\checkmark$  &    & 24.72  & 0.036  & 0.273 & 13GB & 4 hours  & {12GB}   \\
        INV  & $\checkmark$  &    & 27.12  & 0.025 & 0.086 & 1344MB & 160 hours & {12GB}   \\
        CD-NGP (ours)  & $\checkmark$  &     & 27.15  & 0.039 & 0.188 & \textbf{201MB} & 10 hours  & {12GB}   \\\bottomrule 
      \end{tabular}
      \label{tab: comparison long video}
\end{table*}

\begin{table}[!ht]
\centering
      \caption{Ablation study on hash table sizes.   $^{\dag}$ denotes the default setting.}
      \begin{tabular}{l|lrrrrrr} 
        \toprule
        $(P_1, P_2)$      & Params &   PSNR\uaa   & DSSIM\daa & LPIPS\daa   & Size\daa \\ \midrule
        $(18,18)$ & 151M   & \textbf{30.55} & \textbf{0.024}  & \textbf{0.161}  & 638MB\\
        $(16,16)$ & 41M    & 30.08 & 0.027  & 0.194 & 191MB\\
        $(14,14)$ & 12M    & 29.32 & 0.035  & 0.248  & \textbf{75MB}\\
        $(19,14)^{\dag}$ & 21M & 30.23 & 0.027  & 0.198 & 113MB\\
        $(16,14)$ & 13M    & 29.78     &  0.030    & 0.228  & 79MB\\
        $(19,12)$ & 14M    & 30.09     & 0.028    & 0.209 & 82MB\\
        \bottomrule
      \end{tabular}
      \label{tab:ablation hash size}
  \end{table}

\subsubsection{Temporal representations.}
% T-Code和Freq encoding, Freq + MLP的对比

Our continual learning pipeline is also compatible with a variety of representations for time. 
Currently, there are 2 other prevailing representations for the time axis alone: pure frequency encoding \cite{Dnerf} and MLP with frequency-encoding \cite{tineuvox}. 
We replace the temporal hash encoding in our method with them on the DyNeRF dataset and show the results in \cref{tab:ablation temporal representation}. 
Our time latent code slightly outperforms others on DyNeRF dataset. The robustness against changes in the temporal representations also shows the viability of the continual learning method. 
  \begin{table}[!ht]
    \centering
    \caption{Ablation study on temporal representations.}
    \begin{tabular}{l|lllll}
      \toprule
      Method  & {PSNR\uaa}   & {DSSIM\daa}     & {LPIPS\daa}  & \\ 
      \midrule
      CD-NGP      & \textbf{30.23} & \textbf{0.0269} & \textbf{0.198}           \\ 
      Time-MLP & 30.20  & 0.0273 & 0.202           \\ 
      Time-Freq   & 30.19  & 0.0272 & 0.203         \\ 
      \bottomrule
    \end{tabular}
    \label{tab:ablation temporal representation}
  \end{table}
\begin{table}[!ht]
  \centering
    \caption{Ablation study on the field composition methods.}
    \begin{tabular}{l|llll}
      \toprule
      Method  & {PSNR\uaa}   & {DSSIM\daa}     & {LPIPS\daa}   \\ 
      \midrule
      CD-NGP      & \textbf{30.23} & \textbf{0.0269} & \textbf{0.198}           \\ 
      Full composition & 29.24  & 0.0367 & 0.228        \\ 
      Static-dynamic   & 28.27  & 0.0568  & 0.291        \\ 
      \bottomrule
    \end{tabular}
    \label{tab:ablation field composition}
\end{table}

\subsubsection{Field composition methods.}
\begin{figure}[!ht]
  \centering
  \captionsetup{skip=0pt}
  \captionsetup[sub]{font=normalsize}  % 设置subcaption的字体大小和间距
  \begin{subfigure}[b]{0.48\linewidth}
      \centering
      \includegraphics[width=\linewidth]{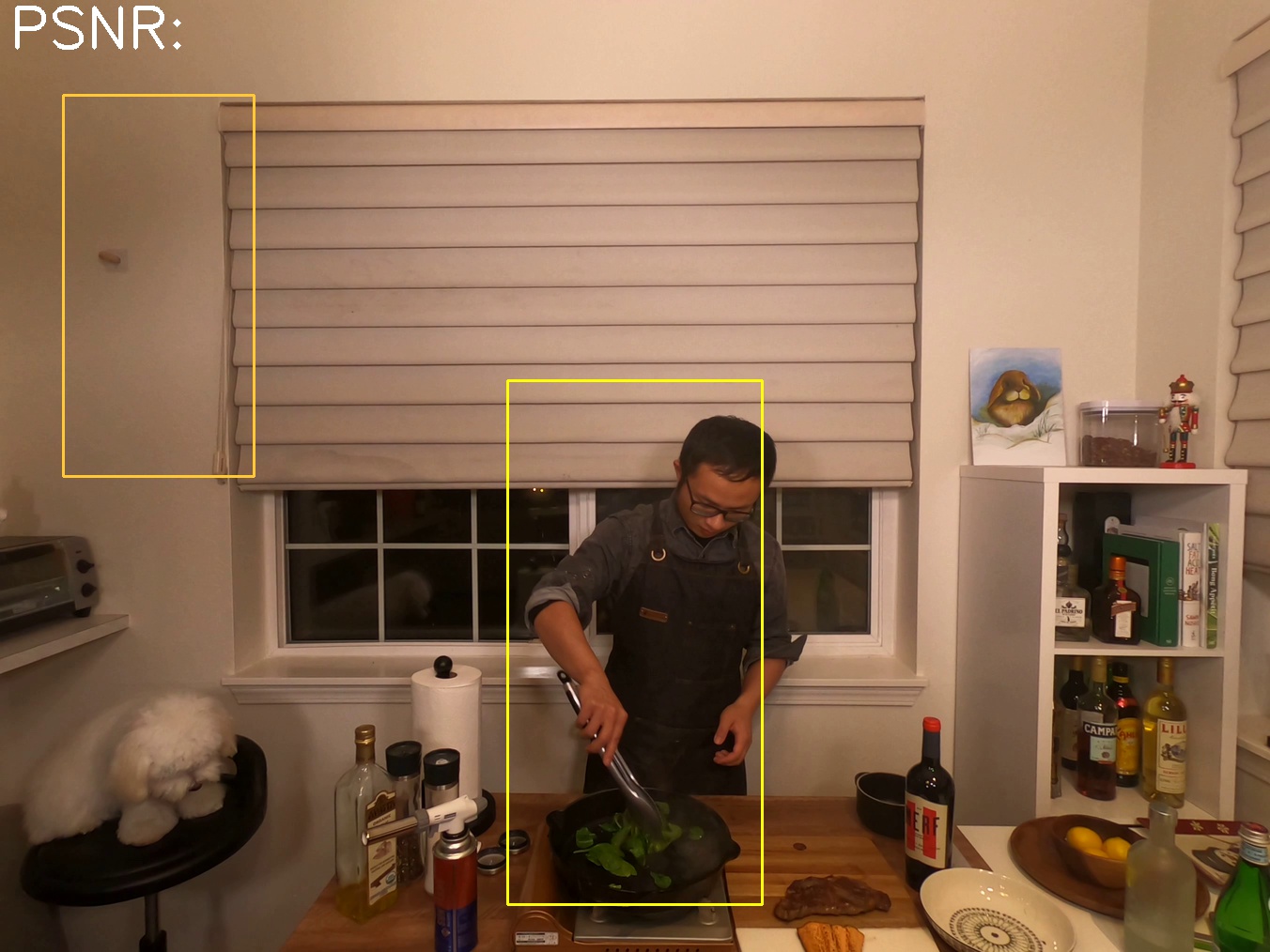}
      \caption{Ground truth}
      \label{fig:ab_field_comp_gt}
  \end{subfigure}
  \begin{subfigure}[b]{0.48\linewidth}
      \centering
      \includegraphics[width=\linewidth]{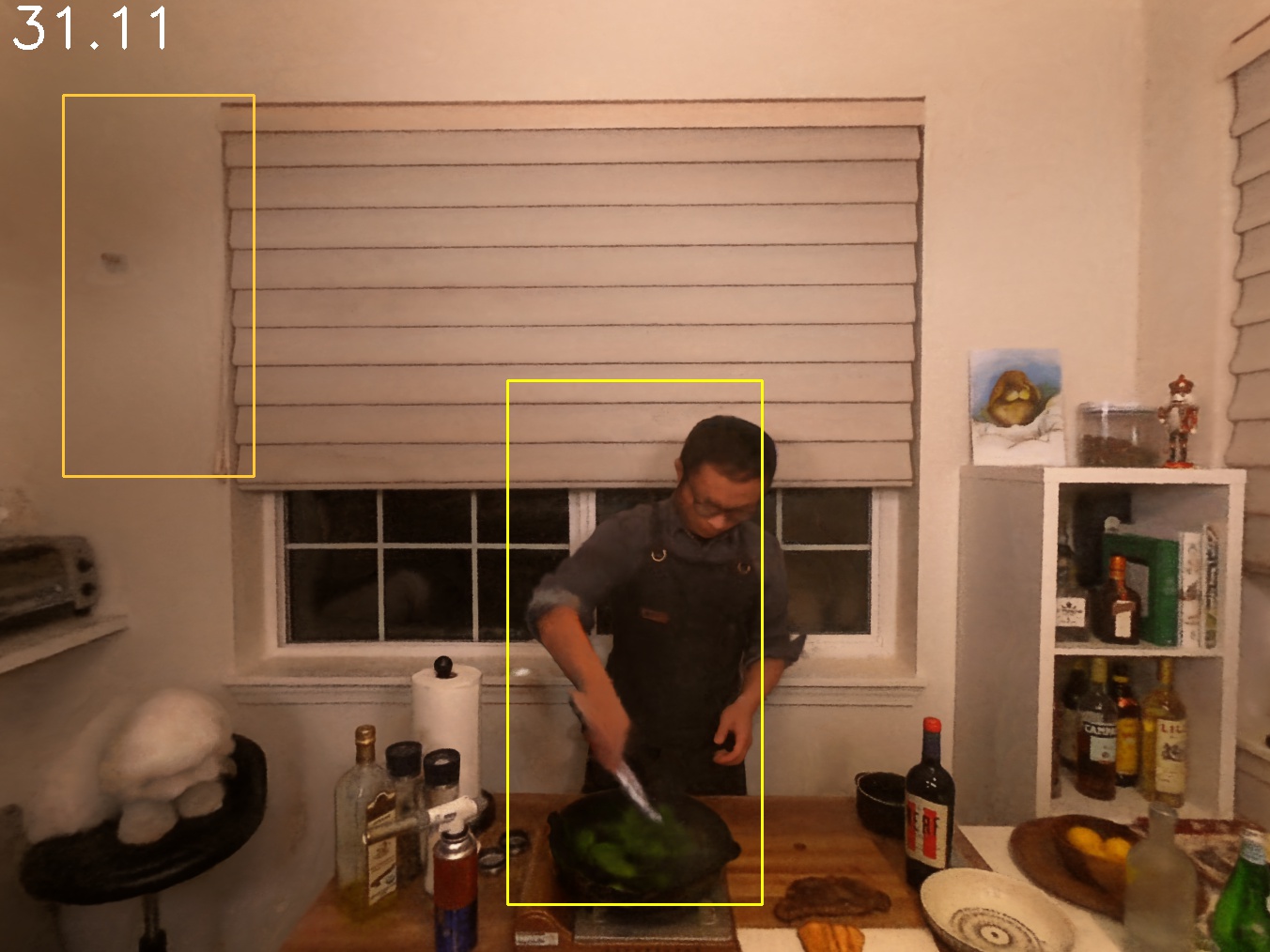}
      \caption{CD-NGP (ours)}
      \label{fig:ab_field_comp_CD-NGP}
  \end{subfigure}
  \begin{subfigure}[b]{0.48\linewidth}
      \centering
      \includegraphics[width=\linewidth]{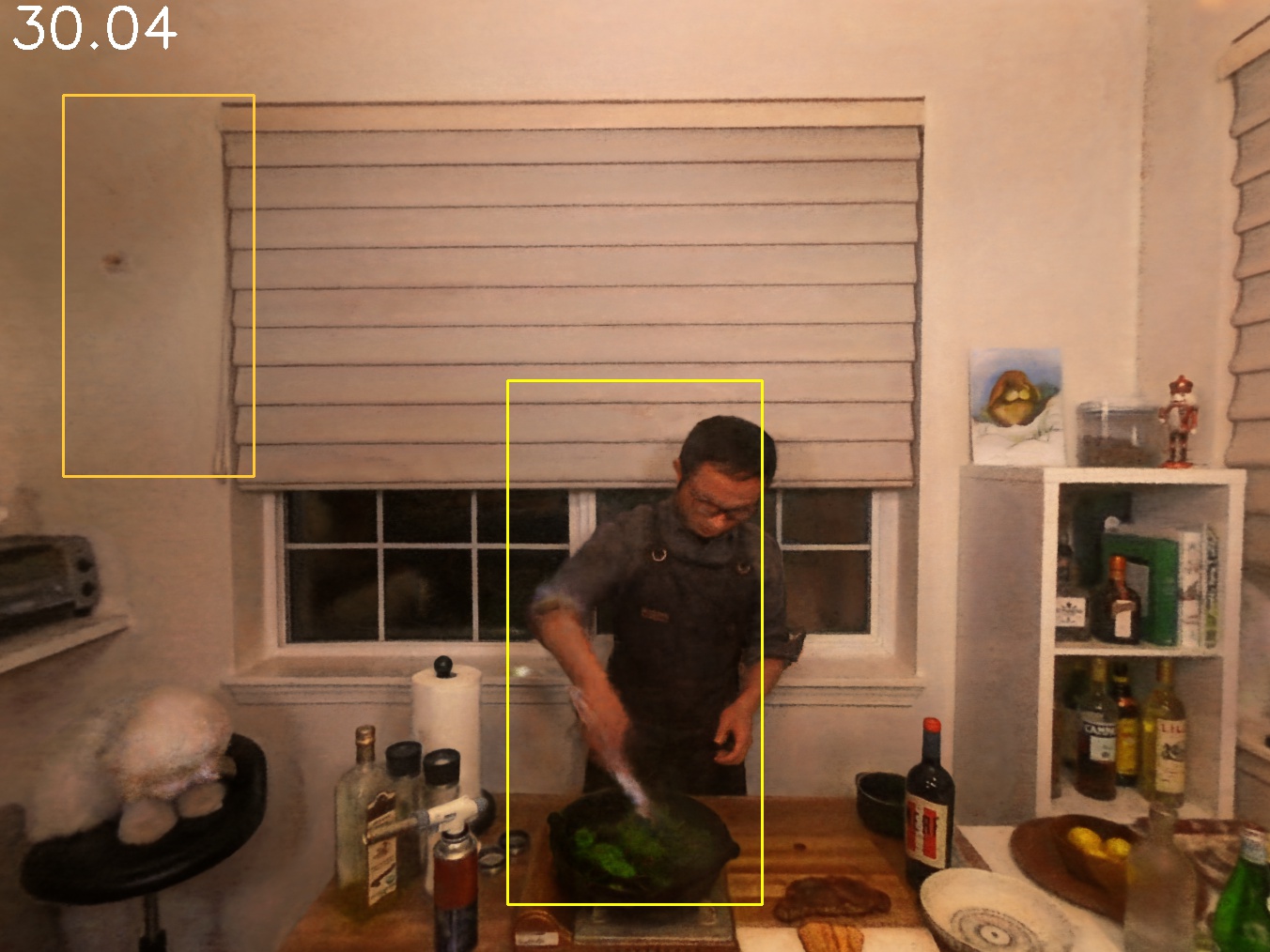}
      \caption{Full composition}
      \label{fig:ab_field_comp_full_composition}
  \end{subfigure}
  \begin{subfigure}[b]{0.48\linewidth}
      \centering
      \includegraphics[width=\linewidth]{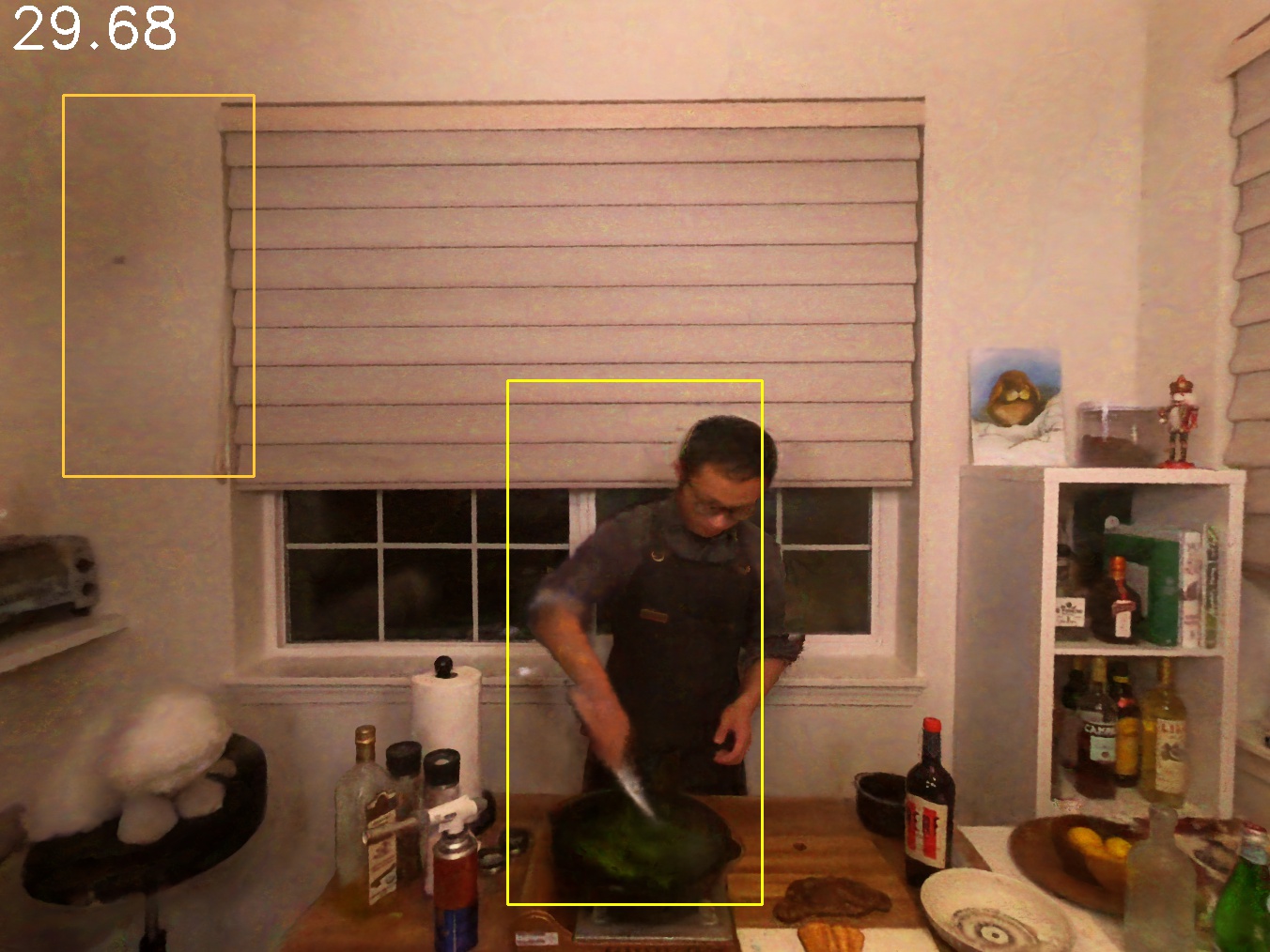}
      \caption{Static-dynamic}
      \label{fig:ab_field_comp_static_dynamic}
  \end{subfigure}
  \caption{Comparison of different field composition methods.}
  \label{fig:ablation_field_composition}
\end{figure}

% T-Code和Freq encoding, Freq + MLP的对比
We also conduct ablation studies on the field composition methods and show the results in \cref{tab:ablation field composition,fig:ablation_field_composition}. The field composition process is defined as:
%---------------------------Equation-----------------------
\begin{equation}
  (\sigma,\mathbf{c})= (\sigma_i + \sigma_a, \mathbf{c}_i + \mathbf{c}_a).
  \label{eq:field composition}
\end{equation}
%-----------------------------------------------------------
\noindent \emph{Full composition} in \cref{tab:ablation field composition} denotes the naive composition of two fields from the base branch and the current branch respectively, \ie $\sigma_i$ and $\mathbf{c}_i$ are the field from the base branch and $\sigma_a$ and $\mathbf{c}_a$ in \cref{eq:field composition} are the field from the current branch. Both branches are encoded with the same architecture as \cref{our_dnerf_sigma}. 
\emph{Static-dynamic} in \cref{tab:ablation field composition} denotes a naive composition of fields from a static branch by the base encoder and a dynamic branch by the auxiliary encoders similar to \cite{mixvoxels}, \ie $\sigma_i$ and $\mathbf{c}_i$ in \cref{eq:field composition} are encoded without the $\Gamma(t)$ in \cref{our_dnerf_sigma}. 
The online training strategies remain the same. Our CD-NGP outperforms the other methods in all metrics.
  Furthermore, as compared in \cref{fig:ablation_field_composition}, the compared composition methods in \cref{fig:ab_field_comp_full_composition,fig:ab_field_comp_static_dynamic} suffer from salt and pepper noise because of the explicit composition, while our CD-NGP circumvents the problem by fusing features in the latent space and produces better results. Moreover, the feature combination 
  in the latent space speeds up the inference by reducing the number of MLP queries. Please refer to the supplementary materials for more details.

%--------------------------------------------- Figure ------------------------------------

%重新排版
\begin{table}[!ht]
  \centering
  \caption{Ablation study on the chunk sizes. The iteration numbers of each chunk are held constant under different $T_{chunk}$ settings.}
  \begin{tabular}{l|lllr}
    \toprule
    $T_{chunk}$  & {PSNR\uaa}   & {DSSIM\daa}     & {LPIPS\daa}  & {Size\daa}   \\ 
    \midrule
    30      & 30.31 & 0.02654 & 0.1842  & \textbf{80MB}            \\ 
    15 & 30.54  & 0.02557 & 0.1816     & 97MB\\ 
    10   & 30.76  & 0.02440 & 0.1793   & 113MB\\ 
    5   & \textbf{30.99}  & \textbf{0.02268} & \textbf{0.1674}  & 162MB        \\ 
    \bottomrule
  \end{tabular}
  \label{tab:ablation: chunk config-hold chunk iter}
\end{table}
\begin{table}[!ht]
  \centering
  \caption{Ablation study on the chunk sizes. The total iterations of the whole 300-frame video are held constant under different $T_{chunk}$ settings.}
  \begin{tabular}{l|lllr}
    \toprule
    $T_{chunk}$  & {PSNR\uaa}   & {DSSIM\daa}     & {LPIPS\daa}  & {Size\daa}   \\ 
    \midrule
    30      & \textbf{30.84} & \textbf{0.02392} & \textbf{0.1505}  & \textbf{80MB}         \\ 
    15 & 30.76  & 0.02413 & 0.1684 & 97MB       \\ 
    10   & 30.76  & 0.02440  & 0.1793 &113MB       \\ 
    5   & 30.39  & 0.02602  & 0.2075  &162MB     \\ 
    \bottomrule
  \end{tabular}
  \label{tab:ablation: chunk config-hold total iter}
\end{table}

\begin{table}[!htbp]
  \centering
  \caption{Comparison between CD-NGP and MVSGS-FT on DyNeRF dataset.}
  \begin{tabular}{l|ccc}
    \toprule
    Model     & PSNR$\uparrow$ & DSSIM$\downarrow$ & LPIPS$\downarrow$ \\
    \midrule
    CD-NGP    & \textbf{30.23}          & \textbf{0.027}             & \textbf{0.198}             \\
    MVSGS-FT  & 26.05          & 0.086             & 0.258             \\
    \bottomrule
  \end{tabular}
  \label{tab:metrics_avg_mvsgs_summary}
\end{table}

%---------------------------------------------------------------------------------

\subsubsection{Chunk size.}

We also conduct ablation studies on the chunk sizes and show the results in \cref{tab:ablation: chunk config-hold chunk iter,tab:ablation: chunk config-hold total iter}. 
We only report the results on the \emph{cook spinach} scene in the DyNeRF dataset for brevity.
  As in \cref{tab:ablation: chunk config-hold chunk iter}, the smaller chunk size leads to better results when controlling the iteration numbers of each chunk. 
  When the total iterations of the whole 300-frame sequence are held constant, the larger chunk size leads to better results as in \cref{tab:ablation: chunk config-hold total iter}. The selection of the chunk size is a trade-off between the quality, robustness, training time, and host memory occupation and should depend on the data,   and our method is robust within a wide range of chunk sizes. We select $T_{chunk}=10$ as the default configuration.

    % 审稿人2要求nuanced discussion
\subsection{Further Discussion}
CD-NGP leverages a highly scalable continual learning framework to tackle dynamic-scene NVS, with a principal benefit being a substantially reduced memory footprint (the feasible 14GB usage on the DyNeRF dataset). The multi-branch feature reusing design leads to fast convergence and high fidelity, and the encoder-decoder structure inherited from neural radiance fields (NeRF) \cite{nerfeccv} enables the model to represent the scenes compactly. CD-NGP supports varied kinds of encoders and field composition methods and exhibits high scalability in long video sequences. However, the volume rendering process is still required, which limits the real-time rendering capability of our method. Please see the supplementary materials for detailed latency analysis. Moreover, the current parameter isolation-based design does not support fusing features from different model branches, and thus the reconstruction quality of our method is lower than the state-of-the-art offline baselines. In addition, due to the hybrid hash encoder representations, our method suffers from more quality downgrades than the explicit baseline \cite{streamRF} in sparse-view scenarios.  

We acknowledge that several recent cost volume-based multi-view stereo (MVS) NVS techniques also support real-time processing with low memory requirements and thus hold promise for dynamic-scene applications. To further evaluate the unique value of continual learning in this context, we include the state-of-the-art method MVSGS \cite{mvsgs} as a representative baseline. MVSGS \cite{mvsgs} leverages a pre-trained model to directly render novel views, as well as generating Gaussian point clouds to support per-scene fine-tuning for best quality.  
We refer to the direct-inference results as MVSGS-D and the fine-tuning results as MVSGS-FT. We note that the cost volume module of MVSGS does not support the standard $1352\times 1014$ resolution in the DyNeRF dataset, we thus use the default setup (including resolution) for MVSGS-D. Although processing the frames at $\sim 1$ FPS,  MVSGS-D produces limited reconstruction quality on the DyNeRF dataset.  
We further compare MVSGS-FT and our CD-NGP in \cref{tab:metrics_avg_mvsgs_summary}. We train the Gaussians for 5k iterations for each frame in the DyNeRF dataset. The training takes around 1.5 minutes for each frame, i.e. $\sim8$ hours for the whole 300-frame sequence. The checkpoints consume around 60MB per frame (18 GB in total), which is significantly larger than the most recent explicit offline baselines for dynamic scenes.  
By comparing the quality metrics in \cref{tab:metrics_avg_mvsgs_summary} and the resource consumption, it is clear that our CD-NGP is more efficient: ours provides high-quality results, but consumes much less model size (113MB in total) and training time (75 minutes) by leveraging the re-use of the spatial features in the first branch. Hence, the significance of continual learning in dynamic-scene NVS is further validated. Please see the supplementary materials for details.

\section{Conclusion and Limitation}
We propose CD-NGP, a continual learning framework for multi-view dynamic scenes based on parameter isolation, along with a self-collected dataset for novel view synthesis under multi-view long-view scenarios. Our approach partitions the input multi-view videos into different chunks and uses individual model branches to reconstruct the scene accordingly. Each branch encodes spatial and temporal features separately using the multi-resolution hash encoding structure. CD-NGP leverages the redundancy of the dynamic scenes and reuses the spatial features in the first branch to achieve fast convergence, high fidelity, and superior scalability. On the prevailing DyNeRF dataset, it consumes much less memory, achieves comparable quality with a smaller model size than most recent explicit offline baselines, and maintains the speed advantage over traditional pure MLP-based offline methods. 
 Compared with online baselines, it delivers state-of-the-art quality and achieves a new balance among reconstruction quality, training speed, bandwidth occupation, and memory cost. On the long video dataset, our method delivers high quality and achieves fast convergence with only a fraction of the storage cost. 
However, like most NeRF-based methods, it still requires querying the field function and is not suitable for real-time rendering. Further efforts could be made to apply baking methods or Gaussian point clouds to continual learning, enabling real-time rendering in long video sequences for wide applications. 

\bibliographystyle{abbrv-doi-hyperref}
\bibliography{main}

%-------------------------biography 

\newcommand{\biophotowidth}{1in}
\newcommand{\biophotoheight}{1in}

\newcommand{\wfwid}{0.9in}
\begin{wrapfigure}{l}{\wfwid} 
  \vspace{-20pt}
  \includegraphics[width=\biophotowidth,height=\biophotoheight,clip,keepaspectratio]{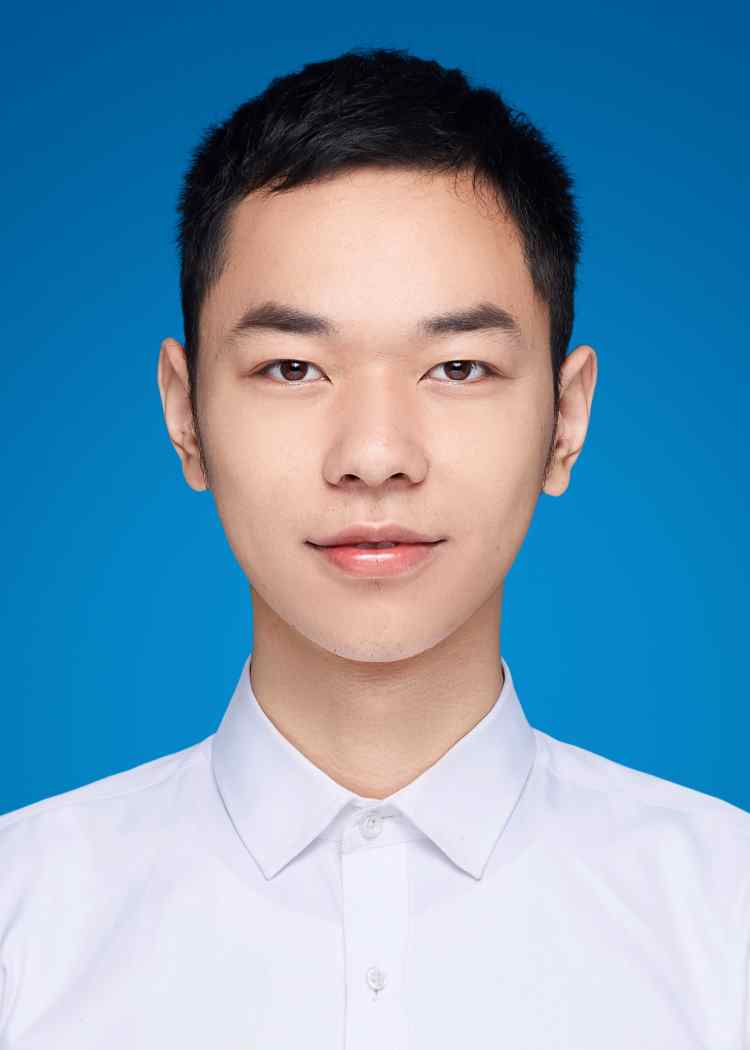}
  \vspace{-10pt}
\end{wrapfigure} 
\textbf{Zhenhuan Liu} received his B.S. degree in automation from Shanghai Jiao Tong University, Shanghai, China in 2022. He is pursuing a Master’s at the Institute of Image Processing and Pattern Recognition, Shanghai Jiao Tong University. His research interest is 3D reconstruction and novel view synthesis in dynamic scenes. 

\vspace{50pt}

\begin{wrapfigure}{l}{\wfwid} 
  \vspace{-20pt}
  \includegraphics[width=\biophotowidth,height=\biophotoheight,clip,keepaspectratio]{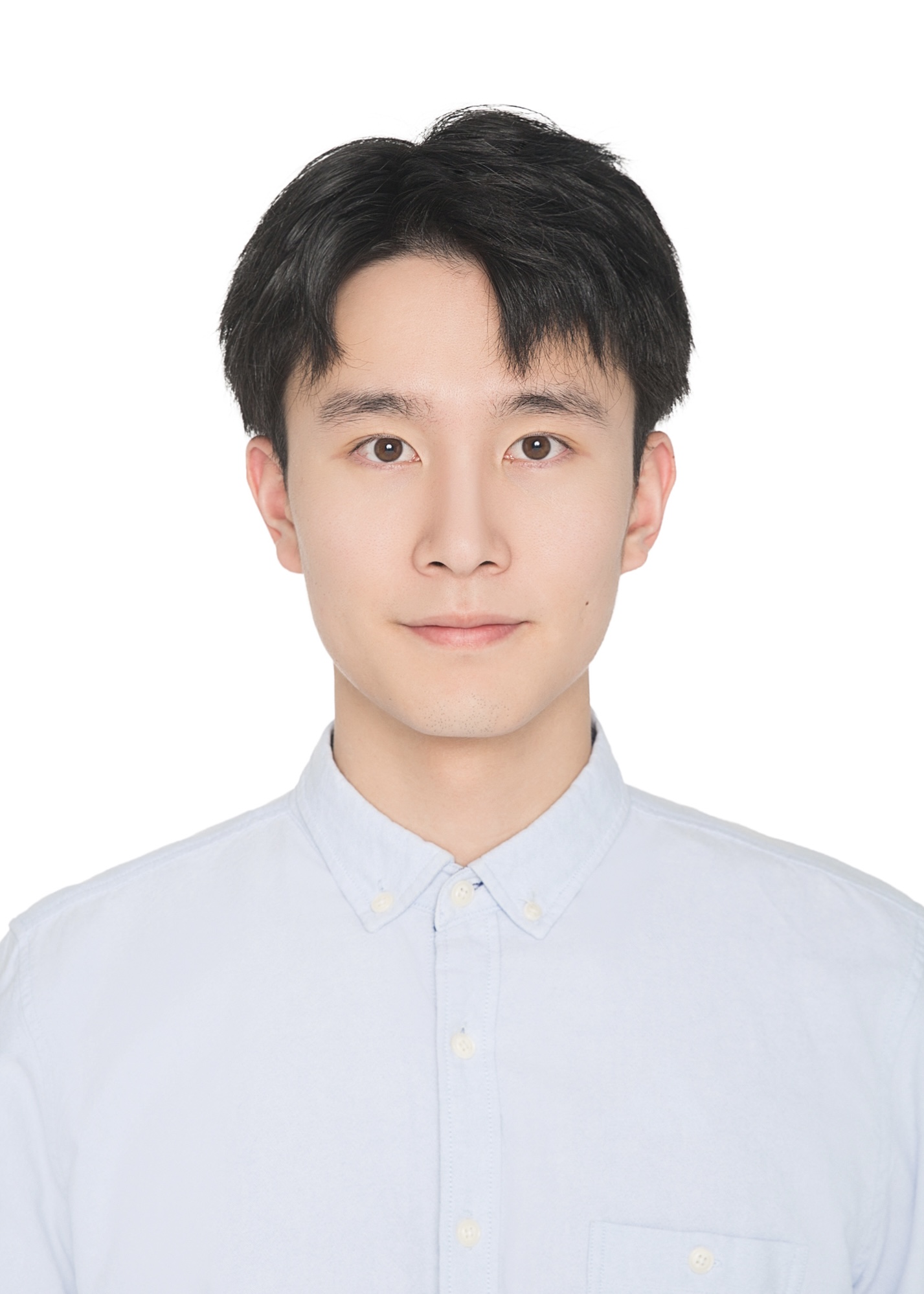}
  \vspace{-10pt}
\end{wrapfigure}
\textbf{Shuai Liu} received his B.S. degree in automation from Dalian Maritime University, Dalian, China in 2023. He is pursuing a Master’s at the Institute of Image Processing and Pattern Recognition, Shanghai Jiao Tong University. His research focuses on 3D reconstruction.
\vspace{20pt}

\begin{wrapfigure}{l}{\wfwid} 
  \vspace{-20pt}
  \includegraphics[width=\biophotowidth,height=\biophotoheight,clip,keepaspectratio]{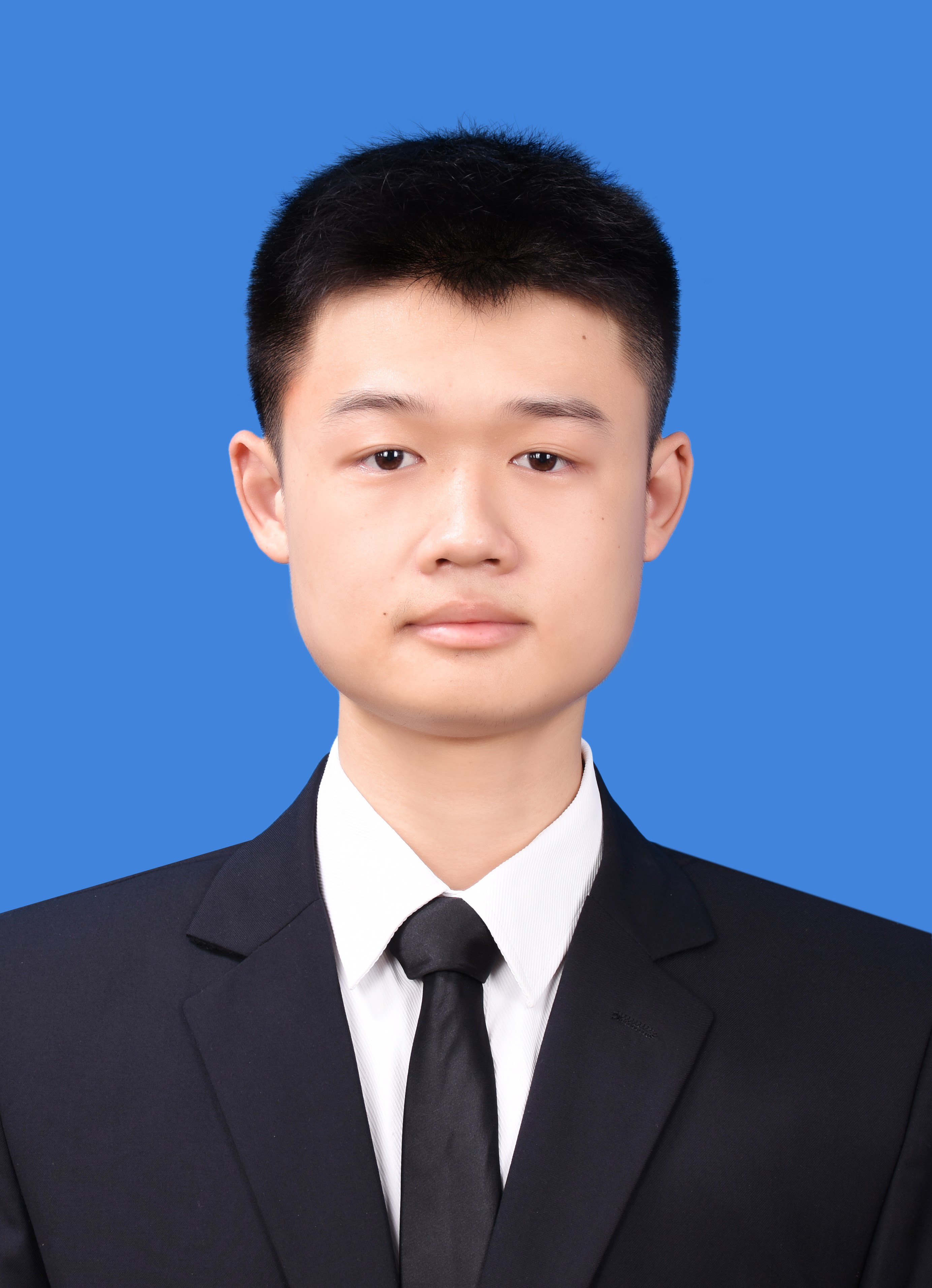}
  \vspace{-10pt}
\end{wrapfigure}
\textbf{Zhiwei Ning} received his B.S. degree in control science and engineering from Xi’an Jiaotong University, Xi’an, China in 2023. He is now pursuing the Ph.D. degree at the Institute of Image Processing and Pattern Recognition, Shanghai Jiao Tong University, Shanghai, China. His research interest is 3D object detection. 
\vspace{10pt}

\begin{wrapfigure}{l}{\wfwid} 
  \vspace{-10pt}
  \includegraphics[width=\biophotowidth,height=\biophotoheight,clip,keepaspectratio]{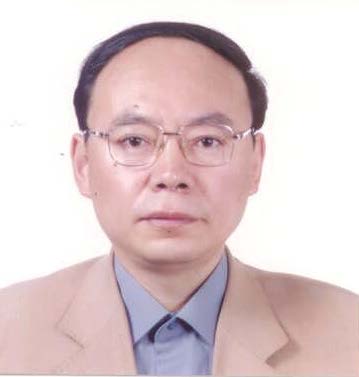}
  \vspace{-10pt}
\end{wrapfigure}
\textbf{Jie Yang} received his Ph.D. from the Department of Computer Science, Hamburg University, Hamburg, Germany, in 1994. Currently, he is a professor at the Institute of Image Processing and Pattern recognition, Shanghai Jiao Tong University, Shanghai, China. He has led many research projects (e.g., National Science Foundation, 863 National High Technique Plan), had one book published in Germany, and authored more than 300 journal papers. His major research interests are object detection and recognition, data fusion and data mining, and medical image processing. 

\vspace{10pt}

\begin{wrapfigure}{l}{\wfwid} 
  \includegraphics[width=\biophotowidth,height=\biophotoheight,clip,keepaspectratio]{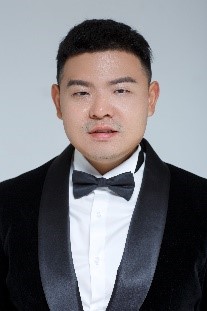}
\end{wrapfigure}
\textbf{Yifan Zuo} received the Ph.D. degree from the University of Technology Sydney, Ultimo, NSW, Australia, in 2018. He is currently an Associate Professor with the School of Computing and Artificial Intelligence, Jiangxi University of Finance and Economics. His research interests include Image/Point Cloud Processing. The corresponding papers have been published in major international journals such as IEEE Transactions on Image Processing, IEEE Transactions on Circuits and Systems for Video Technology, IEEE Transactions on Multimedia, and top conferences such as SIGGRAPH, AAAI.
\vspace{10pt}

\begin{wrapfigure}{l}{\wfwid} 
  \includegraphics[width=\biophotowidth,height=\biophotoheight,clip,keepaspectratio]{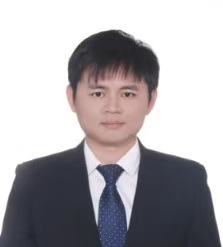}
\end{wrapfigure}
\textbf{Yuming Fang} (M’13-SM’17) received the B.E. degree from Sichuan University, Chengdu, China, the M.S. degree from the Beijing University of Technology, Beijing, China, and the Ph.D. degree from Nanyang Technological University, Singapore. He is currently a Professor with the School of Information Management, Jiangxi University of Finance and Economics, Nanchang, China. His research interests include visual attention modeling, visual quality assessment, computer vision, and 3D image/video processing. He serves on the editorial board for IEEE Transactions on Multimedia and Signal Processing: Image Communication.
\vspace{10pt}

\begin{wrapfigure}{l}{\wfwid} 
  \includegraphics[width=\biophotowidth,height=\biophotoheight,clip,keepaspectratio]{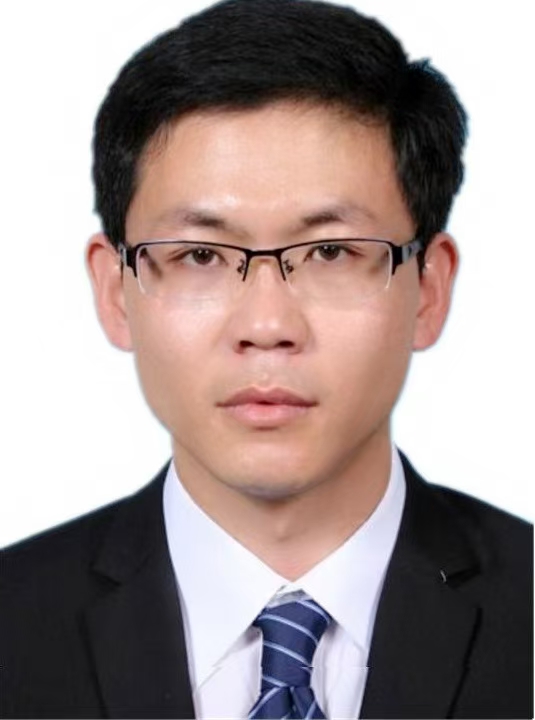}
\end{wrapfigure}
\textbf{Wei Liu} received the B.S. degree from Xi'an Jiaotong University, Xi'an, China, in 2012. He received the Ph.D. degree from Shanghai Jiao Tong University, Shanghai, China in 2019. He was a research fellow in The University of Adelaide and The University of Hong Kong from 2018 to 2021 and 2021 to 2022, respectively. He has been working as an associate professor in Shanghai Jiao Tong University since 2022. His current research areas include image filtering, 3D detection, 3D reconstruction and self-supervised depth estimation.

\clearpage
\onecolumn
\appendix
\section{Per-scene results}
We list the per-scene results of the proposed continual learning method with different spatial representations on the DyNeRF dataset in \cref{tab:perscene_dynerf}, and the per-scene results of the methods in \cref{tab: comparison long video} on our self-collected long video dataset in \cref{tab:perscene_long_video}.

\newcommand{\resizeone}{0.77\textwidth}
\newcommand{\resizetwo}{0.5\textwidth}

%------------------table-detailed-Dynerf
\begin{table*}[!htbp]
\centering
\caption{Detailed results of the proposed CD-NGP and other hash-based methods on the DyNeRF dataset.}
\resizebox{\resizeone}{!}{
\begin{tabular}{l|lllllllll}
\toprule
\multirow{2}{*}{Model} & \multicolumn{3}{c}{Flame Salmon} & \multicolumn{3}{c}{Cook Spinach} & \multicolumn{3}{c}{Cut Roasted Beef} \\
& PSNR$\uparrow$ & DSSIM$\downarrow$ & \multicolumn{1}{l|}{LPIPS$\downarrow$} & PSNR$\uparrow$ & DSSIM$\downarrow$ & \multicolumn{1}{l|}{LPIPS$\downarrow$} & PSNR$\uparrow$ & DSSIM$\downarrow$ & LPIPS$\downarrow$ \\ \midrule
CD-NGP & 26.37 & 0.046 & 0.289 & 30.76 & 0.024 & 0.179 & 31.03 & 0.024 & 0.180 \\
3D+4D & 25.72 &0.059 &0.355 & 29.73 & 0.032 & 0.220 &30.06 &0.031 &0.224 \\
CD-NGP-C & 26.65 & 0.042 & 0.268 & 31.03 & 0.022 & 0.159 & 31.26 & 0.022 & 0.158 \\
CD-Plane & 25.48 & 0.060 & 0.317 & 29.41 & 0.034 & 0.201 & 30.11 & 0.033 & 0.195 \\
CD-MERF & 26.15 & 0.049 & 0.292 & 30.35 & 0.029 & 0.186 & 30.71 & 0.028 & 0.186 \\
CD-MERF-C & 26.56 & 0.044 & 0.255 & 30.69 & 0.026 & 0.158 & 31.04 & 0.025 & 0.163 \\ \midrule
\multirow{2}{*}{Model} & \multicolumn{3}{c}{Flame Steak} & \multicolumn{3}{c}{Sear Steak} & \multicolumn{3}{c}{Average} \\
& PSNR$\uparrow$ & DSSIM$\downarrow$ & \multicolumn{1}{l|}{LPIPS$\downarrow$} & PSNR$\uparrow$ & DSSIM$\downarrow$ & \multicolumn{1}{l|}{LPIPS$\downarrow$} & PSNR$\uparrow$ & DSSIM$\downarrow$ & LPIPS$\downarrow$ \\ \midrule
CD-NGP & 31.27 & 0.021 & 0.174 & 31.71 & 0.019 & 0.170 & 30.23 & 0.027 & 0.198 \\
3D+4D & 30.33 & 0.028 & 0.205 & 30.73 & 0.026 &0.207 & 29.31 & 0.035 & 0.242 \\
CD-NGP-C & 31.64 & 0.018 & 0.140 & 32.02 & 0.017 & 0.142 & 30.52 & 0.025 & 0.174 \\
CD-Plane & 30.26 & 0.029 & 0.180 & 30.56 & 0.027 & 0.181 & 29.16 & 0.037 & 0.215 \\
CD-MERF & 31.01 & 0.025 & 0.172 & 31.44 & 0.022 & 0.170 & 29.93 & 0.031 & 0.201 \\
CD-MERF-C & 31.13 & 0.024 & 0.151 & 31.66 & 0.020 & 0.141 & 30.22 & 0.028 & 0.174 \\ \bottomrule
\end{tabular}
}
\label{tab:perscene_dynerf}
\end{table*}
%-----------------------------------------
\vspace{1pt}
\begin{table*}[!ht]
\centering
\caption{Per-scene results of the compared methods on the meetroom dataset in \cref{tab:comparison_meetroom_dataset}.}
\resizebox{\resizetwo}{!}{
\begin{tabular}{l|llllll}
\toprule
\multirow{2}{*}{Model} & \multicolumn{3}{c}{Trimming} & \multicolumn{3}{c}{Discussion} \\
& PSNR$\uparrow$ & DSSIM$\downarrow$ & \multicolumn{1}{l|}{LPIPS$\downarrow$} & PSNR$\uparrow$ & DSSIM$\downarrow$ & LPIPS$\downarrow$ \\ \midrule
MixVoxels & 27.21 & 0.026 & 0.093 & 27.42 & 0.026 & 0.110 \\
HexPlane & 22.13 & 0.062 & 0.263 & 26.28 & 0.032 & 0.212 \\ \midrule
HexPlane-PR & 17.30 & 0.140 & 0.404 & 18.53 & 0.111 & 0.355 \\
HexPlane-GR & 12.80 & 0.184 & 0.554 & 23.31 & 0.061 & 0.330 \\
StreamRF & 26.72 & 0.029 & 0.170 & 26.65 & 0.028 & 0.169 \\
INV & 27.08 & 0.026 & 0.089 & 24.57 & 0.036 & 0.110 \\
CD-NGP & 27.58 & 0.040 & 0.164 & 26.01 & 0.040 & 0.214 \\ \midrule
\multirow{2}{*}{Model} & \multicolumn{3}{c}{VRheadset} & \multicolumn{3}{c}{Average} \\
& PSNR$\uparrow$ & DSSIM$\downarrow$ & \multicolumn{1}{l|}{LPIPS$\downarrow$} & PSNR$\uparrow$ & DSSIM$\downarrow$ & LPIPS$\downarrow$ \\ \midrule
MixVoxels & 26.25 & 0.029 & 0.107 & 26.96 & 0.027 & 0.103 \\
HexPlane & 25.72 & 0.035 & 0.187 & 24.71 & 0.043 & 0.221 \\ \midrule
HexPlane-PR & 20.05 & 0.085 & 0.317 & 18.63 & 0.112 & 0.359 \\
HexPlane-GR & 22.37 & 0.054 & 0.316 & 19.49 & 0.099 & 0.400 \\
StreamRF & 26.06 & 0.029 & 0.171 & 26.48 & 0.029 & 0.170 \\
INV & 19.95 & 0.061 & 0.160 & 23.87 & 0.041 & 0.120 \\
CD-NGP & 24.43 & 0.037 & 0.193 & 26.01 & 0.039 & 0.191 \\ \bottomrule
\end{tabular}
}
\label{tab:perscene_meetroom_dataset}
\end{table*}
\vspace{1pt}
%------------------table-detailed-long video
\begin{table*}[!htbp]
\centering
\caption{Detailed results of the proposed CD-NGP and other methods on our long video dataset.}
\resizebox{\resizeone}{!}{
\begin{tabular}{l|lllllllll}
\toprule
\multirow{2}{*}{Model} & \multicolumn{3}{c}{Demo} & \multicolumn{3}{c}{Discussion} & \multicolumn{3}{c}{Kungfu} \\
& PSNR$\uparrow$ & DSSIM$\downarrow$ & \multicolumn{1}{l|}{LPIPS$\downarrow$} & PSNR$\uparrow$ & DSSIM$\downarrow$ & \multicolumn{1}{l|}{LPIPS$\downarrow$} & PSNR$\uparrow$ & DSSIM$\downarrow$ & LPIPS$\downarrow$ \\ \midrule
MixVoxels & 27.50 & 0.021 & 0.103 & 28.01 & 0.016 & 0.097 & 24.43 & 0.033 & 0.140 \\
HexPlane & 28.92 & 0.016 & 0.095 & 27.17 & 0.024 & 0.136 & 25.14 & 0.032 & 0.136 \\
4DGS & 28.38 & 0.022 & 0.092 & 29.92 & 0.016 & 0.064 & 25.53 & 0.029 & 0.099 \\ \midrule
HexPlane-PR & 21.10 & 0.090 & 0.268 & 20.98 & 0.094 & 0.315 & 18.85 & 0.112 & 0.309 \\
HexPlane-GR & 24.85 & 0.041 & 0.243 & 24.03 & 0.050 & 0.296 & 22.15 & 0.072 & 0.336 \\
StreamRF & 24.86 & 0.032 & 0.233 & 24.41 & 0.029 & 0.258 & 23.75 & 0.043 & 0.301 \\
INV & 26.55 & 0.025 & 0.093 & 28.51 & 0.017 & 0.054 & 24.57 & 0.037 & 0.120 \\
CD-NGP & 26.66 & 0.030 & 0.161 & 29.66 & 0.057 & 0.137 & 23.66 & 0.055 & 0.257 \\ \midrule
\multirow{2}{*}{Model} & \multicolumn{3}{c}{Reading} & \multicolumn{3}{c}{Seminar} & \multicolumn{3}{c}{Average} \\
& PSNR$\uparrow$ & DSSIM$\downarrow$ & \multicolumn{1}{l|}{LPIPS$\downarrow$} & PSNR$\uparrow$ & DSSIM$\downarrow$ & \multicolumn{1}{l|}{LPIPS$\downarrow$} & PSNR$\uparrow$ & DSSIM$\downarrow$ & LPIPS$\downarrow$ \\ \midrule
MixVoxels & 25.99 & 0.021 & 0.144 & 28.78 & 0.018 & 0.108 & 26.94 & 0.022 & 0.118 \\
HexPlane & 27.88 & 0.018 & 0.099 & 29.59 & 0.018 & 0.120 & 27.74 & 0.022 & 0.117 \\
4DGS & 29.38 & 0.014 & 0.067 & 31.02 & 0.015 & 0.067 & 28.85 & 0.019 & 0.078 \\ \midrule
HexPlane-PR & 21.85 & 0.087 & 0.287 & 21.78 & 0.092 & 0.280 & 20.91 & 0.095 & 0.292 \\
HexPlane-GR & 26.52 & 0.040 & 0.266 & 25.77 & 0.053 & 0.306 & 24.66 & 0.051 & 0.289 \\
StreamRF & 24.06 & 0.040 & 0.325 & 26.53 & 0.037 & 0.250 & 24.72 & 0.036 & 0.273 \\
INV & 26.39 & 0.026 & 0.106 & 29.60 & 0.018 & 0.054 & 27.12 & 0.025 & 0.085 \\
CD-NGP & 26.70 & 0.031 & 0.212 & 28.97 & 0.024 & 0.169 & 27.13 & 0.039 & 0.201 \\ \bottomrule
\end{tabular}
}
\label{tab:perscene_long_video}
\end{table*}

%-----------------------------------------

\section{Access to the long video dataset.}

We provide a public link to access our self-collected long video dataset. 
It can be accessed via the following BaiduNetdisk link:

\url{https://pan.baidu.com/s/1zN_K_PwTqdunt2J9vxKPkA?pwd=dv2d}.

Or via the following Google Drive link:
\url{https://drive.google.com/drive/folders/1eIT-Uk4R49P6Tm_2AirWj4kgymzOowKT?usp=sharing}. 

\section{Discussion on rendering latency.}

The proposed method CD-NGP fuses spatial and temporal features in a multi-branch NeRF model, which is efficient in terms of training and memory consumption. Moreover, the feature combination in the latent space avoids querying multiple fields for rendering, and thus achieves better rendering latency.
However, the volume rendering process adopted from NeRF-variants is inherently computationally intensive, which limits the real-time rendering capability. 
In this section, we provide a detailed analysis of the rendering latency of our method on the DyNeRF dataset. 
Taking the \emph{cut roasted beef} scene in the DyNeRF dataset (at $1352\times 1014$ resolution) as an example, we provide a detailed analysis of the rendering latency in \cref{tab:rendering_time}. The rendering process of CD-NGP is divided into three parts: ray marching, NeRF model, and pixel-level volume rendering as follows: 
\begin{itemize}
\item Ray marching: The ray marching process is the same as that in the implementation \cite{ngppl} of Instant-NGP \cite{ingp}, which is a standard process in hash-based NeRF. This process generates a split (batch) of sampling coordinates from the camera to the scene along each ray, and each image requires multiple passes to render. Typically, each pass generates a split of up to 9M samples on the 1M rays (1M pixels), and rendering the entire image requires around 200M samples.
\item NeRF model: As elaborated in the manuscript, our CD-NGP is a multi-branch NeRF model that incorporates hash-based encoders and MLP-based decoders. As shown in \cref{tab:rendering_time}, both the spatial and temporal hash encoders are efficient and lightweight, requiring only 7 ms and 5 ms, respectively. The primary computational cost arises from the two MLP decoders (55 ms in total), which is common and inevitable in NeRF-based methods. Although CD-NGP avoids multiple passes of MLP decoders by fusing features in the latent space, one fused decoding pass of MLP decoders is still required for each sampled point. This fused pass takes approximately 33 ms for the XYZT-fusion MLP and 22 ms for the RGB MLP, resulting in a total NeRF model latency of around 76 ms per pass. 
\item Volume rendering: This process loads the optical attributes $\sigma, \textbf{c}$ of the sampled points and computes the pixel color $\textbf{C}$ by numerical integral (accumulating the optical attributes along the ray). The rendering process is similar to that of Instant-NGP \cite{ingp} and other NeRF-based methods \cite{ngppl}. If the accumulated opacity of a ray (pixel) reaches $10^{-4}$, the sampling and rendering are terminated. Thus, the whole rendering time is smaller than the product of the number of passes and the time of each pass.
\end{itemize}

\begin{table}[!ht]
\centering
\caption{Rendering time breakdown and point scale of each component.}
\begin{tabular}{@{}l@{\hspace{16pt}}r@{\hspace{16pt}}l@{}}
\toprule
\textbf{Module} & \textbf{Time} & \textbf{Scale} \\
\cmidrule(lr){1-3}
Ray marching & 3ms & $\sim$1M pixels \\
\cmidrule(lr){1-3}
\textbf{NeRF model} & & \\
\quad\quad Spatial representation & 7ms & $\sim$9M samples \\
\quad\quad Time representation & 5ms & $\sim$9M samples \\
\quad\quad XYZT-fusion MLP & 33ms & $\sim$9M samples \\
\quad\quad RGB MLP & 22ms & $\sim$9M samples \\
\quad\quad Total (NeRF model) & 76ms & $\sim$9M samples \\
\cmidrule(lr){1-3}
Rendering a split & 2ms & $\sim$1M pixels \\
Rendering the whole image & 500ms & $\sim$200M samples \\
\bottomrule
\end{tabular}
\label{tab:rendering_time}
\end{table}

As illustrated, each query of the hash table-based NeRF model takes less than 100ms, but the pixel-based volume rendering process requires multiple passes to render the whole image. This limitation is a common issue in all NeRF-based methods and is orthogonal to the contribution of the proposed feature fusion method and continual learning pipeline.

\section{Comparison with real-time processing method.}

% 我们的方法着重于使用高scalability的持续学习方法解决动态场景NVS问题，其一大优势是减少memory consumption。然而，也有一些NVS方法支持real-time processing，也能减少memory consumption，拥有在动态场景上进行应用的潜质。在这里，我们选取一个具有代表性的方法MVSGS进行比较，从而验证continual learning在动态场景NVS问题的必要性。
%
%

\begin{figure*}[!ht]
\center{
\begin{tabular}{c} %
\includegraphics[width=0.8\textwidth]{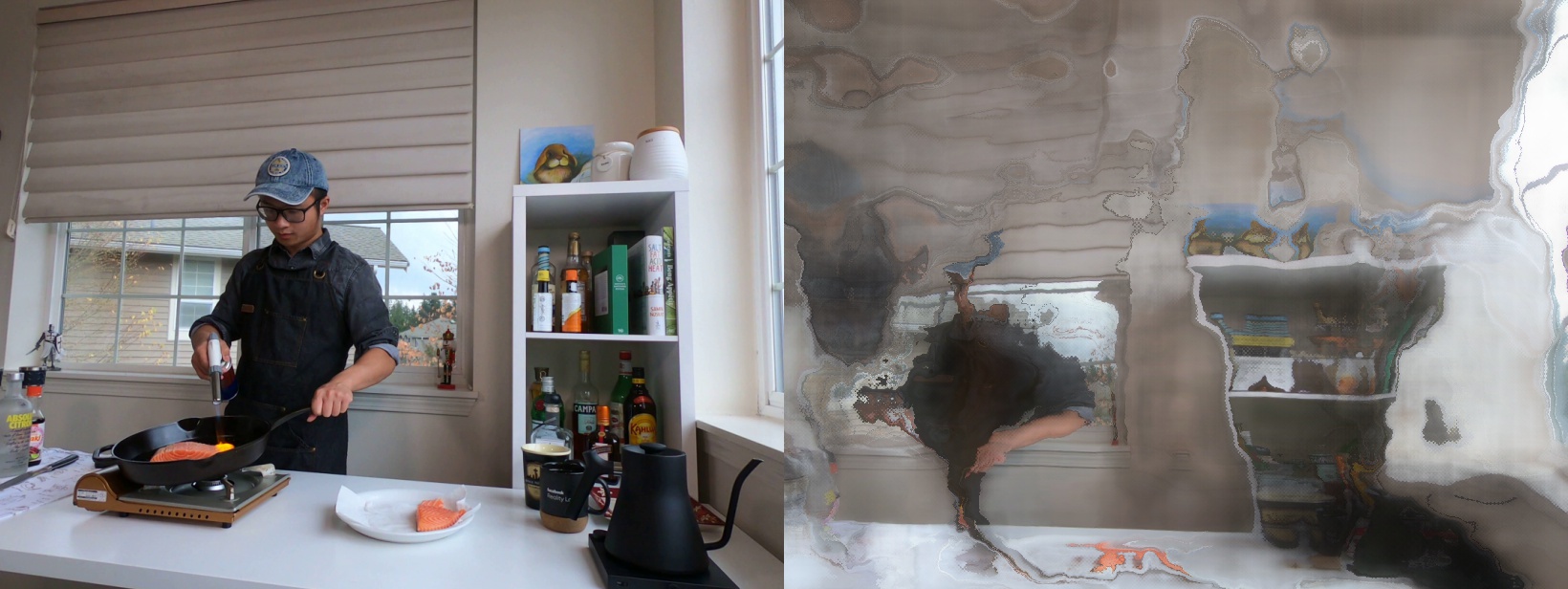}\\
\includegraphics[width=0.8\textwidth]{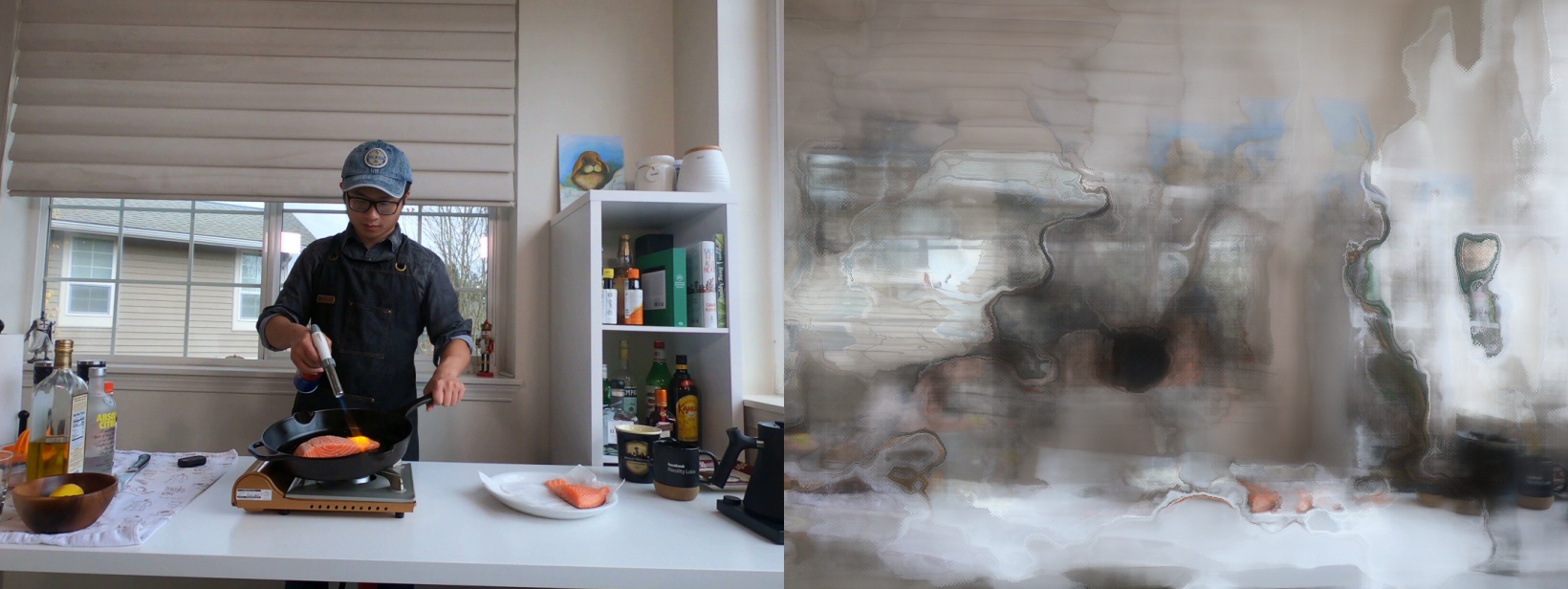}\\
\includegraphics[width=0.8\textwidth]{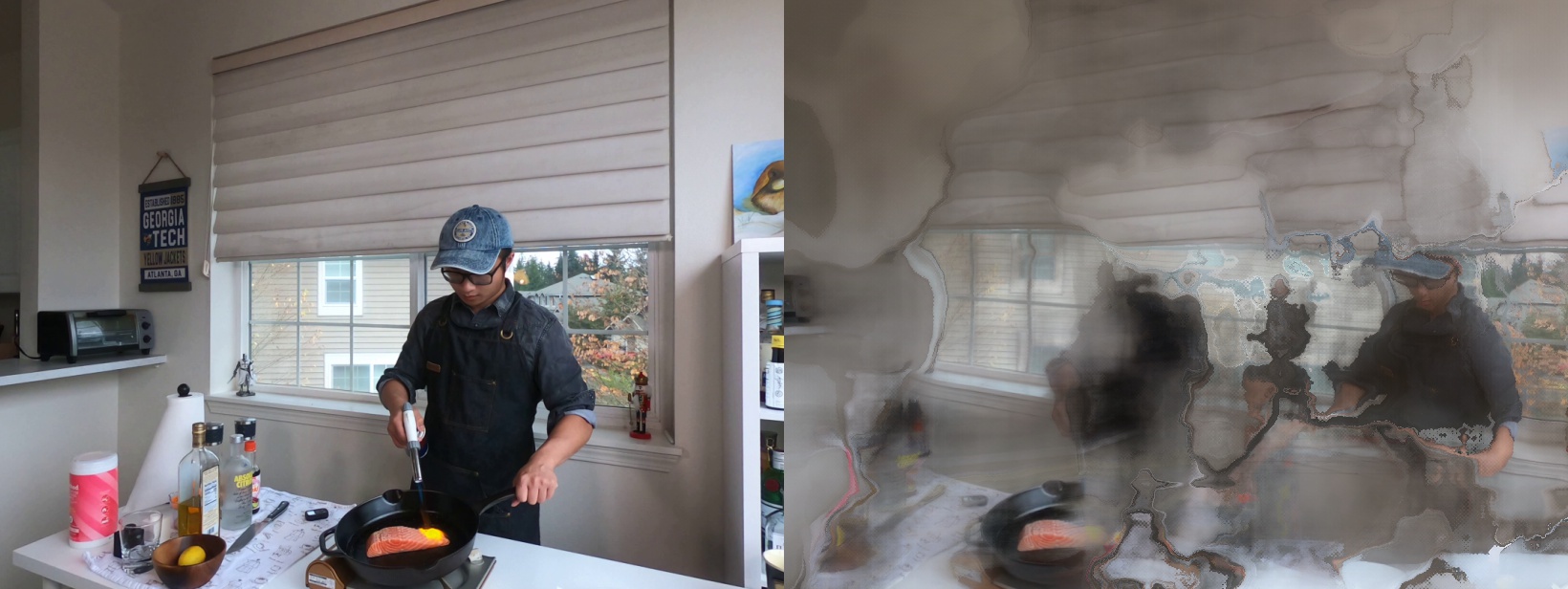}\\
\includegraphics[width=0.8\textwidth]{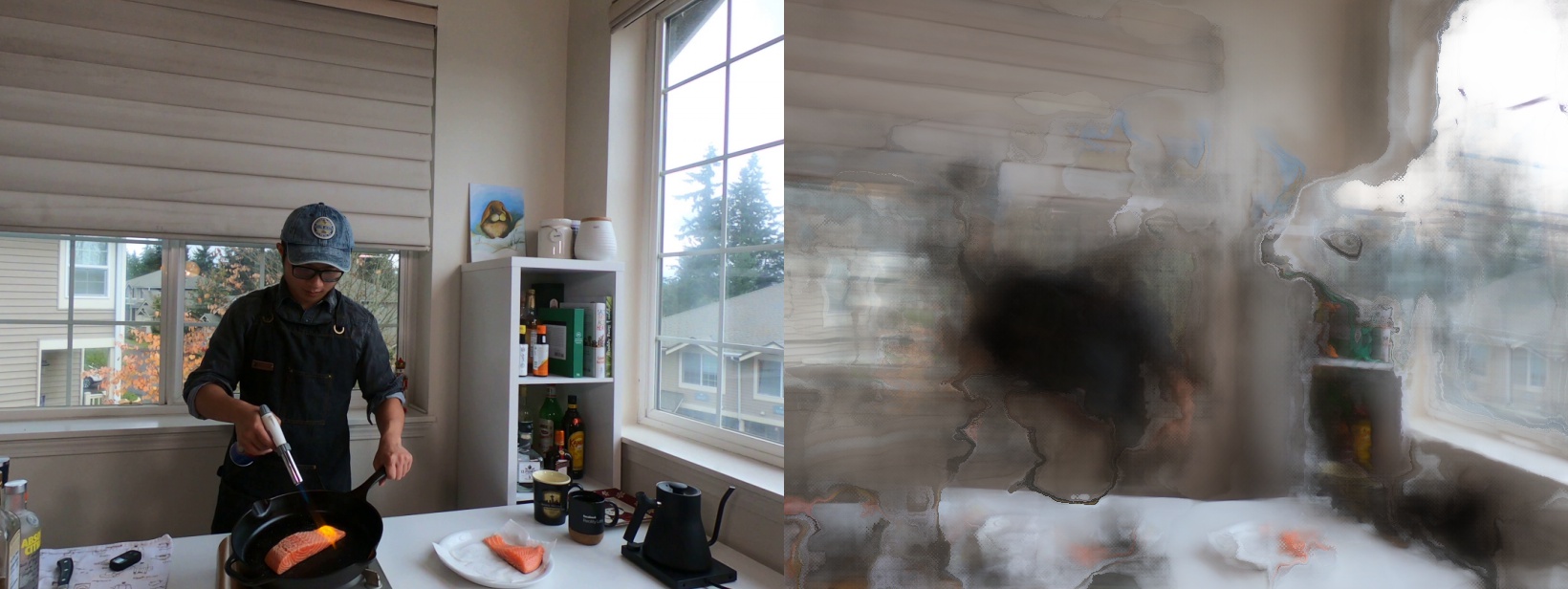}\\
\end{tabular}}
\caption{Output images of MVSGS-D on the DyNeRF dataset. The left side is the ground truth, and the right side is the rendering output. }
\label{fig:mvsgs_d_render}
\end{figure*}

We compare the per-scene results of our method and MVSGS-FT on the DyNeRF dataset in \cref{tab:metrics_comp_with_mvsgs}. Our method produces comparable quality with MVSGS-FT in the first 4 scenes and is more robust in the last \emph{sear steak} scene. The method MVSGS-D does not support the standard $1352\times 1014$ resolution on the DyNeRF dataset and produces limited quality as shown in \cref{fig:mvsgs_d_render}, we thus exclude it from \cref{tab:metrics_comp_with_mvsgs}.

\begin{figure*}[!ht]
\center{
\begin{tabular}{cc} %
\includegraphics[width=0.4\textwidth]{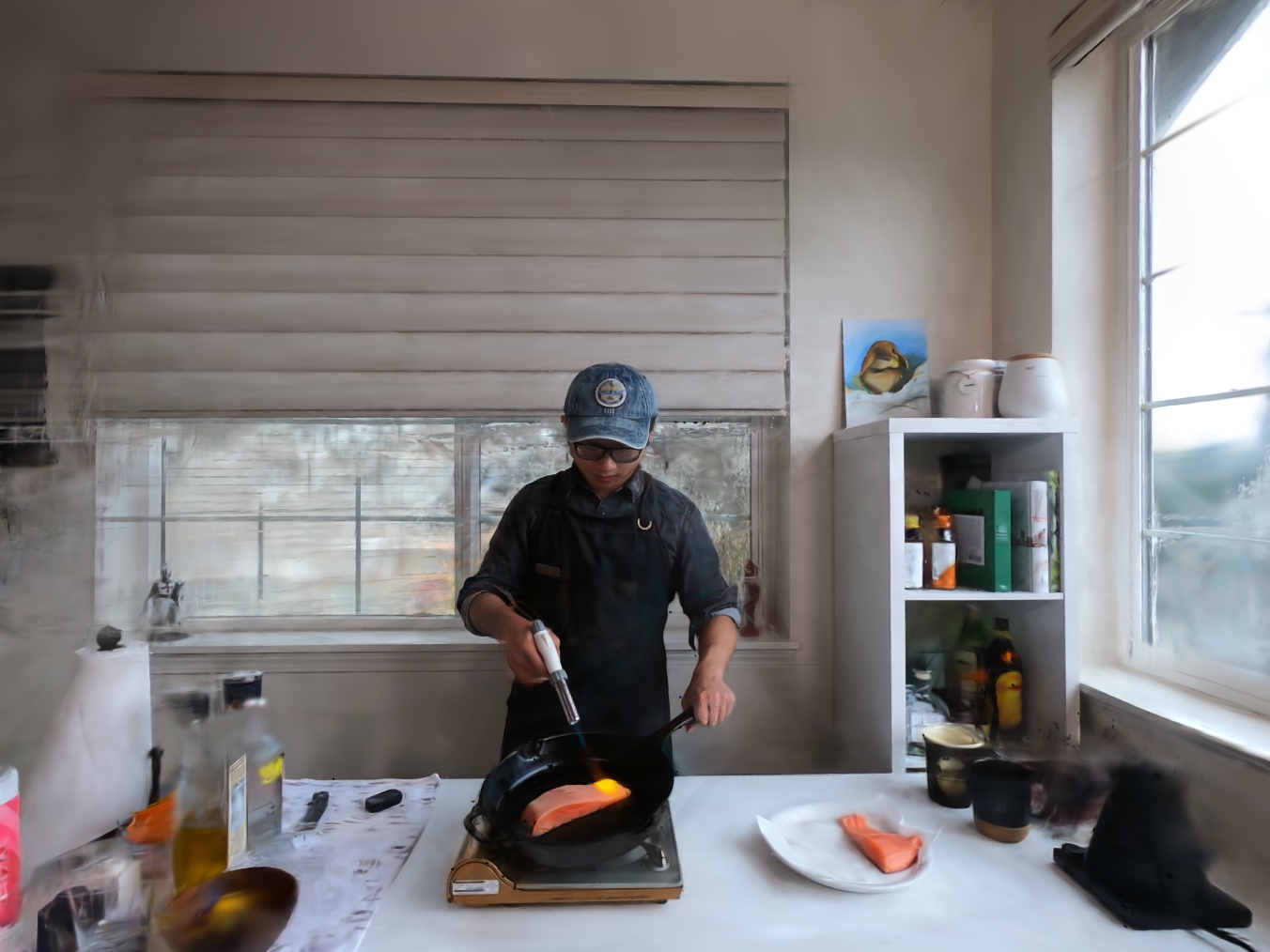} &
\includegraphics[width=0.4\textwidth]{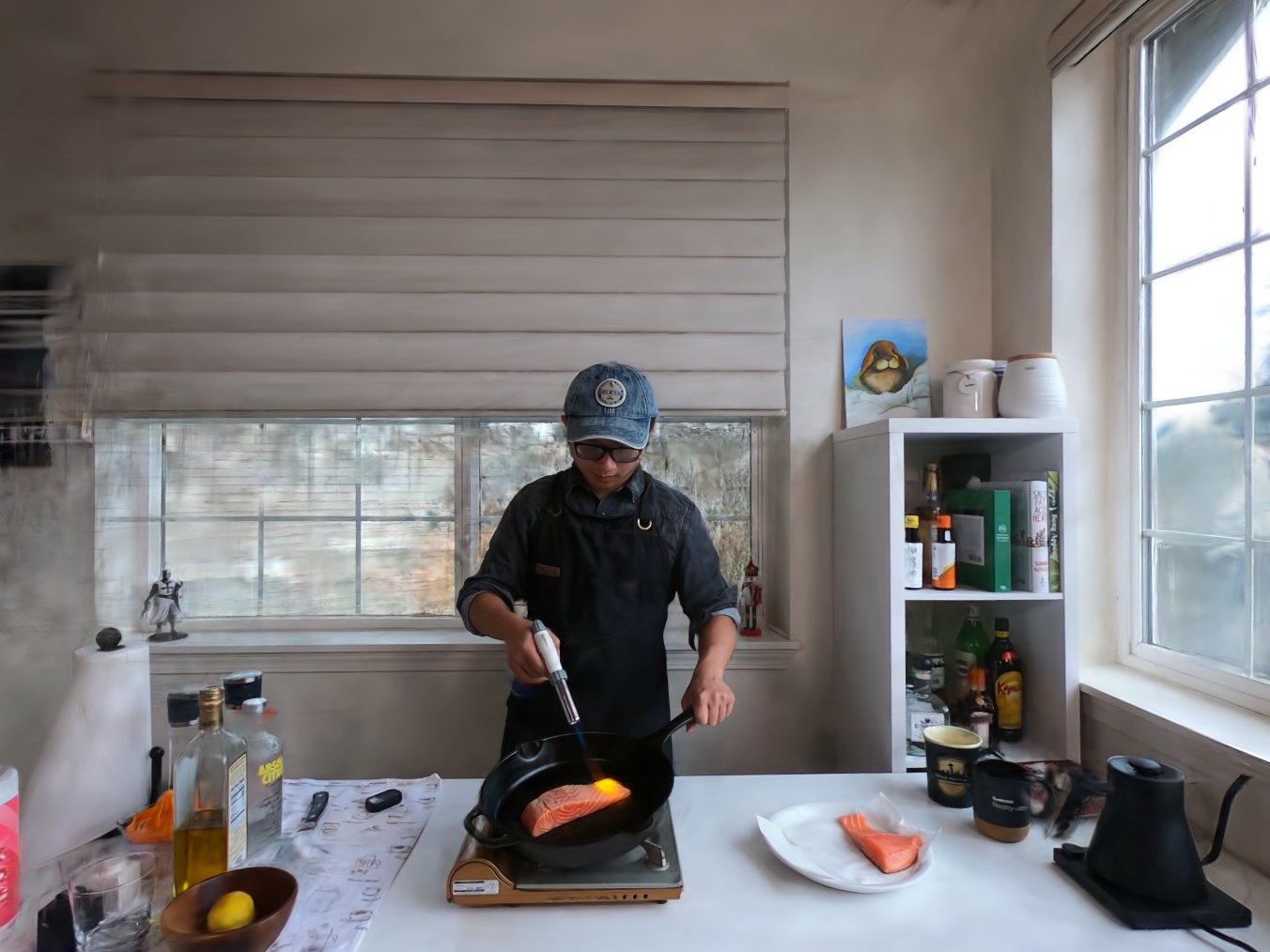} \\
(a) & (b)\\
\end{tabular}}
\caption{Output images of MVSGS-FT on the DyNeRF dataset. (a) iteration 3,000, (b) iteration 5,000. }
\label{fig:mvsgs_ft_render}
\end{figure*}

\begin{table*}[!htbp]
\centering
\caption{Detailed results of the proposed CD-NGP and real-time processing method MVSGS.}
\resizebox{\resizeone}{!}{
\begin{tabular}{l|lllllllll}
\toprule
\multirow{2}{*}{Model} & \multicolumn{3}{c}{Flame Salmon} & \multicolumn{3}{c}{Cook Spinach} & \multicolumn{3}{c}{Cut Roasted Beef} \\
& PSNR$\uparrow$ & DSSIM$\downarrow$ & \multicolumn{1}{l|}{LPIPS$\downarrow$} & PSNR$\uparrow$ & DSSIM$\downarrow$ & \multicolumn{1}{l|}{LPIPS$\downarrow$} & PSNR$\uparrow$ & DSSIM$\downarrow$ & LPIPS$\downarrow$ \\ \midrule
CD-NGP & 26.37 & 0.046 & 0.289 & 30.76 & 0.024 & 0.179 & 31.03 & 0.024 & 0.180 \\
MVSGS-FT & 24.75 & 0.054 & 0.190 & 30.57 & 0.023 & 0.123 & 29.68 & 0.025 & 0.131 \\ \midrule

\multirow{2}{*}{Model} & \multicolumn{3}{c}{Flame Steak} & \multicolumn{3}{c}{Sear Steak} & \multicolumn{3}{c}{Average} \\
& PSNR$\uparrow$ & DSSIM$\downarrow$ & \multicolumn{1}{l|}{LPIPS$\downarrow$} & PSNR$\uparrow$ & DSSIM$\downarrow$ & \multicolumn{1}{l|}{LPIPS$\downarrow$} & PSNR$\uparrow$ & DSSIM$\downarrow$ & LPIPS$\downarrow$ \\ \midrule
CD-NGP & 31.27 & 0.021 & 0.174 & 31.71 & 0.019 & 0.170 & 30.23 & 0.027 & 0.198 \\
MVSGS-FT & 30.02 & 0.027 & 0.126 & 15.26 & 0.301 & 0.722 & 26.05 & 0.086 & 0.258 \\ \bottomrule
\end{tabular}
}
\label{tab:metrics_comp_with_mvsgs}
\end{table*}

We provide the rendered images of MVSGS-FT for better visualization in \cref{fig:mvsgs_ft_render}. We note that there are two approaches for fine-tuning the MVSGS: the first is to use the output of the MVS network to re-initialize Gaussians for each frame, and the second is to use the Gaussians of the previous frame for initialization. However, we find that the second approach leads to unstable training and produces unsatisfactory results (much lower PSNRs). Therefore, we only report the metrics of the first approach and refer to it as MVSGS-FT.

\section{Implementation Details about the spatial and temporal hash encoding}

\subsection{Implementation}
In the implementation of tiny-cuda-nn, the authors have provided interfaces for hash-encoders with different input dimensions, such as 1D, 2D, and 3D. However, the 1D hash encoding is not enabled by default. In our implementation, we enable the interfaces for one-dimension hash encodings of tiny-cuda-nn \cite{tcnn} to support the temporal encoding in our continual learning framework. Specifically, we use the 1D hash encoding for the time dimension and the 3D hash encoding for the spatial dimensions.

\subsection{Scalability on size. }
We measure the hash encoder size and the inference latency of different configurations of the encodings in \cref{tab:hash_config_scalability_detail_com}. The latency is measured on a single NVIDIA RTX 3090 GPU.
Since all hash tables are stored in float32 precision, each element of the table consumes $F$ parameters, and the hash encoder contains $L$ hash tables for different encoding resolutions. The upper bound of the total hash encoder size is $4 \times L \times F \times 2^{P}$ bytes. Since some spatial hash table resolutions are coarse (e.g., $16 \times 16 \times 16 = 2^{12} < 2^{14} < 2^{P_1=19}$), and the tables are not fully filled, the actual hash encoder size is smaller than the upper bound.
Clearly, the hash encoder size is primarily influenced by the length of the hash tables (i.e., $2^P$). By controlling the length of the hash tables and reusing features from the base spatial hash table, we can achieve a good trade-off between hash encoder size and rendering quality.

\subsection{Scalability on inference speed. }
\Cref{tab:hash_config_scalability_detail_com} additionally presents the inference speed across different configurations. We render the first frame of the \emph{cut roasted beef} scene from the DyNeRF dataset, which has a resolution of $1352\times 1014$. The latency increases approximately linearly with both the number of layers $L$ and the feature vector length $F$, with $L$ exhibiting a greater impact. Increasing the hash table length $P$ also affects inference speed, primarily due to reduced memory locality for larger table sizes.

\begin{table}[htbp]
\centering
\caption{Inference speed and model size under different configurations. L: levels, F: feature dim, P: log2 values of the hash table length. Base and auxiliary denote the hash encoders of the base spatial representation and auxiliary spatial representation, respectively. Temporal denotes the hash encoder of the time dimension. }
\begin{tabular}{lcccccc}
\toprule
\textbf{Type} & \textbf{L} & \textbf{F} & \textbf{P} & \textbf{Size (MB)} & \textbf{Samples} & \textbf{Latency(ms)} \\
\midrule
base & 12 & 2 & 19 & 37.83 & 5483712 & 3.33 \\
auxiliary & 12 & 2 & 14 & 1.41 & 5483712 & 3.16 \\
temporal & 2 &20 & 7 & 0.006 & 5483712 & 2.89 \\ \midrule
base & 16 & 2 & 19 & 51.03 & 5483712 & 4.48 \\
auxiliary & 16 & 2 & 14 & 1.91 & 5483712 & 4.27 \\
temporal & 2 &20 & 7 & 0.006 & 5483712 & 2.96 \\ \midrule
base & 24 & 2 & 19 & 77.53 & 5483712 & 6.82 \\
auxiliary & 24 & 2 & 14 & 2.86 & 5483712 & 6.51 \\
temporal & 2 &20 & 7 & 0.006 & 5483712 & 2.89 \\ \midrule
base & 12 & 4 & 19 & 75.65 & 5483712 & 5.33 \\
auxiliary & 12 & 4 & 14 & 2.82 & 5483712 & 5.08 \\
temporal & 2 &20 & 7 & 0.006 & 5483712 & 3.01 \\ \midrule
base & 12 & 2 & 21 &140.83 & 5483712 & 3.64 \\
auxiliary & 12 & 2 & 14 & 1.41 & 5483712 & 3.16 \\
temporal & 2 &20 & 7 & 0.006 & 5483712 & 3.19 \\ \midrule
base & 12 & 2 & 20 & 73.82 & 5483712 & 3.45 \\
auxiliary & 12 & 2 & 14 & 1.41 & 5483712 & 3.16 \\
temporal & 2 &20 & 7 & 0.006 & 5483712 & 3.23 \\ \midrule
base & 12 & 2 & 19 & 37.83 & 5483712 & 3.31 \\
auxiliary & 12 & 2 & 16 & 5.26 & 5483712 & 3.24 \\
temporal & 2 &20 & 7 & 0.006 & 5483712 & 2.87 \\ \midrule
base & 12 & 2 & 19 & 37.83 & 5483712 & 3.30 \\
auxiliary & 12 & 2 & 14 & 1.41 & 5483712 & 3.14 \\
temporal & 1 &20 & 7 & 0.003 & 5483712 & 1.39 \\ \midrule
base & 12 & 2 & 19 & 37.83 & 5483712 & 3.29 \\
auxiliary & 12 & 2 & 14 & 1.41 & 5483712 & 3.15 \\
temporal & 1 &40 & 7 & 0.006 & 5483712 & 2.87 \\ \bottomrule
\end{tabular}
\label{tab:hash_config_scalability_detail_com}
\end{table}

\end{document}